\documentclass[journal]{IEEEtran}
\usepackage{graphicx,color,psfrag, caption, subcaption}
\usepackage[cmex10]{amsmath}
\usepackage{amssymb} 
\usepackage{moreverb,url}

\begin{document}
\onecolumn
This manuscript has been submitted to Robotics and Autonomous Systems for possible publication. Once this manuscript is accepted for publication, the copyright will be transferred without notice. 
\newpage
\twocolumn	
	%
	\title{Optical-Flow based Self-Supervised Learning of Obstacle Appearance applied to MAV Landing}
	%
	%
	%
	
	\author{H.~W.~Ho,
		C.~De~Wagter,
		B.~D.~W.~Remes,
		and~G.~C.~H.~E.~de~Croon
		\thanks{H.~W.~Ho is with the Micro Air Vehicle laboratory of the Faculty of Aerospace Engineering, Delft University of Technology, The Netherlands and School of Aerospace Engineering, Universiti Sains Malaysia, Malaysia \texttt{H.W.Ho@tudelft.nl}.}
		\thanks{C.~De~Wagter, B.~D.~W.~Remes, G.~C.~H.~E.~de~Croon are with the Micro Air Vehicle laboratory of the Faculty of Aerospace Engineering, Delft University of Technology, The Netherlands (e-mail: C.deWagter@TUDelft.nl; B.D.W.Remes@TUDelft.nl; G.C.H.E.deCroon@TUDelft.nl).}
	}

	\maketitle
	
	
	\begin{abstract}
		Monocular optical flow has been widely used to detect obstacles in Micro Air Vehicles (MAVs) during visual navigation. However, this approach requires significant movement, which reduces the efficiency of navigation and may even introduce risks in narrow spaces. In this paper, we introduce a novel setup of self-supervised learning (SSL), in which optical flow cues serve as a scaffold to learn the visual appearance of obstacles in the environment. We apply it to a landing task, in which initially `surface roughness' is estimated from the optical flow field in order to detect obstacles. Subsequently, a linear regression function is learned that maps appearance features represented by texton distributions to the roughness estimate. After learning, the MAV can detect obstacles by just analyzing a still image. This allows the MAV to search for a landing spot without moving. We first demonstrate this principle to work with offline tests involving images captured from an on-board camera, and then demonstrate the principle in flight. Although surface roughness is a property of the entire flow field in the global image, the appearance learning even allows for the pixel-wise segmentation of obstacles.  
	\end{abstract}
	
	
	%
	\IEEEpeerreviewmaketitle

\section{Introduction}
\label{sec:Introduction}
\IEEEPARstart{T}{o} reduce the risk and cost of human intervention, autonomous flight of Micro Air Vehicles (MAVs) is highly desired in many circumstances. In particular, autonomous landing is a challenging, but essential task of the flight, as it needs to be done in a limited space and time \cite{sanchez2014approach}. Hence, quickly searching for a safe landing spot is required during landing in autonomous flights \cite{mejias2006two}. 

Many existing methods for finding a suitable landing spot use multiple cameras \cite{goldberg2002stereo,brockers2014micro,park2012landing} or active sensors such as a laser range finder \cite{achtelik2009stereo,bachrach2009autonomous} to estimate the distance to many points on the landing surface. While both methods can provide accurate measurements, their perception range is limited and they are heavy and costly for small MAVs. Therefore, use of a single camera is preferable as it is light-weight and has low power consumption \cite{li2015monocular}.

State-of-the-art algorithms for autonomous landing purely rely on motion cues. There are two main approaches: (1) visual Simultaneous Localization and Mapping (SLAM) and (2) bio-inspired optical flow control. 

The first approach can be categorized into feature-based and direct methods to solve the SLAM problems, i.e., to determine the vehicle's location and 3D-structure of the surrounding environment. The feature-based method \cite{davison2007monoslam,li2013high} decouples these SLAM problems into extraction of features and computation of camera pose and scene geometry based on tracking these features \cite{celik2008mvcslam,blosch2010vision,weiss2011monocular}. However, this approach is not sufficiently robust to challenging scenes where the features are hardly detected. The direct method tries to avoid this limitation by using image intensities directly to generate (semi-) dense maps \cite{newcombe2011dtam,engel2013semi,schops2014semi}. Although the computational efficiency and accuracy of visual SLAM have been improved over the years \cite{li2013high,engel2014lsd,forster2014svo,Desaraju-RSS-14,mur2015orb}, this approach still uses more computational resources than are strictly necessary. 

The second approach is inspired by flying insects, which heavily rely on optical flow for navigation. Biologists first found that honeybees perform a grazing landing by keeping the ventral flow (lateral velocities divided by height) constant \cite{baird2005visual,baird2006visual,franceschini2009neuromimetic,ruffier2008aerial}. This approach guarantees a soft landing but does not control its vertical dynamics. To deal with that, recent studies proposed using time-to-contact (height divided by vertical velocity) \cite{herisse2010landing,izzo2012landing,baird2013universal}. The optical flow field can also be used during landing to identify and avoid obstacles \cite{roberts2013optical,de2013optic}.

For sensing obstacles with motion cues, either the vehicle or the obstacle obviously needs to move. This requirement is a drawback of motion cue approaches because it would be both safer and more efficient for MAVs that have hovering capability to detect the obstacles underneath the vehicle in hover. This would require MAVs to exploit currently unused \emph{appearance} cues - the information contained in still images. Human pilots are very able to identify potential landing sites in still images, by recognizing obstacles and flat landing areas in view. This is based on years of experience, to learn what obstacles look like.

\begin{figure*}[thpb]
	\centering
	\captionsetup{justification=centering}
%
%
\begin{psfrags}%
\psfragscanon%
\newcommand{\tsize}{0.7}
\newcommand{\tsizeb}{0.5}
%
\psfrag{O}[t][t][\tsize]{\color[rgb]{0.4,0.4,0.4}\setlength{\tabcolsep}{0pt}\begin{tabular}{c}Optical Flow\end{tabular}}%
\psfrag{L}[t][t][\tsize]{\color[rgb]{0.4,0.4,0.4}\setlength{\tabcolsep}{0pt}\begin{tabular}{c}Learning\end{tabular}}%
\psfrag{P}[t][t][\tsize]{\color[rgb]{0.4,0.4,0.4}\setlength{\tabcolsep}{0pt}\begin{tabular}{c}Appearance\end{tabular}}%
\psfrag{q}[r][r][\tsize]{\color[rgb]{0,0,0}\setlength{\tabcolsep}{0pt}\begin{tabular}{c}$\mathbf{q}_1$\end{tabular}}%
\psfrag{w}[t][t][\tsize]{\color[rgb]{0,0,0}\setlength{\tabcolsep}{0pt}\begin{tabular}{c}$\mathbf{q}_{n-1}$\end{tabular}}%
\psfrag{T}[t][t][\tsize]{\color[rgb]{0,0,0}\setlength{\tabcolsep}{0pt}\begin{tabular}{c}Texton\\distribution\end{tabular}}%
\psfrag{f}[t][t][\tsize]{\color[rgb]{0,0,0}\setlength{\tabcolsep}{0pt}\begin{tabular}{c}$f(\mathbf{q}_i)=\epsilon^*_i$\end{tabular}}%
\psfrag{x}[t][t][\tsize]{\color[rgb]{0,0,0}\setlength{\tabcolsep}{0pt}\begin{tabular}{c}x\end{tabular}}%
\psfrag{y}[t][t][\tsize]{\color[rgb]{0,0,0}\setlength{\tabcolsep}{0pt}\begin{tabular}{c}y\end{tabular}}%
\psfrag{i}[t][t][\tsize]{\color[rgb]{0,0,0}\setlength{\tabcolsep}{0pt}\begin{tabular}{c}$I_{1,2}$\end{tabular}}%
\psfrag{j}[t][t][\tsize]{\color[rgb]{0,0,0}\setlength{\tabcolsep}{0pt}\begin{tabular}{c}$I_{n-1,n}$\end{tabular}}%
\psfrag{k}[t][t][\tsize]{\color[rgb]{0,0,0}\setlength{\tabcolsep}{0pt}\begin{tabular}{c}$\epsilon^*_1$\end{tabular}}%
\psfrag{l}[t][t][\tsize]{\color[rgb]{0,0,0}\setlength{\tabcolsep}{0pt}\begin{tabular}{c}$\epsilon^*_{n}$\end{tabular}}%
\psfrag{m}[t][t][\tsize]{\color[rgb]{0,0,0}\setlength{\tabcolsep}{0pt}\begin{tabular}{c}$I_{1}$\end{tabular}}%
\psfrag{n}[t][t][\tsize]{\color[rgb]{0,0,0}\setlength{\tabcolsep}{0pt}\begin{tabular}{c}$I_{n}$\end{tabular}}%
\psfrag{r}[t][t][\tsize]{\color[rgb]{0,0,0}\setlength{\tabcolsep}{0pt}\begin{tabular}{c}$\hat{\epsilon}_i$\end{tabular}}%
\psfrag{s}[t][t][\tsize]{\color[rgb]{0,0,0}\setlength{\tabcolsep}{0pt}\begin{tabular}{c}$\epsilon^*_i$\end{tabular}}%
\psfrag{C}[t][t][\tsize]{\color[rgb]{0.4,0.4,0.4}\setlength{\tabcolsep}{0pt}\begin{tabular}{c}Control\end{tabular}}%
\psfrag{u}[t][t][\tsize]{\color[rgb]{0,0,0}\setlength{\tabcolsep}{0pt}\begin{tabular}{c}$\epsilon^*_i$\end{tabular}}%
\psfrag{v}[r][r][\tsize]{\color[rgb]{0,0,0}\setlength{\tabcolsep}{0pt}\begin{tabular}{c}$\mathbf{q}_i$\end{tabular}}%
%
\includegraphics[width=\textwidth]{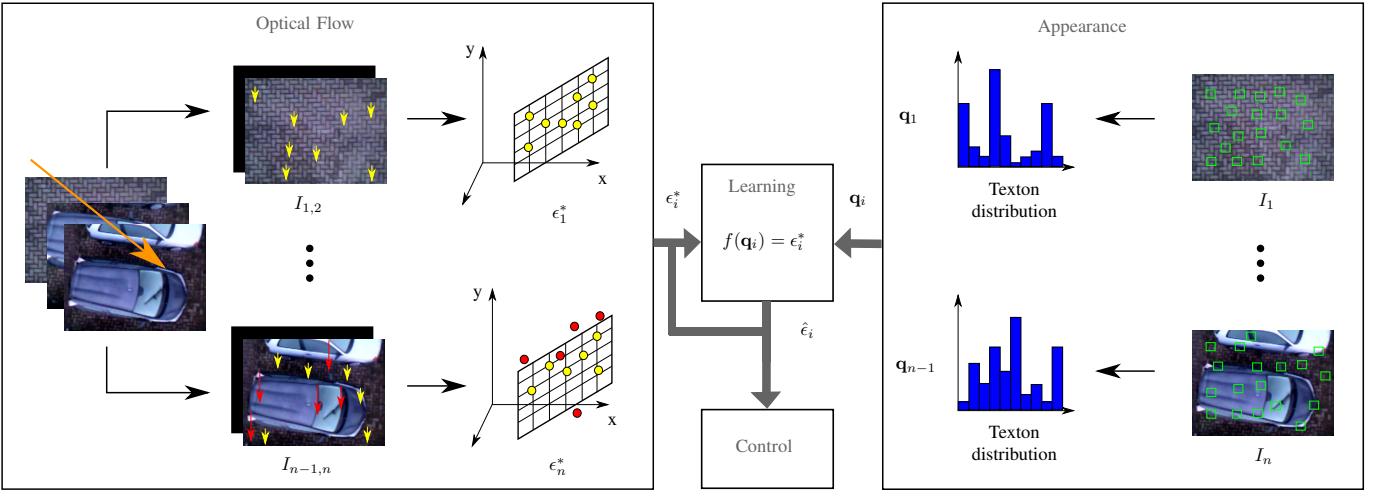}
\end{psfrags}%
%

	\caption{Overview of the novel self-supervised learning (SSL) setup. The MAV starts flying using a roughness measure $\epsilon^*$ extracted from optical flow algorithm (left). A function is learned that maps appearance features, $\mathbf{q}$ from a \emph{still} image to an estimate of the roughness measure $\hat{\epsilon}$ (right). After learning, the MAV can determine whether there are obstacles below based on a still image, allowing landing site selection in hover.}
	\label{fig:overview}
\end{figure*}

In this paper, we propose an approach that allows MAVs to learn about the appearance of obstacles and suitable landing sites completely by themselves. The approach involves a novel setup of Self-Supervised Learning (SSL), in which motion cues provide the targets for the supervised learning of a function that maps appearance features to a surface roughness measure (see Fig.~\ref{fig:overview} for an overview). We showed the feasibility of the proposed SSL concept in \cite{HannWoei2015OFSSL}. The current article takes into account the influence of MAV height on obstacle detection with optical flow and significantly extends upon \cite{HannWoei2015OFSSL} by means of a systematic analysis with experiments. This includes threshold selection based on height information and experiments in which the learned appearance is used in the control loop and a study of the generalization of the learned appearance to various indoor and outdoor environments.

The remainder of the article is set up as follows: In Section~\ref{sec:RelatedWork}, we discuss related work on self-supervised learning in more detail. In Section~\ref{sec:Method}, we describe our proposed concept to learn the visual appearance of obstacles based on \textit{surface roughness}, $\epsilon^*$ from optical flow. Section~\ref{sec:Results} presents the results of both optical flow- and appearance- based obstacle detection, and Section~\ref{sec:Generalization} explains generalization of our SSL approach to different environments. Then, Section~\ref{sec:FlightTests} demonstrates landing experiments where the proposed algorithms run onboard an MAV. 
Finally, we draw conclusions in Section~\ref{sec:Conclusion}.

\section{Related Work}
\label{sec:RelatedWork}
There are several remarkable achievements with SSL on autonomous driving cars, where stereo vision, laser scanners, or bumpers provided supervised outputs to learn the appearance of obstacles on the road \cite{dahlkamp2006self,thrun2006stanley,muller2013real,milella2015self}. In \cite{dahlkamp2006self,thrun2006stanley}, a close-range map is generated by the laser scanners to identify a nearby patch of drivable surface. This patch is used to train appearance models to extend the road detection range. In \cite{muller2013real}, terrain classification obtained from 3D information with a stereo camera is used for training a convolutional neural network (ConvNet). The input image patches to the trained ConvNet need to be normalized based on the estimated distance so that the obstacles always have the same size. Then, the trained model can be used to detect obstacles beyond the range of a stereo camera. In \cite{milella2015self}, ground and non-ground regions in images can be segmented using radar and monocular vision. This is done by serving the estimate of the ground region from radar as the supervised output to learn visual appearance of the ground. After learning, the unmanned ground robot can detect drivable surface using monocular vision. 

In \cite{lookingbill2007reverse}, optical flow was used for tracing back the obstacles in time when they are far away. The supervised outputs were still provided by stereo vision and bumpers. A weakly-supervised approach was used to segment drivable paths of a road vehicle \cite{barnes2017find}. This approach first labeled the training images by estimating the vehicle motion using a stereo camera and detecting 3D objects using a laser scanner and then used these labeled images to train a deep semantic segmentation network. The trained network can provide path segmentations using images from a single camera for autonomous driving of a road vehicle. Another drivable path segmentations were achieved by using a ConvNet \cite{wang2017self} or a fully convolutional network (FCN) \cite{Sanberg2017} which was trained using images labeled with a stereo camera. A nonlinear regression based depth estimation method using only a monocular camera was used for MAVs to learn a deliberate scheme for navigating through a cluttered environment \cite{Dey_2015_7909}. The training set was made using a stereo vision system on a ground robot and offboard image processing was done in the flight. 

A major difference of the approach we propose and previous work on SSL is that optical flow from monocular vision is used for generating the supervised outputs. To the best of our knowledge, there is only one other SSL study that also used optical flow to provide the supervised outputs. The study in \cite{lee2015online} used optical flow from a camera mounted on a car to learn a ground color model, assuming knowledge of the camera position relative to the ground. The learned ground color model aided in filtering optical flow vectors in order to improve the accuracy of the optical-flow based visual odometry. 

In this article, we use the optical flow from a downward-looking camera mounted on an MAV to learn the appearance of obstacles. In contrast to \cite{lee2015online}, we intend for the MAV to be able to use the learned appearance of obstacles even in the absence of supervisory cue of optical flow. This leads to a very interesting extension of the MAV's autonomous flight capabilities: while the robot initially only uses motion cues, and hence needs to move significantly in order to see obstacles, after learning it is able to see obstacles without moving. 

\section{Method}
\label{sec:Method}
An overview of the proposed method is shown in Fig.~\ref{fig:overview}. In the proposed setup of SSL, the MAV first uses optical flow to determine the landing surface roughness $\epsilon^*$. Simultaneously, the MAV extracts appearance features from each image, resulting in a feature vector $\mathbf{q}$. During operation, it uses the $\epsilon^*$ as targets for the supervised learning of $f(\mathbf{q})$, a function that maps appearance features to the surface roughness. After learning, the MAV can map $\mathbf{q}$ directly to an estimate of surface roughness $\hat{\epsilon}$. 

In this study, we focus on computationally efficient methods for the real world application to small MAVs which have limited computing capabilities. Therefore, we select a fast computer vision method and straightforward learning model which are suitable for real-time implementation. However, the proposed SSL setup does not preclude the use of other advanced computer vision and learning methods which could give even better results in terms of accuracy and generalization, at the cost of more computational effort. 

We explain our method in three subsections. First, in Subsection~\ref{subsec:OpticalFlowEquation} the optical flow algorithm to estimate $\epsilon^*$ is introduced. Second, in Subsection~\ref{subsec:AppearanceTexton} the texton method to represent appearance with $\mathbf{q}$ is explained. Third, in Subsection~\ref{subsec:RegressionLearning} the learning of the function $\mathbf{f(q)}$ is described. Finally, in Subsection~\ref{subsec:RoughnessThresholdSelectionfromHeight} the selection of roughness threshold is presented.

\subsection{Surface Roughness from Optical Flow}
\label{subsec:OpticalFlowEquation}
\begin{figure}[thpb]
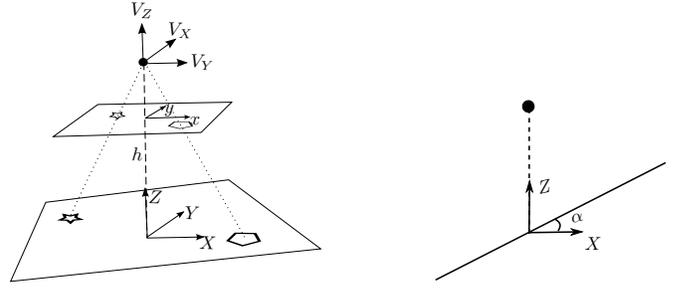

	\centering
	\captionsetup{justification=centering}
	\input{camera_model.tex} \hspace{1cm} \input{slope_model.tex}
	\caption{Left: A pin hole model. Right: An inclined ground surface with slope $\alpha$.}
	\label{fig:camera_model}
\end{figure}

In this section, a computationally efficient method is proposed to extract information from the optical flow field, which will allow the MAV to detect obstacles and determine if a landing spot is safe. The optical flow algorithm used to determine a safe landing spot is based on early findings in \cite{longuet1980interpretation}. The algorithm was developed in previous research \cite{de2013optic} by assuming that (a) a pinhole camera model pointing downward is used, (b) the surface in sight is planar, and (c) the angular rates of the camera can be measured and used to de-rotate the optical flow. Under these assumptions, the equation of the optical flow vectors can be expressed as follows:
\begin{equation}
u = -\vartheta_x + (\vartheta_x a + \vartheta_z) x + \vartheta_x b y - a \vartheta_z x^2 - b \vartheta_z xy,
\label{equation:optic_flow_u}
\end{equation}
\begin{equation}
v = -\vartheta_y + \vartheta_y a x + (\vartheta_y b + \vartheta_z) y - b \vartheta_z y^2 - a \vartheta_z xy,
\label{equation:optic_flow_v}
\end{equation}

\noindent where $u$ and $v$ are the optical flow vectors in $x$ and $y$ image coordinates system, respectively (see Fig.~\ref{fig:camera_model}). $\vartheta_x = V_x/h$, $\vartheta_y = V_y/h$, and $\vartheta_z = V_z/h$ are the corresponding velocities in $X$, $Y$, and $Z$ directions scaled with respect to the height $h$. 
Slope angles of the surface, $\alpha$ and $\beta$ are the arctangent of $a$ and $b$, respectively in Eqs.~(\ref{equation:optic_flow_u}), (\ref{equation:optic_flow_v}). In this work, we compute optical flow using
the sparse corner detection method with FAST \cite{rosten2006machine,rosten2005fusing} and Lucas-Kanade tracker \cite{bouquet2000pyramidal} to reduce computation for on-board processing. Note that since this proposed concept does not constrain the way to compute optical flow, other methods computing optical flow, e.g. dense optical flow can also be used.

By re-writing Eqs.~(\ref{equation:optic_flow_u}) and (\ref{equation:optic_flow_v}) into matrix form as shown below, the parameter vectors $\mathbf{p_u} = [p_{u1},p_{u2},p_{u3},p_{u4},p_{u5}]$ and $\mathbf{p_v} = [p_{v1},p_{v2},p_{v3},p_{v4},p_{v5}]$ can be estimated using a maximal-likelihood linear least-squares estimate within a robust random sample consensus (RANSAC) estimation procedure \cite{fischler1981random}:
\begin{equation}
u = \mathbf{p_u} [1, x, y, x^2, xy]^T,
\label{equation:pu}
\end{equation}
\begin{equation}
v = \mathbf{p_v} [1, x, y, y^2, xy]^T.
\label{equation:pv}
\end{equation}

\noindent These fits using RANSAC returns the number of inliers and the fitting error. If there are obstacles sticking out of the landing surface, their optical flow vectors will not fit with the second assumption of a planar landing surface. This leads to a higher fitting error, $\epsilon^*$ which can thus be interpreted as a measure of average \textit{surface roughness} of the entire area in the field of view in Eq.~(\ref{equation:3D}):

\begin{equation}
\epsilon^* = \epsilon_{u} + \epsilon_{v},
\label{equation:3D}
\end{equation}
\noindent with $\epsilon_{u}$ and $\epsilon_{v}$ sum of absolute errors of the RANSAC estimation in Eqs.~(\ref{equation:pu}) and (\ref{equation:pv}) divided by the number of tracked corners. Thus, we can use $\epsilon^*$ to detect obstacles near the ground surface by fitting the optical flow field. Note that Eqs.~(\ref{equation:optic_flow_u}) and (\ref{equation:optic_flow_v}) can be simplified by neglecting the second-order terms if the MAV only moves laterally. Therefore, a linear fit of the optical flow field can be used.

\subsection{Appearance from Texton Distribution}
\label{subsec:AppearanceTexton}
\begin{figure}[thpb]
	\centering
	\captionsetup{justification=centering}
%
%
\begin{psfrags}%
\psfragscanon%
\newcommand{\tsize}{0.7}
\newcommand{\tsizeb}{0.65}
%
\psfrag{a}[t][t][\tsize]{\color[rgb]{0,0,0}\setlength{\tabcolsep}{0pt}\begin{tabular}{c}Image\\Patch\end{tabular}}%
\psfrag{b}[t][t][\tsize]{\color[rgb]{0,0,0}\setlength{\tabcolsep}{0pt}\begin{tabular}{c}Dictionary\end{tabular}}%
\psfrag{c}[t][t][\tsize]{\color[rgb]{0,0,0}\setlength{\tabcolsep}{0pt}\begin{tabular}{c}Image\end{tabular}}%
\psfrag{d}[t][t][\tsize]{\color[rgb]{0,0,0}\setlength{\tabcolsep}{0pt}\begin{tabular}{c}Textons\end{tabular}}%
\psfrag{e}[t][t][\tsize]{\color[rgb]{0,0,0}\setlength{\tabcolsep}{0pt}\begin{tabular}{c}Probability\\Distribution\end{tabular}}%
%
\includegraphics[trim = 7mm 0mm 5mm 10mm, clip, width=0.5\textwidth]{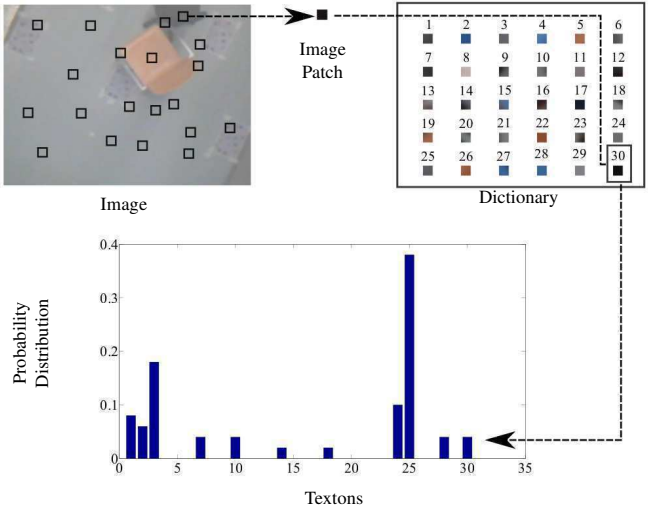}
\end{psfrags}%
%

	\caption{Texton Method: A number of image patches is randomly selected from an image. The patches are then compared to textons in a dictionary to form a texton probability distribution of the image.}
	\label{fig:textonDescription}
\end{figure}

In this study the visual appearance is described using the \textit{texton} method \cite{varma2003texture}, based on the extraction of small image patches (see Fig.~\ref{fig:textonDescription}). The advantage of using texton method is that it not only encodes the color distribution of an image, but also their textures. This improves the accuracy of the appearance representation while maintaining its efficiency for real-time application. 

With this method, first a \emph{dictionary} is created consisting of \emph{textons}, i.e., the cluster centroids of small image patches. For our implementation, we follow our previous work in \cite{de2012appearance}, and learn the dictionary with Kohonen clustering \cite{kohonen1990self}. After creation of the dictionary, image appearance can be represented as a texton distribution. To this end, a number of image patches are randomly extracted from an image and per patch the closest texton can be added to a corresponding bin in a histogram. By normalizing it with the number of patches, a maximum likelihood estimate of the texton probability distribution $\mathbf{q}$ is obtained. The texton method showed competitive results on texture classification tasks \cite{varma2003texture} with respect to computationally much more complex methods such as Gabor filter banks. In previous research, the texton method has been used to calculate the appearance variation cue \cite{de2009design} and to learn how to recognize heights and obstacles \cite{de2012appearance} but it was never applied to SSL.

\subsection{Regression Learning of Surface Roughness}
\label{subsec:RegressionLearning}
In order to learn the visual appearance of obstacles on the landing surface, a regression model is learned that maps a texton distribution to a surface roughness value. This function is learned by using the optical flow based roughness estimate $\epsilon^*$ as the regressand. In this subsection, we choose one of the regression methods for SSL based on the preliminary test results. 

To show feasibility and reliability of the relationship, various regression methods (such as Linear, Ridge, LASSO, Kernel smoother, Pseudo-inverse, Partial least squares, k-nearest neighbor regressions) were tried out to perform the learning using \textit{prtools} \cite{duin2000matlab} in MATLAB. Preliminary tests have shown little difference between the learning methods. For instance, the normalized Root Mean Square Errors (NRMSE) computed using Eq.~(\ref{equation:NRMSE}) on the test sets are $\sim10\%$ for the learning methods.

\begin{equation}
\mathit{NRMSE} = \frac{1}{\epsilon^*_{max}-\epsilon^*_{min}}\sqrt{\frac{\sum_{t=1}^n (\hat{\epsilon_t} - \epsilon^*_t)^2}{n}},
\label{equation:NRMSE}
\end{equation}

Since they all give reasonably good results, the linear regression method is used for this study, due to its simplicity and computational efficiency. After obtaining the texton distributions $\mathbf{q}$ of $m$ number of visual words and the roughness $\epsilon^*$ for $n$ images, a linear regression model expressed in Eq.~(\ref{equation:linearRegression}) can be trained.  
\begin{equation}
f(q_i)=\rho_1q_{i1}+...+\rho_mq_{im}+\Lambda,\quad i=1,...,n
\label{equation:linearRegression}
\end{equation}
\noindent where $\Lambda$ is a bias and $\mathbf{\rho}$ are the regression coefficients, which are optimized so that $f(\mathbf{q}) \approx \epsilon^*$.

\subsection{Roughness Threshold Selection for obstacle detection}
\label{subsec:RoughnessThresholdSelectionfromHeight}
One issue with our previous SSL work \cite{HannWoei2015OFSSL} is that it does not take the height of the MAV into account. In fact, the optical flow measurements are dependent on the MAV's velocity and the distance to the features. When the MAV is moving laterally at a constant speed, we measure smaller optical flow at larger heights than at lower heights. 
As a consequence, something that seems like an obstacle at a low height may not be detected as an obstacle at a large height. 
Therefore, we have to adapt a threshold for the roughness to the height and velocity for obstacle detection.

\begin{figure}[thpb]
	\centering
	\captionsetup{justification=centering}
	\begin{psfrags}%
\psfragscanon%
\newcommand{\tsize}{0.7}
%
\psfrag{F}[b][b][\tsize]{\color[rgb]{0,0,0}\setlength{\tabcolsep}{0pt}\begin{tabular}{c}Focal point\end{tabular}}%
\psfrag{i}[b][b][\tsize]{\color[rgb]{0,0,0}\setlength{\tabcolsep}{0pt}\begin{tabular}{c}Image plane\end{tabular}}%
\psfrag{g}[b][b][\tsize]{\color[rgb]{0,0,0}\setlength{\tabcolsep}{0pt}\begin{tabular}{c}Ground\end{tabular}}%
\psfrag{P}[b][b][\tsize]{\color[rgb]{0,0,0}\setlength{\tabcolsep}{0pt}\begin{tabular}{c}Features\end{tabular}}%
\psfrag{f}[b][b][\tsize]{\color[rgb]{0,0,0}\setlength{\tabcolsep}{0pt}\begin{tabular}{c}$\tilde{f}$\end{tabular}}%
\psfrag{d}[c][c][\tsize]{\color[rgb]{0,0,0}\setlength{\tabcolsep}{0pt}\begin{tabular}{c}$l$\end{tabular}}%
\psfrag{s}[t][t][\tsize]{\color[rgb]{0,0,0}\setlength{\tabcolsep}{0pt}\begin{tabular}{c}$V_X$\end{tabular}}%
\psfrag{Z}[b][b][\tsize]{\color[rgb]{0,0,0}\setlength{\tabcolsep}{0pt}\begin{tabular}{c}$h$\end{tabular}}%
\psfrag{m}[b][b][\tsize]{\color[rgb]{0,0,0}\setlength{\tabcolsep}{0pt}\begin{tabular}{c}$\Delta h$\end{tabular}}%
\psfrag{o}[b][b][\tsize]{\color[rgb]{0,0,0}\setlength{\tabcolsep}{0pt}\begin{tabular}{c}Object\end{tabular}}%

\psfrag{z}[t][t][\tsize]{\color[rgb]{0,0,0}\setlength{\tabcolsep}{0pt}\begin{tabular}{c}$h$\end{tabular}}%
\psfrag{h}[t][t][\tsize]{\color[rgb]{0,0,0}\setlength{\tabcolsep}{0pt}\begin{tabular}{c}$h-\Delta h$\end{tabular}}%
\psfrag{x}[t][t][\tsize]{\color[rgb]{0,0,0}\setlength{\tabcolsep}{0pt}\begin{tabular}{c}$u_x$\end{tabular}}%
\psfrag{y}[t][t][\tsize]{\color[rgb]{1,0,0}\setlength{\tabcolsep}{0pt}\begin{tabular}{c}$u_x+\Delta u_x$\end{tabular}}%

\includegraphics[width=0.45\textwidth]{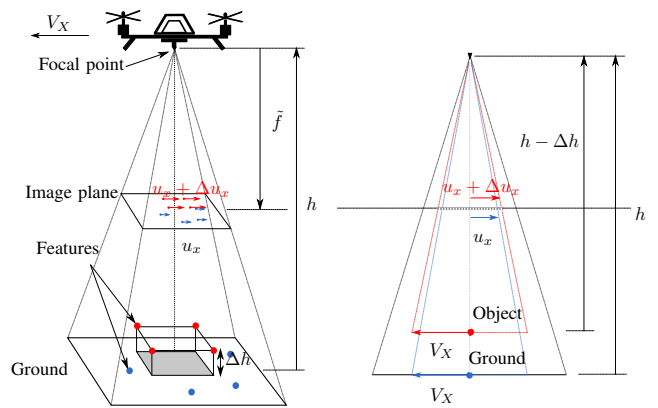}
\end{psfrags}%
%

	\caption{Estimation of roughness threshold for obstacle detection using surface roughness. \textit{Left}: The MAV moves laterally with a velocity of $V_X$ at the height of $h$. A downward-looking camera of a focal length of $\tilde{f}$ is used to observe de-rotated optical flow vectors on the ground $u_x$ and on the 3D object $u_x + \Delta u_x$. \textit{Right}: The features closer to the camera give larger flow vectors and result in an error $\Delta u_x$ in the flow field fitting. This error can be used as the roughness threshold for obstacle detection.} 
	\label{fig:threshold_selection_height_equation}
\end{figure}

In order to estimate this threshold, we use an example shown in Fig.~\ref{fig:threshold_selection_height_equation} in which an MAV moves laterally at a velocity of $V_X$ at the height of $h$. An onboard camera with a focal length of $\tilde{f}$ points vertically down to measure optical flow. The optical flow vectors, as mentioned in Subsection~\ref{subsec:OpticalFlowEquation}, are de-rotated using the angular rates measured from the IMU. Thus, only the translational components of the flow are used. By using the triangulation and assuming that there are features located in the center of view, optical flow vectors perceived from these features on the ground as a result of this movement can be written as:
\begin{equation}
u_x = \frac{V_X}{h}\tilde{f}.
\label{equation:OFVector}
\end{equation}

Since the features on a 3D object with a height of $\Delta h$ are closer to the camera, their optical flow vectors are larger, e.g., $u_x + \Delta u_x$. In the optical flow algorithm presented in Subsection \ref{subsec:OpticalFlowEquation}, the surface in sight is assumed to be planar. Therefore, $\Delta u_x$ can be considered as the outliers' error of optical flow field fitting and can be estimated as follow:
\begin{equation}
\begin{split}
\Delta u_x
=& V_X \tilde{f} \left[\frac{1}{h - \Delta h}-\frac{1}{h}\right] \\
=& V_X \tilde{f} \left[\frac{\Delta h}{h\left(h - \Delta h\right)}\right].
\end{split} 
\label{equation:roughnessThresholdEst}
\end{equation}

\begin{figure}[thpb]
	\centering
	\captionsetup{justification=centering}
%
%
\begin{psfrags}%
\psfragscanon%
\newcommand{\tsize}{1}
\newcommand{\tsizeb}{0.6}
%
\psfrag{e}[b][b][\tsize]{\color[rgb]{0.15,0.15,0.15}\setlength{\tabcolsep}{0pt}\begin{tabular}{c}$\epsilon^*$, $\hat{\epsilon}$, $\Delta u_x$\end{tabular}}%
%
\color[rgb]{0.15,0.15,0.15}%
%
\psfrag{2m}[t][t][\tsize]{2m}%
\psfrag{3m}[t][t][\tsize]{3m}%
\psfrag{4m}[t][t][\tsize]{4m}%
%
\psfrag{0}[r][r][\tsize]{0}%
\psfrag{2}[r][r][\tsize]{2}%
\psfrag{4}[r][r][\tsize]{4}%
\psfrag{6}[r][r][\tsize]{6}%
\psfrag{8}[r][r][\tsize]{8}%
\psfrag{10}[r][r][\tsize]{10}%
\psfrag{12}[r][r][\tsize]{12}%
%
\psfrag{A}[r][r][\tsizeb]{$\Delta u_x$ for $\Delta h = 0.01 m$}%
\psfrag{B}[r][r][\tsizeb]{$\Delta u_x$ for $\Delta h = 0.03 m$}%
\psfrag{C}[r][r][\tsizeb]{$\Delta u_x$ for $\Delta h = 0.1 m$}%
\psfrag{X}[r][r][\tsize]{$2~m$}%
\psfrag{Y}[r][r][\tsize]{$3~m$}%
\psfrag{Z}[r][r][\tsize]{$4~m$}%
\includegraphics[trim = 10mm 0mm 20mm 0mm, clip, width=0.5\textwidth]{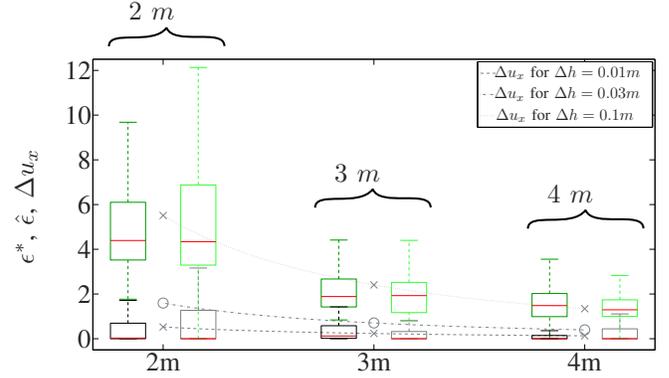}
\end{psfrags}%
%

	\caption{Effect of height on roughness. Dark green and light green boxplots represent the roughness $\epsilon^*$ and $\hat{\epsilon}$, respectively measured in the presence of obstacles. Black and gray boxplots represent the roughness $\epsilon^*$ and $\hat{\epsilon}$, respectively measured in the absence of obstacles. A dashed line, a dash-dot line, and a dotted line are plotted to show $\Delta u_x$ at $V_X=0.6~m/s$ for $\Delta h = 0.01~m$, $0.03~m$, and $0.1~m$, respectively. The dash-dot line represents the threshold line to detect obstacles.}
	\label{fig:FlatnessHeights}
\end{figure}

In Eq.~\ref{equation:roughnessThresholdEst}, $\tilde{f}$ can be found from the specification of the camera, $h$ can be measured from an onboard sonar sensor, and then $V_X$ is also known. By choosing the allowable height $\Delta h$ of the object which we would like to detect, $\Delta u_x$ can be computed. We know from Eq.~\ref{equation:3D} that the roughness estimate $\epsilon^*$ is the average error of all flow vectors including both inliers and outliers. Thus, $\epsilon^*$ obtained in the presence of an obstacle can be smaller than $\Delta u_x$ at the same height when averaging all the flow fitting errors (see Fig.~\ref{fig:FlatnessHeights}). 

\begin{figure}[thpb]
	\centering
	\captionsetup{justification=centering}
%
%
\begin{psfrags}%
\psfragscanon%
\newcommand{\tsize}{0.7}
\newcommand{\tsizeb}{0.5}
%
\psfrag{s14}[t][t]{\color[rgb]{0,0,0}\setlength{\tabcolsep}{0pt}\begin{tabular}{c}$h$~(m)\end{tabular}}%
\psfrag{s15}[b][b]{\color[rgb]{0,0,0}\setlength{\tabcolsep}{0pt}\begin{tabular}{c}$\Delta u_x$\end{tabular}}%
\psfrag{s18}[][]{\color[rgb]{0,0,0}\setlength{\tabcolsep}{0pt}\begin{tabular}{c} \end{tabular}}%
\psfrag{s19}[][]{\color[rgb]{0,0,0}\setlength{\tabcolsep}{0pt}\begin{tabular}{c} \end{tabular}}%
\psfrag{s20}[l][l][\tsizeb]{\color[rgb]{0,0,0}$2.0~m/s$}%
\psfrag{s21}[l][l][\tsizeb]{\color[rgb]{0,0,0}$0.2~m/s$}%
\psfrag{s22}[l][l][\tsizeb]{\color[rgb]{0,0,0}$0.4~m/s$}%
\psfrag{s23}[l][l][\tsizeb]{\color[rgb]{0,0,0}$0.6~m/s$}%
\psfrag{s24}[l][l][\tsizeb]{\color[rgb]{0,0,0}$0.8~m/s$}%
\psfrag{s25}[l][l][\tsizeb]{\color[rgb]{0,0,0}$1.0~m/s$}%
\psfrag{s26}[l][l][\tsizeb]{\color[rgb]{0,0,0}$1.2~m/s$}%
\psfrag{s27}[l][l][\tsizeb]{\color[rgb]{0,0,0}$1.4~m/s$}%
\psfrag{s28}[l][l][\tsizeb]{\color[rgb]{0,0,0}$1.6~m/s$}%
\psfrag{s29}[l][l][\tsizeb]{\color[rgb]{0,0,0}$1.8~m/s$}%
\psfrag{s30}[l][l][\tsizeb]{\color[rgb]{0,0,0}$2.0~m/s$}%
%
\psfrag{x01}[t][t][\tsize]{1}%
\psfrag{x02}[t][t][\tsize]{1.5}%
\psfrag{x03}[t][t][\tsize]{2}%
\psfrag{x04}[t][t][\tsize]{2.5}%
\psfrag{x05}[t][t][\tsize]{3}%
\psfrag{x06}[t][t][\tsize]{3.5}%
\psfrag{x07}[t][t][\tsize]{4}%
\psfrag{x08}[t][t][\tsize]{10}%
\psfrag{x09}[t][t][\tsize]{}%
\psfrag{x10}[t][t][\tsize]{11}%
\psfrag{x11}[t][t][\tsize]{}%
\psfrag{x12}[t][t][\tsize]{12}%
\psfrag{x13}[t][t][\tsize]{}%
\psfrag{x14}[t][t][\tsize]{13}%
\psfrag{x15}[t][t][\tsize]{}%
\psfrag{x16}[t][t][\tsize]{14}%
\psfrag{x17}[t][t][\tsize]{}%
\psfrag{x18}[t][t][\tsize]{15}%
\psfrag{x19}[t][t]{0}%
\psfrag{x20}[t][t]{2}%
\psfrag{x21}[t][t]{4}%
\psfrag{x22}[t][t]{6}%
\psfrag{x23}[t][t]{8}%
\psfrag{x24}[t][t]{10}%
\psfrag{x25}[t][t]{12}%
\psfrag{x26}[t][t]{14}%
\psfrag{x27}[t][t]{16}%
%
\psfrag{v01}[r][r][\tsize]{5}%
\psfrag{v02}[r][r][\tsize]{10}%
\psfrag{v03}[r][r][\tsize]{15}%
\psfrag{v04}[r][r][\tsize]{20}%
\psfrag{v05}[r][r][\tsize]{25}%
\psfrag{v06}[r][r][\tsize]{0.05}%
\psfrag{v07}[r][r][\tsize]{0.1}%
\psfrag{v08}[r][r][\tsize]{0.15}%
\psfrag{v09}[r][r][\tsize]{0.2}%
\psfrag{v10}[r][r][\tsize]{}%
\psfrag{v11}[r][r]{0}%
\psfrag{v12}[r][r]{10}%
\psfrag{v13}[r][r]{20}%
\psfrag{v14}[r][r]{30}%
\psfrag{v15}[r][r]{40}%
\psfrag{v16}[r][r]{50}%
\psfrag{v17}[r][r]{60}%
\psfrag{v18}[r][r]{70}%
\psfrag{v19}[r][r]{80}%
\psfrag{v20}[r][r]{90}%
%
\includegraphics[trim = 0mm 0mm 0mm 0mm, clip, width=0.48\textwidth]{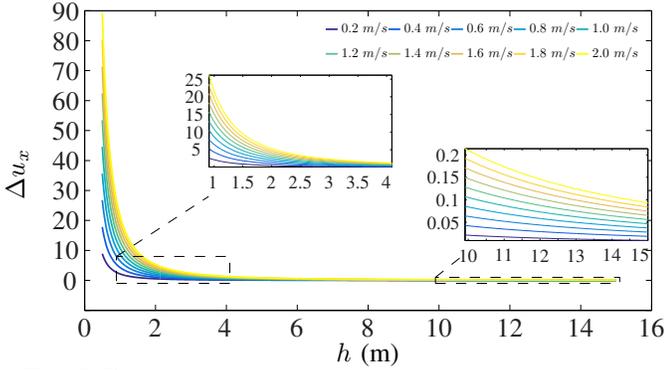}
\end{psfrags}%
%

	\caption{Roughness threshold selection for obstacle detection ($\Delta h = 0.03~m$). Two enlarged views are used to select the roughness threshold for the indoor (left) and outdoor (right) flights which are presented in this paper.}
	\label{fig:threshold_selection_height}
\end{figure}

Fig.~\ref{fig:threshold_selection_height} plots $\Delta u_x$ against $h$ ($0.5-15~m$) at different $V_X$ ($0.2-2.0~m/s$) for $\Delta h$ of $0.03~m$. 
This value of $\Delta h$ is selected because we would like to detect obstacles above this height which can endanger the MAVs. From this figure, we can choose a roughness threshold $\epsilon^*_{th},\widehat{\epsilon}_{th} \approx \Delta u_x$ based on the known $h$ and $V_X$. When the roughness estimate is larger than the threshold, it means that the surface in sight contains obstacle(s). Two enlarged views in this figure present $\Delta u_x$ for the range of the heights in which the MAV flew in indoor (left) and outdoor (right) environments for the experiments presented in this paper.

In order to validate the feasibility of the selected roughness threshold used for obstacle detection, we conducted experiments flying the vehicle over an obstacle (e.g., a chair with the height of $\approx 0.9~m$) at various heights ($2~m$, $3~m$, and $4~m$) and the same velocity ($\approx0.6~m/s$) in the same structured indoor environment. Fig.~\ref{fig:FlatnessHeights} shows surface roughness estimated from the optical flow algorithm and SSL method at different heights, in which the dark and light green boxplots are the roughness estimates in the presence of obstacles while the black and gray boxplots are the roughness estimates in the absence of obstacles. These boxplots show the influence of height on the roughness estimates. 

In this figure, a dashed line, a dash-dot line, and a dotted line are plotted to show $\Delta u_x$ at $V_X=0.6~m/s$ for $\Delta h = 0.01~m$, $0.03~m$, and $0.1~m$, respectively. We can observe from the boxplots and these lines that the roughness estimates in the absence of obstacles are resulted from the inliers' errors, which correspond to $\Delta u_x$ at $\Delta h = 0.01~m$. While in the presence of obstacles, these inliers' errors cause the roughness estimates much smaller after averaging all the flow fitting errors. For example, the roughness estimates in the presence of the chair with the height of $\approx 0.9~m$ correspond to the $\Delta u_x$ at $\Delta h = 0.1~m$. 

However, most importantly, if we choose the $\Delta h$ which is slightly larger than $\Delta h$ which causes inliers' errors, we can select a suitable threshold for obstacle detection using surface roughness. For example, a dash-dot line is also plotted in this figure to represent $\Delta u_x$ at $V_X=0.6~m/s$ for $\Delta h = 0.03~m$ (slightly larger than $0.01~m$). It is clear that, if the roughness estimates are greater than this $\Delta u_x$ (same as the threshold plotted in Fig.~\ref{fig:threshold_selection_height}), the obstacles can be detected in most cases. From this result, we show that, by using the roughness threshold selected based on Eq.~\ref{equation:roughnessThresholdEst} or Fig.~\ref{fig:threshold_selection_height}, the MAV can automatically detect obstacles using surface roughness. 

Additionally, from Figs.~\ref{fig:threshold_selection_height} and \ref{fig:FlatnessHeights}, the roughness estimates resulting from the obstacles, which have the same heights over $2~m,3~m,4~m$ in the experiments, get closer and closer to that of having no obstacles as the MAV flies higher. Thus, we know that small obstacles cannot be well observed with optical flow at large heights, and these obstacles are also much smaller in the image and so are much harder to learn. In contrast, higher obstacles typically are also larger, so that they can be spotted at larger heights. This corresponds well to our intuition that at larger heights an MAV should avoid to land, e.g., on a roof of a building, while when closer to the landing surface it prefers to touch down without being tilted/ flipped over.

\section{Obstacle Detection}
\label{sec:Results}
\begin{figure}[thpb]
	\captionsetup[subfigure]{justification=centering}
	\centering
	\begin{subfigure}[thpb]{0.45\textwidth}
%
%
\begin{psfrags}%
\psfragscanon%
\newcommand{\tsize}{0.6}
\newcommand{\tsizeb}{0.5}
%
\psfrag{s28}[t][t][\tsize]{\color[rgb]{0,0,0}\setlength{\tabcolsep}{0pt}\begin{tabular}{c}S1\end{tabular}}%
\psfrag{s29}[t][t][\tsize]{\color[rgb]{0,0,0}\setlength{\tabcolsep}{0pt}\begin{tabular}{c}S2\end{tabular}}%
\psfrag{s30}[t][t][\tsize]{\color[rgb]{0,0,0}\setlength{\tabcolsep}{0pt}\begin{tabular}{c}S3\end{tabular}}%
\psfrag{s31}[t][t][\tsize]{\color[rgb]{0,0,0}\setlength{\tabcolsep}{0pt}\begin{tabular}{c}S4\end{tabular}}%
\psfrag{s32}[t][t][\tsize]{\color[rgb]{0,0,0}\setlength{\tabcolsep}{0pt}\begin{tabular}{c}S5\end{tabular}}%
\psfrag{s33}[t][t][\tsize]{\color[rgb]{0,0,0}\setlength{\tabcolsep}{0pt}\begin{tabular}{c}S6\end{tabular}}%
\psfrag{s34}[t][t][\tsize]{\color[rgb]{0,0,0}\setlength{\tabcolsep}{0pt}\begin{tabular}{c}S7\end{tabular}}%
\psfrag{s35}[t][t][\tsize]{\color[rgb]{0,0,0}\setlength{\tabcolsep}{0pt}\begin{tabular}{c}S8\end{tabular}}%
\psfrag{s36}[t][t][\tsize]{\color[rgb]{0,0,0}\setlength{\tabcolsep}{0pt}\begin{tabular}{c}S9\end{tabular}}%
%
\includegraphics[width=\textwidth]{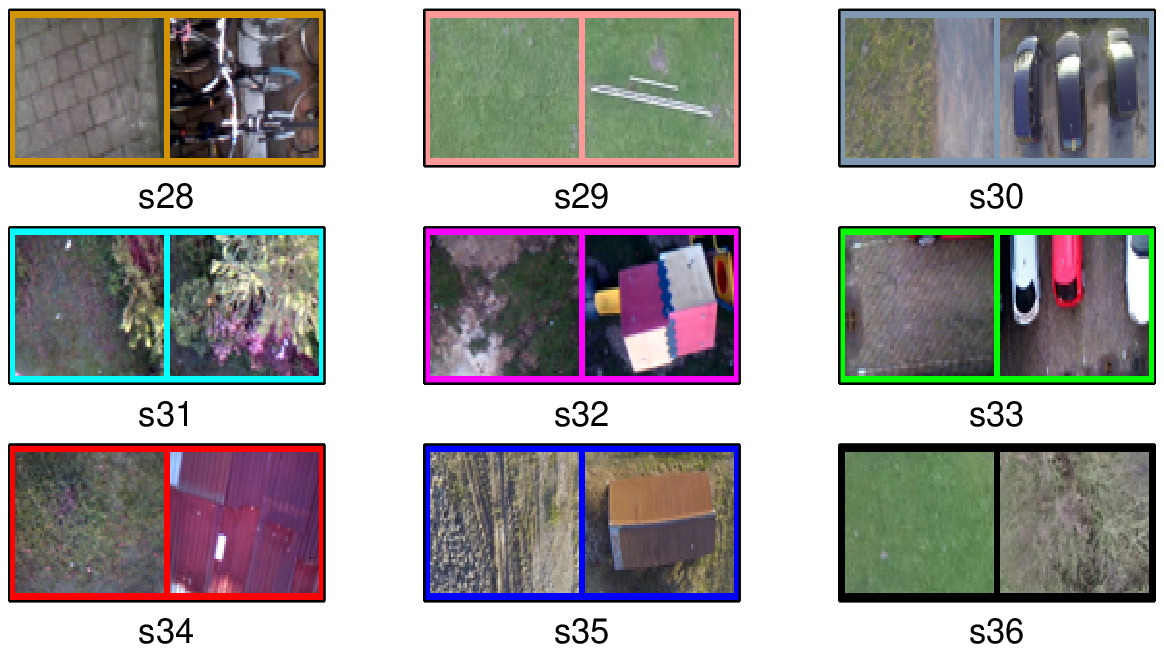}%
\end{psfrags}%
%

		\caption{Outdoor datasets}
		\label{fig:OutdoorDataset}
	\end{subfigure} \qquad
	\begin{subfigure}[thpb]{0.45\textwidth}
%
%
\begin{psfrags}%
\psfragscanon%
\newcommand{\tsize}{0.6}
\newcommand{\tsizeb}{0.5}
%
\psfrag{s28}[t][t][\tsize]{\color[rgb]{0,0,0}\setlength{\tabcolsep}{0pt}\begin{tabular}{c}S1\end{tabular}}%
\psfrag{s29}[t][t][\tsize]{\color[rgb]{0,0,0}\setlength{\tabcolsep}{0pt}\begin{tabular}{c}S2\end{tabular}}%
\psfrag{s30}[t][t][\tsize]{\color[rgb]{0,0,0}\setlength{\tabcolsep}{0pt}\begin{tabular}{c}S3\end{tabular}}%
\psfrag{s31}[t][t][\tsize]{\color[rgb]{0,0,0}\setlength{\tabcolsep}{0pt}\begin{tabular}{c}S4\end{tabular}}%
\psfrag{s32}[t][t][\tsize]{\color[rgb]{0,0,0}\setlength{\tabcolsep}{0pt}\begin{tabular}{c}S5\end{tabular}}%
\psfrag{s33}[t][t][\tsize]{\color[rgb]{0,0,0}\setlength{\tabcolsep}{0pt}\begin{tabular}{c}S6\end{tabular}}%
\psfrag{s34}[t][t][\tsize]{\color[rgb]{0,0,0}\setlength{\tabcolsep}{0pt}\begin{tabular}{c}S7\end{tabular}}%
\psfrag{s35}[t][t][\tsize]{\color[rgb]{0,0,0}\setlength{\tabcolsep}{0pt}\begin{tabular}{c}S8\end{tabular}}%
\psfrag{s36}[t][t][\tsize]{\color[rgb]{0,0,0}\setlength{\tabcolsep}{0pt}\begin{tabular}{c}S9\end{tabular}}%
%
\includegraphics[width=\textwidth]{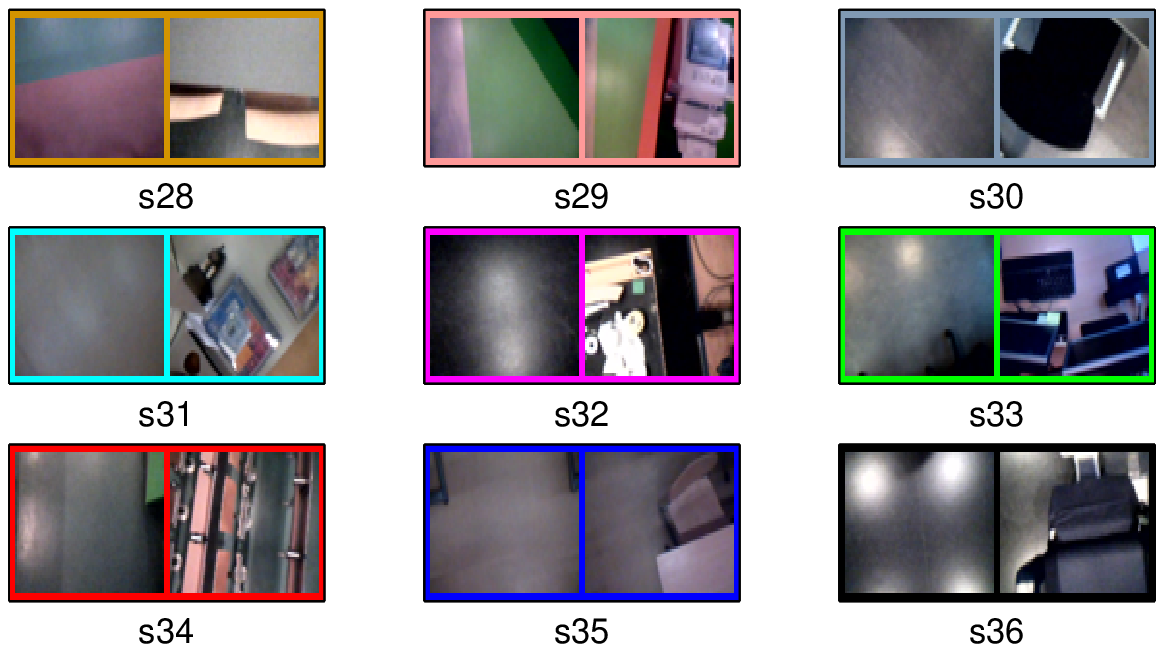}%
\end{psfrags}%
%

		\caption{Indoor datasets}
		\label{fig:IndoorDataset}
	\end{subfigure}
	\caption{Image datasets collected from outdoor and indoor environments (Scenes S1-S9). For each environment, an image is shown without obstacle (left) and with obstacle (right).}
	\label{fig:Dataset}
\end{figure}

In this section, we test how well the methods in Section~\ref{sec:Method} work to detect obstacles. To this end, we made multiple image sets of onboard MAV images from a downward-looking camera. Outdoors MAV flew at $\approx10\sim 15$ meters high in 9 different environments, ranging from a car parking lot to a park. Indoors MAV flew at $\approx1\sim 4$ meters high in 9 different indoor environments, ranging from the canteen to office spaces. Both the indoor and outdoor datasets show considerable variation in appearance of the landing surface and obstacles, and hence represent a significant challenge for machine learning methods. 
Fig.~\ref{fig:Dataset} shows the onboard images without (left) and with (right) obstacle of each environment. 

\subsection{Obstacle Detection with Surface Roughness}
\label{subsec:ObstacleDetectionRoughness}
\begin{figure}[thpb]
	\captionsetup[subfigure]{justification=centering}
	\centering
	\begin{subfigure}[thpb]{0.45\textwidth}
%
%
\begin{psfrags}%
\psfragscanon%
\newcommand{\tsize}{0.7}
\newcommand{\tsizeb}{0.65}
%
\psfrag{s01}[t][t][\tsize]{\color[rgb]{0,0,0}\setlength{\tabcolsep}{0pt}\begin{tabular}{c}x\end{tabular}}%
\psfrag{s02}[b][b][\tsize]{\color[rgb]{0,0,0}\setlength{\tabcolsep}{0pt}\begin{tabular}{c}y\end{tabular}}%
\psfrag{s03}[b][b][\tsize]{\color[rgb]{0,0,0}\setlength{\tabcolsep}{0pt}\begin{tabular}{c}\end{tabular}}%
\psfrag{s05}[l][l][\tsize]{\color[rgb]{0,0,0}\setlength{\tabcolsep}{0pt}\begin{tabular}{r}x\end{tabular}}%
\psfrag{s06}[r][r][\tsize]{\color[rgb]{0,0,0}\setlength{\tabcolsep}{0pt}\begin{tabular}{l}y\end{tabular}}%
\psfrag{s07}[t][t][\tsize]{\color[rgb]{0,0,0}\setlength{\tabcolsep}{0pt}\begin{tabular}{c}u\end{tabular}}%
\psfrag{s09}[l][l][\tsize]{\color[rgb]{0,0,0}\setlength{\tabcolsep}{0pt}\begin{tabular}{r}x\end{tabular}}%
\psfrag{s10}[r][r][\tsize]{\color[rgb]{0,0,0}\setlength{\tabcolsep}{0pt}\begin{tabular}{l}y\end{tabular}}%
\psfrag{s11}[t][t][\tsize]{\color[rgb]{0,0,0}\setlength{\tabcolsep}{0pt}\begin{tabular}{c}v\end{tabular}}%
%
\psfrag{x01}[t][t][\tsize]{}%
\psfrag{x02}[t][t][\tsize]{}%
\psfrag{x03}[t][t][\tsize]{}%
\psfrag{x04}[t][t][\tsize]{}%
\psfrag{x05}[t][t][\tsize]{}%
\psfrag{x06}[t][t][\tsize]{}%
\psfrag{x07}[t][t][\tsize]{}%
\psfrag{x08}[t][t][\tsize]{}%
\psfrag{x09}[t][t][\tsize]{}%
\psfrag{x10}[t][t][\tsize]{}%
\psfrag{x11}[t][t][\tsize]{}%
\psfrag{x12}[t][t][\tsize]{\scriptsize0}%
\psfrag{x13}[t][t][\tsize]{}%
\psfrag{x14}[t][t][\tsize]{\scriptsize300}%
\psfrag{x15}[t][t][\tsize]{\scriptsize0}%
\psfrag{x16}[t][t][\tsize]{}%
\psfrag{x17}[t][t][\tsize]{\scriptsize300}%
\psfrag{x18}[t][t][\tsize]{}%
\psfrag{x19}[t][t][\tsize]{100}%
\psfrag{x20}[t][t][\tsize]{}%
\psfrag{x21}[t][t][\tsize]{200}%
\psfrag{x22}[t][t][\tsize]{}%
\psfrag{x23}[t][t][\tsize]{300}%
%
\psfrag{v01}[r][r][\tsize]{}%
\psfrag{v02}[r][r][\tsize]{}%
\psfrag{v03}[r][r][\tsize]{}%
\psfrag{v04}[r][r][\tsize]{}%
\psfrag{v05}[r][r][\tsize]{}%
\psfrag{v06}[r][r][\tsize]{}%
\psfrag{v07}[r][r][\tsize]{}%
\psfrag{v08}[r][r][\tsize]{}%
\psfrag{v09}[r][r][\tsize]{}%
\psfrag{v10}[r][r][\tsize]{}%
\psfrag{v11}[r][r][\tsize]{}%
\psfrag{v12}[r][r][\tsize]{}%
\psfrag{v13}[r][r][\tsize]{\scriptsize40}%
\psfrag{v14}[r][r][\tsize]{}%
\psfrag{v15}[r][r][\tsize]{}%
\psfrag{v16}[r][r][\tsize]{}%
\psfrag{v17}[r][r][\tsize]{\scriptsize80}%
\psfrag{v18}[r][r][\tsize]{}%
\psfrag{v19}[r][r][\tsize]{}%
\psfrag{v20}[r][r][\tsize]{}%
\psfrag{v21}[r][r][\tsize]{}%
\psfrag{v22}[r][r][\tsize]{\scriptsize40}%
\psfrag{v23}[r][r][\tsize]{}%
\psfrag{v24}[r][r][\tsize]{}%
\psfrag{v25}[r][r][\tsize]{}%
\psfrag{v26}[r][r][\tsize]{\scriptsize80}%
\psfrag{v27}[r][r][\tsize]{}%
\psfrag{v28}[r][r][\tsize]{}%
\psfrag{v29}[r][r][\tsize]{}%
\psfrag{v30}[r][r][\tsize]{}%
\psfrag{v31}[r][r][\tsize]{100}%
\psfrag{v32}[r][r][\tsize]{}%
\psfrag{v33}[r][r][\tsize]{200}%
%
\psfrag{z01}[r][r][\tsize]{\scriptsize2}%
\psfrag{z02}[r][r][\tsize]{}%
\psfrag{z03}[r][r][\tsize]{\scriptsize4}%
\psfrag{z04}[r][r][\tsize]{\scriptsize-1}%
\psfrag{z05}[r][r][\tsize]{}%
\psfrag{z06}[r][r][\tsize]{\scriptsize0}%
\psfrag{z07}[r][r][\tsize]{}%
%
\includegraphics[width=\textwidth]{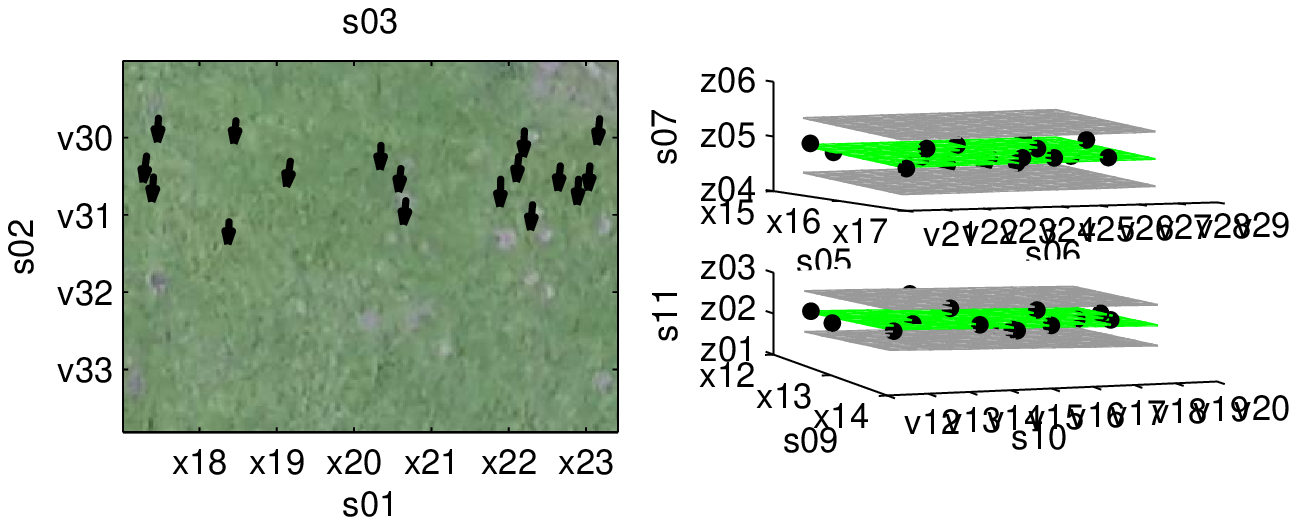}%
\end{psfrags}%
%

		\caption{Non-obstacle (grass field)}
		\label{fig:FittingGrassObject}
	\end{subfigure}
	\begin{subfigure}[thpb]{0.45\textwidth}
%
%
\begin{psfrags}%
\psfragscanon%
\newcommand{\tsize}{0.7}
\newcommand{\tsizeb}{0.65}
%
\psfrag{s04}[t][t][\tsize]{\color[rgb]{0,0,0}\setlength{\tabcolsep}{0pt}\begin{tabular}{c}x\end{tabular}}%
\psfrag{s05}[b][b][\tsize]{\color[rgb]{0,0,0}\setlength{\tabcolsep}{0pt}\begin{tabular}{c}y\end{tabular}}%
\psfrag{s06}[b][b][\tsize]{\color[rgb]{0,0,0}\setlength{\tabcolsep}{0pt}\begin{tabular}{c}\end{tabular}}%
\psfrag{s07}[lt][lt][\tsize]{\color[rgb]{0,0,0}\setlength{\tabcolsep}{0pt}\begin{tabular}{r}x\end{tabular}}%
\psfrag{s08}[rt][rt][\tsize]{\color[rgb]{0,0,0}\setlength{\tabcolsep}{0pt}\begin{tabular}{l}y\end{tabular}}%
\psfrag{s09}[t][t][\tsize]{\color[rgb]{0,0,0}\setlength{\tabcolsep}{0pt}\begin{tabular}{c}u\end{tabular}}%
\psfrag{s10}[lt][lt][\tsize]{\color[rgb]{0,0,0}\setlength{\tabcolsep}{0pt}\begin{tabular}{r}x\end{tabular}}%
\psfrag{s11}[rt][rt][\tsize]{\color[rgb]{0,0,0}\setlength{\tabcolsep}{0pt}\begin{tabular}{l}y\end{tabular}}%
\psfrag{s12}[t][t][\tsize]{\color[rgb]{0,0,0}\setlength{\tabcolsep}{0pt}\begin{tabular}{c}v\end{tabular}}%
%
\psfrag{x01}[t][t][\tsize]{\scriptsize100}%
\psfrag{x02}[t][t][\tsize]{}%
\psfrag{x03}[t][t][\tsize]{\scriptsize300}%
\psfrag{x04}[t][t][\tsize]{}%
\psfrag{x05}[t][t][\tsize]{\scriptsize100}%
\psfrag{x06}[t][t][\tsize]{}%
\psfrag{x07}[t][t][\tsize]{\scriptsize300}%
\psfrag{x08}[t][t][\tsize]{}%
\psfrag{x09}[t][t][\tsize]{}%
\psfrag{x10}[t][t][\tsize]{100}%
\psfrag{x11}[t][t][\tsize]{}%
\psfrag{x12}[t][t][\tsize]{200}%
\psfrag{x13}[t][t][\tsize]{}%
\psfrag{x14}[t][t][\tsize]{300}%
%
\psfrag{v01}[r][r][\tsize]{}%
\psfrag{v02}[r][r][\tsize]{}%
\psfrag{v03}[r][r][\tsize]{\scriptsize100}%
\psfrag{v04}[r][r][\tsize]{}%
\psfrag{v05}[r][r][\tsize]{}%
\psfrag{v06}[r][r][\tsize]{}%
\psfrag{v07}[r][r][\tsize]{}%
\psfrag{v08}[r][r][\tsize]{\scriptsize200}%
\psfrag{v09}[r][r][\tsize]{}%
\psfrag{v10}[r][r][\tsize]{}%
\psfrag{v11}[r][r][\tsize]{\scriptsize100}%
\psfrag{v12}[r][r][\tsize]{}%
\psfrag{v13}[r][r][\tsize]{}%
\psfrag{v14}[r][r][\tsize]{}%
\psfrag{v15}[r][r][\tsize]{}%
\psfrag{v16}[r][r][\tsize]{\scriptsize200}%
\psfrag{v17}[r][r][\tsize]{}%
\psfrag{v18}[r][r][\tsize]{100}%
\psfrag{v19}[r][r][\tsize]{}%
\psfrag{v20}[r][r][\tsize]{200}%
%
\psfrag{z01}[r][r][\tsize]{\scriptsize2}%
\psfrag{z02}[r][r][\tsize]{}%
\psfrag{z03}[r][r][\tsize]{\scriptsize3}%
\psfrag{z04}[r][r][\tsize]{}%
\psfrag{z05}[r][r][\tsize]{\scriptsize-2}%
\psfrag{z06}[r][r][\tsize]{}%
\psfrag{z07}[r][r][\tsize]{\scriptsize0}%
%
\includegraphics[width=\textwidth]{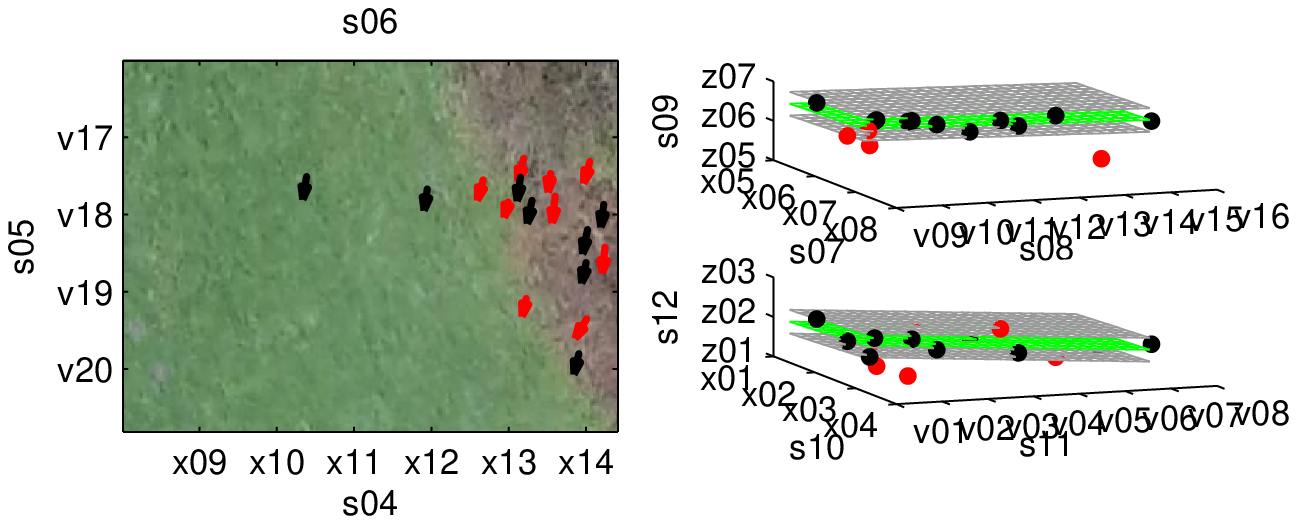}%
\end{psfrags}%
%

		\caption{Obstacle (trees)}
		\label{fig:FittingTrees}
	\end{subfigure}
	\begin{subfigure}[thpb]{0.45\textwidth}
%
%
\begin{psfrags}%
\psfragscanon%
\newcommand{\tsize}{0.7}
\newcommand{\tsizeb}{0.65}
%
\psfrag{s01}[t][t][\tsize]{\color[rgb]{0,0,0}\setlength{\tabcolsep}{0pt}\begin{tabular}{c}x\end{tabular}}%
\psfrag{s02}[b][b][\tsize]{\color[rgb]{0,0,0}\setlength{\tabcolsep}{0pt}\begin{tabular}{c}y\end{tabular}}%
\psfrag{s03}[b][b][\tsize]{\color[rgb]{0,0,0}\setlength{\tabcolsep}{0pt}\begin{tabular}{c}\end{tabular}}%
\psfrag{s05}[lt][lt][\tsize]{\color[rgb]{0,0,0}\setlength{\tabcolsep}{0pt}\begin{tabular}{c}x\end{tabular}}%
\psfrag{s06}[rt][rt][\tsize]{\color[rgb]{0,0,0}\setlength{\tabcolsep}{0pt}\begin{tabular}{c}y\end{tabular}}%
\psfrag{s07}[t][t][\tsize]{\color[rgb]{0,0,0}\setlength{\tabcolsep}{0pt}\begin{tabular}{c}u\end{tabular}}%
\psfrag{s09}[lt][lt][\tsize]{\color[rgb]{0,0,0}\setlength{\tabcolsep}{0pt}\begin{tabular}{r}x\end{tabular}}%
\psfrag{s10}[rt][rt][\tsize]{\color[rgb]{0,0,0}\setlength{\tabcolsep}{0pt}\begin{tabular}{l}y\end{tabular}}%
\psfrag{s11}[t][t][\tsize]{\color[rgb]{0,0,0}\setlength{\tabcolsep}{0pt}\begin{tabular}{c}v\end{tabular}}%
%
\psfrag{x01}[t][t][\tsize]{}%
\psfrag{x02}[t][t][\tsize]{}%
\psfrag{x03}[t][t][\tsize]{}%
\psfrag{x04}[t][t][\tsize]{}%
\psfrag{x05}[t][t][\tsize]{}%
\psfrag{x06}[t][t][\tsize]{}%
\psfrag{x07}[t][t][\tsize]{}%
\psfrag{x08}[t][t][\tsize]{}%
\psfrag{x09}[t][t][\tsize]{}%
\psfrag{x10}[t][t][\tsize]{}%
\psfrag{x11}[t][t][\tsize]{}%
\psfrag{x12}[t][t][\tsize]{\scriptsize100}%
\psfrag{x13}[t][t][\tsize]{}%
\psfrag{x14}[t][t][\tsize]{\scriptsize300}%
\psfrag{x15}[t][t][\tsize]{\scriptsize100}%
\psfrag{x16}[t][t][\tsize]{}%
\psfrag{x17}[t][t][\tsize]{\scriptsize300}%
\psfrag{x18}[t][t][\tsize]{}%
\psfrag{x19}[t][t][\tsize]{100}%
\psfrag{x20}[t][t][\tsize]{}%
\psfrag{x21}[t][t][\tsize]{200}%
\psfrag{x22}[t][t][\tsize]{}%
\psfrag{x23}[t][t][\tsize]{300}%
%
\psfrag{v01}[r][r][\tsize]{}%
\psfrag{v02}[r][r][\tsize]{}%
\psfrag{v03}[r][r][\tsize]{}%
\psfrag{v04}[r][r][\tsize]{}%
\psfrag{v05}[r][r][\tsize]{}%
\psfrag{v06}[r][r][\tsize]{}%
\psfrag{v07}[r][r][\tsize]{}%
\psfrag{v08}[r][r][\tsize]{}%
\psfrag{v09}[r][r][\tsize]{}%
\psfrag{v10}[r][r][\tsize]{}%
\psfrag{v11}[r][r][\tsize]{}%
\psfrag{v12}[r][r][\tsize]{}%
\psfrag{v13}[r][r][\tsize]{}%
\psfrag{v14}[r][r][\tsize]{\scriptsize40}%
\psfrag{v15}[r][r][\tsize]{}%
\psfrag{v16}[r][r][\tsize]{\scriptsize80}%
\psfrag{v17}[r][r][\tsize]{}%
\psfrag{v18}[r][r][\tsize]{}%
\psfrag{v19}[r][r][\tsize]{}%
\psfrag{v20}[r][r][\tsize]{}%
\psfrag{v21}[r][r][\tsize]{\scriptsize40}%
\psfrag{v22}[r][r][\tsize]{}%
\psfrag{v23}[r][r][\tsize]{\scriptsize80}%
\psfrag{v24}[r][r][\tsize]{}%
\psfrag{v25}[r][r][\tsize]{}%
\psfrag{v26}[r][r][\tsize]{}%
\psfrag{v27}[r][r][\tsize]{100}%
\psfrag{v28}[r][r][\tsize]{}%
\psfrag{v29}[r][r][\tsize]{200}%
%
\psfrag{z01}[r][r][\tsize]{\scriptsize0}%
\psfrag{z02}[r][r][\tsize]{}%
\psfrag{z03}[r][r][\tsize]{\scriptsize10}%
\psfrag{z04}[r][r][\tsize]{\scriptsize0}%
\psfrag{z05}[r][r][\tsize]{}%
\psfrag{z06}[r][r][\tsize]{\scriptsize2}%
\psfrag{z07}[r][r][\tsize]{}%
%
\includegraphics[width=\textwidth]{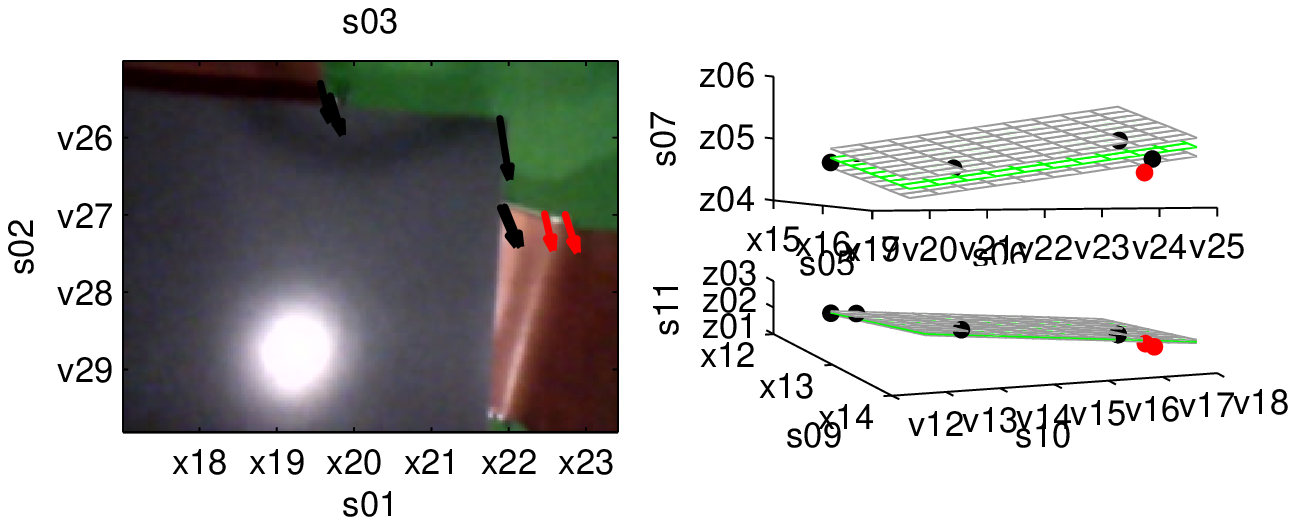}%
\end{psfrags}%
%

		\caption{Obstacle (table and chair)}
		\label{fig:FittingChairCanteen}
	\end{subfigure}	
	\begin{subfigure}[thpb]{0.45\textwidth}
%
%
\begin{psfrags}%
\psfragscanon%
\newcommand{\tsize}{0.7}
\newcommand{\tsizeb}{0.65}
%
\psfrag{s01}[t][t][\tsize]{\color[rgb]{0,0,0}\setlength{\tabcolsep}{0pt}\begin{tabular}{c}x\end{tabular}}%
\psfrag{s02}[b][b][\tsize]{\color[rgb]{0,0,0}\setlength{\tabcolsep}{0pt}\begin{tabular}{c}y\end{tabular}}%
\psfrag{s03}[b][b][\tsize]{\color[rgb]{0,0,0}\setlength{\tabcolsep}{0pt}\begin{tabular}{c}\end{tabular}}%
\psfrag{s05}[lt][lt][\tsize]{\color[rgb]{0,0,0}\setlength{\tabcolsep}{0pt}\begin{tabular}{r}x\end{tabular}}%
\psfrag{s06}[rt][rt][\tsize]{\color[rgb]{0,0,0}\setlength{\tabcolsep}{0pt}\begin{tabular}{l}y\end{tabular}}%
\psfrag{s07}[t][t][\tsize]{\color[rgb]{0,0,0}\setlength{\tabcolsep}{0pt}\begin{tabular}{c}u\end{tabular}}%
\psfrag{s09}[lt][lt][\tsize]{\color[rgb]{0,0,0}\setlength{\tabcolsep}{0pt}\begin{tabular}{c}x\end{tabular}}%
\psfrag{s10}[rt][rt][\tsize]{\color[rgb]{0,0,0}\setlength{\tabcolsep}{0pt}\begin{tabular}{c}y\end{tabular}}%
\psfrag{s11}[t][t][\tsize]{\color[rgb]{0,0,0}\setlength{\tabcolsep}{0pt}\begin{tabular}{c}v\end{tabular}}%
%
\psfrag{x01}[t][t][\tsize]{}%
\psfrag{x02}[t][t][\tsize]{}%
\psfrag{x03}[t][t][\tsize]{}%
\psfrag{x04}[t][t][\tsize]{}%
\psfrag{x05}[t][t][\tsize]{}%
\psfrag{x06}[t][t][\tsize]{}%
\psfrag{x07}[t][t][\tsize]{}%
\psfrag{x08}[t][t][\tsize]{}%
\psfrag{x09}[t][t][\tsize]{}%
\psfrag{x10}[t][t][\tsize]{}%
\psfrag{x11}[t][t][\tsize]{}%
\psfrag{x12}[t][t][\tsize]{}%
\psfrag{x13}[t][t][\tsize]{\scriptsize100}%
\psfrag{x14}[t][t][\tsize]{}%
\psfrag{x15}[t][t][\tsize]{\scriptsize300}%
\psfrag{x16}[t][t][\tsize]{}%
\psfrag{x17}[t][t][\tsize]{\scriptsize100}%
\psfrag{x18}[t][t][\tsize]{}%
\psfrag{x19}[t][t][\tsize]{\scriptsize300}%
\psfrag{x20}[t][t][\tsize]{}%
\psfrag{x21}[t][t][\tsize]{100}%
\psfrag{x22}[t][t][\tsize]{}%
\psfrag{x23}[t][t][\tsize]{200}%
\psfrag{x24}[t][t][\tsize]{}%
\psfrag{x25}[t][t][\tsize]{300}%
%
\psfrag{v01}[r][r][\tsize]{}%
\psfrag{v02}[r][r][\tsize]{}%
\psfrag{v03}[r][r][\tsize]{}%
\psfrag{v04}[r][r][\tsize]{}%
\psfrag{v05}[r][r][\tsize]{}%
\psfrag{v06}[r][r][\tsize]{}%
\psfrag{v07}[r][r][\tsize]{}%
\psfrag{v08}[r][r][\tsize]{}%
\psfrag{v09}[r][r][\tsize]{}%
\psfrag{v10}[r][r][\tsize]{}%
\psfrag{v11}[r][r][\tsize]{}%
\psfrag{v12}[r][r][\tsize]{}%
\psfrag{v13}[r][r][\tsize]{}%
\psfrag{v14}[r][r][\tsize]{\scriptsize20}%
\psfrag{v15}[r][r][\tsize]{}%
\psfrag{v16}[r][r][\tsize]{\scriptsize40}%
\psfrag{v17}[r][r][\tsize]{}%
\psfrag{v18}[r][r][\tsize]{}%
\psfrag{v19}[r][r][\tsize]{}%
\psfrag{v20}[r][r][\tsize]{}%
\psfrag{v21}[r][r][\tsize]{\scriptsize20}%
\psfrag{v22}[r][r][\tsize]{}%
\psfrag{v23}[r][r][\tsize]{}%
\psfrag{v24}[r][r][\tsize]{}%
\psfrag{v25}[r][r][\tsize]{\scriptsize40}%
\psfrag{v26}[r][r][\tsize]{}%
\psfrag{v27}[r][r][\tsize]{}%
\psfrag{v28}[r][r][\tsize]{100}%
\psfrag{v29}[r][r][\tsize]{}%
\psfrag{v30}[r][r][\tsize]{200}%
%
\psfrag{z01}[r][r][\tsize]{\scriptsize2.5}%
\psfrag{z02}[r][r][\tsize]{}%
\psfrag{z03}[r][r][\tsize]{\scriptsize3.5}%
\psfrag{z04}[r][r][\tsize]{}%
\psfrag{z05}[r][r][\tsize]{}%
\psfrag{z06}[r][r][\tsize]{\scriptsize-2}%
\psfrag{z07}[r][r][\tsize]{}%
\psfrag{z08}[r][r][\tsize]{\scriptsize2}%
\psfrag{z09}[r][r][\tsize]{}%
%
\includegraphics[width=\textwidth]{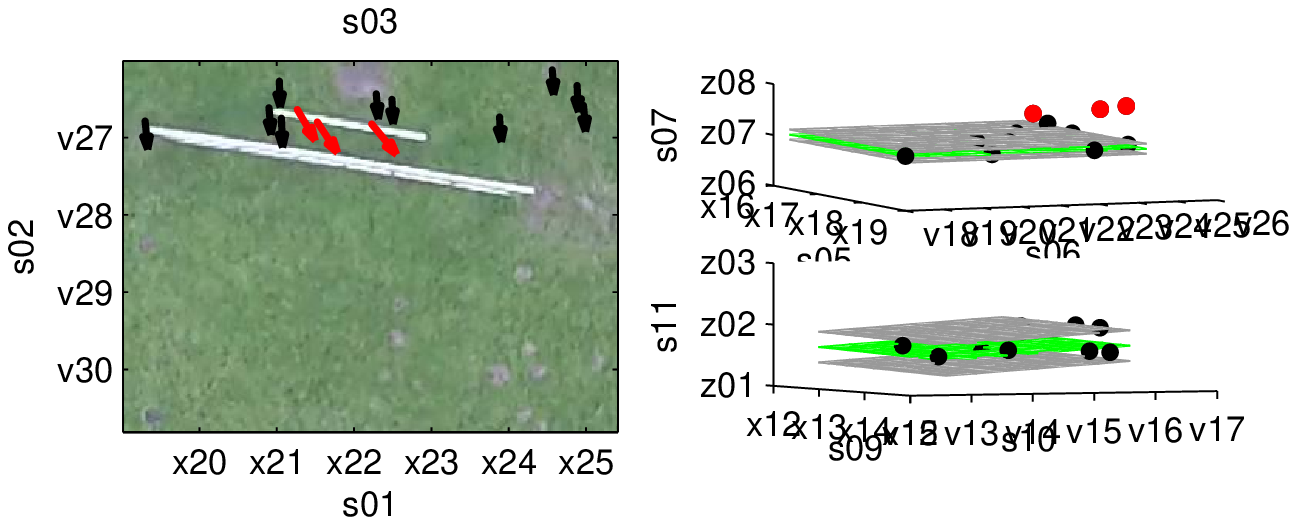}%
\end{psfrags}%
%

		\caption{Obstacle (fence)}
		\label{fig:FittingObject}
	\end{subfigure}
	\caption{Outdoor and indoor images without and with obstacles, and their corresponding optical flow fits. Non-obstacle: Consistent optical flow vectors (black) contribute to low fitting errors. Obstacle: Irregular optical flow vectors (red) lead to high fitting errors.}
	\label{fig:OpticalFlowFitting}
\end{figure}

To show that the proposed surface roughness, $\epsilon^*$ can be used as a metric to detect obstacles, images from the datasets were analyzed. Fig.~\ref{fig:OpticalFlowFitting} presents optical flow vectors measured from the consecutive images, and their fits of the optical flow field. In this figure, optical flow vectors are indicated by arrows in the image while the fit planes are presented in green and the error thresholds are shown as the gray planes. The red arrows in the images and red dots in the fits indicate which optical flow vectors contribute to larger error of the fits, the so-called outliers. When there is no obstacle or 3D structure in the images, the measured optical flow vectors are uniform, and thus most of them lie on the fitted plane within the bound of error threshold. This can be seen from the fits of the optical flow on the grass field in Fig.~\ref{fig:FittingGrassObject}. In contrast, irregular optical flow vectors can be expected from the images containing obstacles. This leads to more outliers or larger errors of the fits as shown in Fig.~\ref{fig:FittingTrees} (trees on the right), Fig.~\ref{fig:FittingChairCanteen} (chair below the table), and Fig.~\ref{fig:FittingObject} (fence on the top). 


\begin{figure}[thpb]
	\captionsetup[subfigure]{justification=centering}
	\centering
	\begin{subfigure}[thpb]{0.45\textwidth}
%
%
\begin{psfrags}%
\psfragscanon%
\newcommand{\tsize}{0.7}
\newcommand{\tsizeb}{0.65}
%
\psfrag{s05}[t][t][\tsize]{\color[rgb]{0,0,0}\setlength{\tabcolsep}{0pt}\begin{tabular}{c}Time~(s)\end{tabular}}%
\psfrag{s06}[b][b][\tsize]{\color[rgb]{0,0,0}\setlength{\tabcolsep}{0pt}\begin{tabular}{c}\Large$\epsilon^*$,~\Large$\hat{\epsilon}$\end{tabular}}%
\psfrag{s10}[][]{\color[rgb]{0,0,0}\setlength{\tabcolsep}{0pt}\begin{tabular}{c} \end{tabular}}%
\psfrag{s11}[][]{\color[rgb]{0,0,0}\setlength{\tabcolsep}{0pt}\begin{tabular}{c} \end{tabular}}%
\psfrag{s12}[l][l][\tsize]{\color[rgb]{0,0,0}$\hat{\epsilon}$}%
\psfrag{s13}[l][l][\tsize]{\color[rgb]{0,0,0}$\epsilon^*$}%
\psfrag{s14}[l][l][\tsize]{\color[rgb]{0,0,0}$\hat{\epsilon}$}%
%
\psfrag{x01}[t][t][\tsize]{0}%
\psfrag{x02}[t][t][\tsize]{5}%
\psfrag{x03}[t][t][\tsize]{10}%
\psfrag{x04}[t][t][\tsize]{15}%
\psfrag{x05}[t][t][\tsize]{20}%
\psfrag{x06}[t][t][\tsize]{25}%
\psfrag{x07}[t][t][\tsize]{30}%
\psfrag{x08}[t][t][\tsize]{35}%
\psfrag{x09}[t][t][\tsize]{40}%
%
\psfrag{v01}[r][r][\tsize]{0}%
\psfrag{v02}[r][r][\tsize]{0.5}%
\psfrag{v03}[r][r][\tsize]{1}%
%
\includegraphics[width=\textwidth]{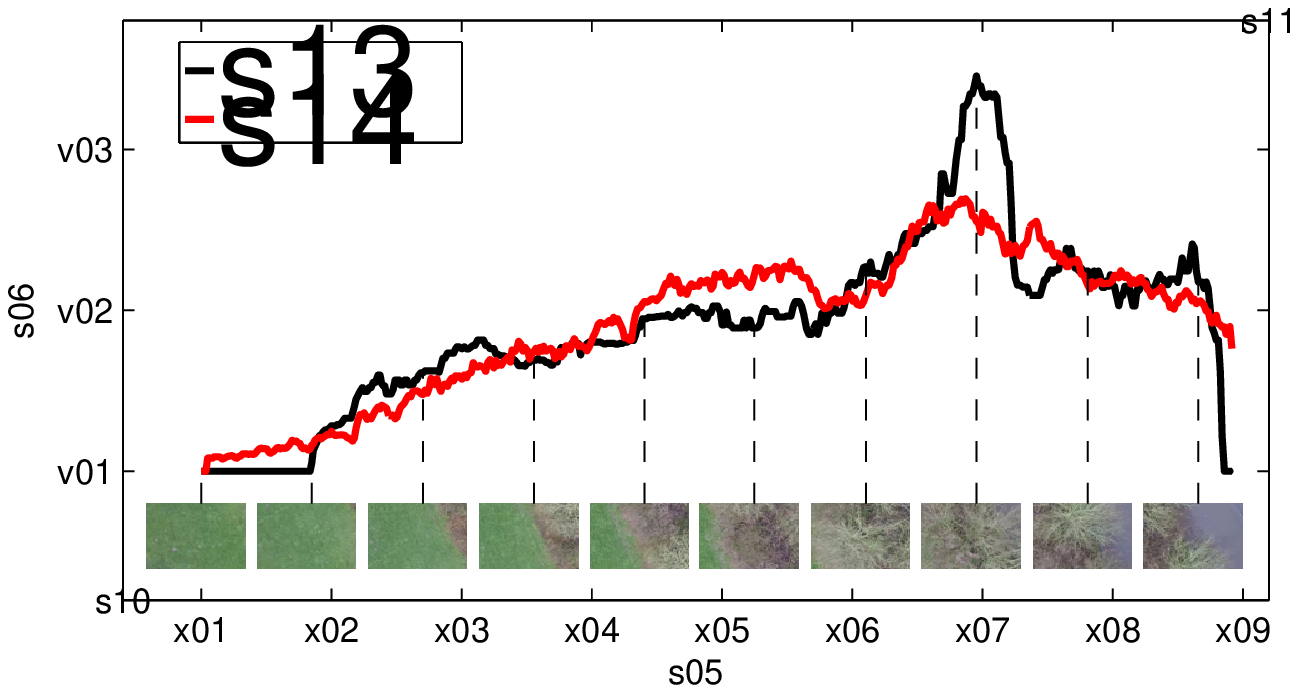}%
\end{psfrags}%
%

		\caption{Outdoor scene I (S9)}
		\label{fig:RoughnessTrees}
	\end{subfigure} \quad
	\begin{subfigure}[thpb]{0.45\textwidth}
%
%
\begin{psfrags}%
\psfragscanon%
\newcommand{\tsize}{0.7}
\newcommand{\tsizeb}{0.65}
%
\psfrag{s05}[t][t][\tsize]{\color[rgb]{0,0,0}\setlength{\tabcolsep}{0pt}\begin{tabular}{c}Time~(s)\end{tabular}}%
\psfrag{s06}[b][b][\tsize]{\color[rgb]{0,0,0}\setlength{\tabcolsep}{0pt}\begin{tabular}{c}\Large$\epsilon^*$,~\Large$\hat{\epsilon}$\end{tabular}}%
\psfrag{s10}[][]{\color[rgb]{0,0,0}\setlength{\tabcolsep}{0pt}\begin{tabular}{c} \end{tabular}}%
\psfrag{s11}[][]{\color[rgb]{0,0,0}\setlength{\tabcolsep}{0pt}\begin{tabular}{c} \end{tabular}}%
\psfrag{s12}[l][l][\tsize]{\color[rgb]{0,0,0}$\hat{\epsilon}$}%
\psfrag{s13}[l][l][\tsize]{\color[rgb]{0,0,0}$\epsilon^*$}%
\psfrag{s14}[l][l][\tsize]{\color[rgb]{0,0,0}$\hat{\epsilon}$}%
%
\psfrag{x01}[t][t][\tsize]{0}%
\psfrag{x02}[t][t][\tsize]{2}%
\psfrag{x03}[t][t][\tsize]{4}%
\psfrag{x04}[t][t][\tsize]{6}%
\psfrag{x05}[t][t][\tsize]{8}%
\psfrag{x06}[t][t][\tsize]{10}%
\psfrag{x07}[t][t][\tsize]{12}%
%
\psfrag{v01}[r][r][\tsize]{}%
\psfrag{v02}[r][r][\tsize]{0}%
\psfrag{v03}[r][r][\tsize]{}%
\psfrag{v04}[r][r][\tsize]{0.4}%
\psfrag{v05}[r][r][\tsize]{}%
\psfrag{v06}[r][r][\tsize]{0.8}%
\psfrag{v07}[r][r][\tsize]{}%
%
\includegraphics[width=\textwidth]{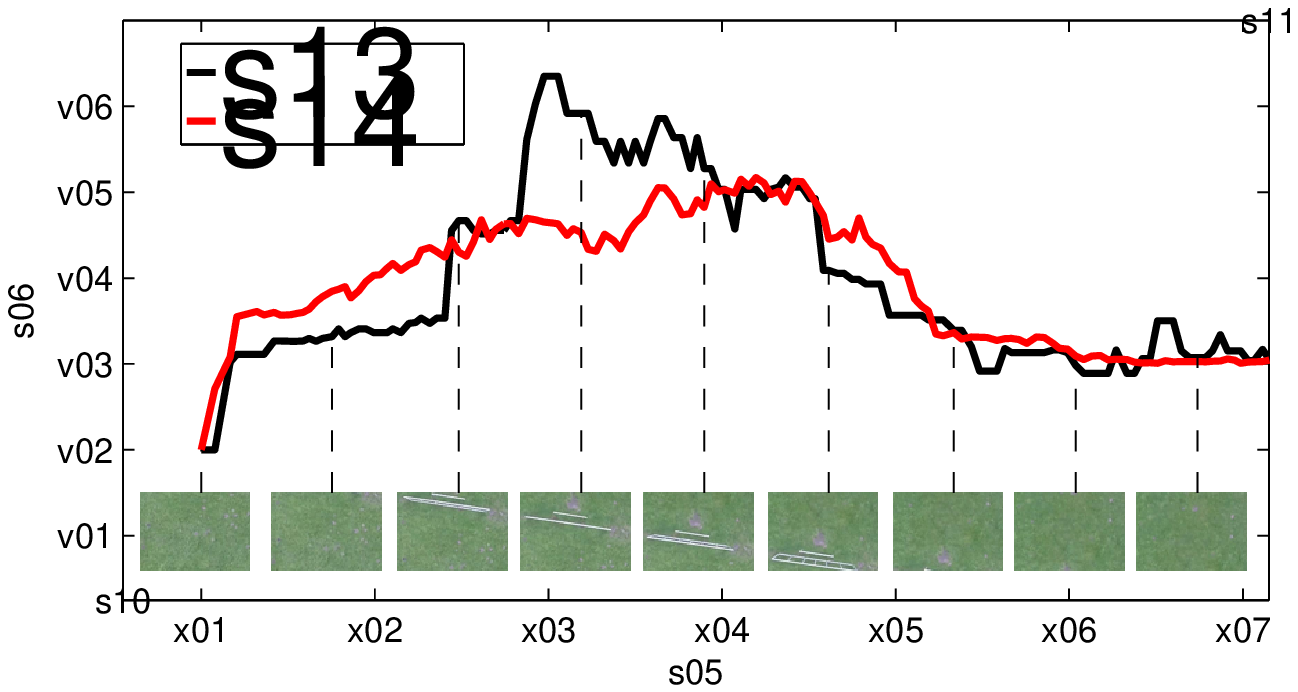}%
\end{psfrags}%
%

		\caption{Outdoor scene II (S2)}
		\label{fig:RoughnessObject}
	\end{subfigure}
	\begin{subfigure}[thpb]{0.45\textwidth}
%
%
\begin{psfrags}%
\psfragscanon%
\newcommand{\tsize}{0.7}
\newcommand{\tsizeb}{0.65}
%
\psfrag{s01}[t][t][\tsize]{\color[rgb]{0,0,0}\setlength{\tabcolsep}{0pt}\begin{tabular}{c}Time~(s)\end{tabular}}%
\psfrag{s02}[b][b][\tsize]{\color[rgb]{0,0,0}\setlength{\tabcolsep}{0pt}\begin{tabular}{c}\Large$\epsilon^*$,~\Large$\hat{\epsilon}$\end{tabular}}%
\psfrag{s06}[][]{\color[rgb]{0,0,0}\setlength{\tabcolsep}{0pt}\begin{tabular}{c} \end{tabular}}%
\psfrag{s07}[][]{\color[rgb]{0,0,0}\setlength{\tabcolsep}{0pt}\begin{tabular}{c} \end{tabular}}%
\psfrag{s08}[l][l][\tsize]{\color[rgb]{0,0,0}$\hat{\epsilon}$}%
\psfrag{s17}[l][l][\tsize]{\color[rgb]{0,0,0}$\epsilon^*$}%
\psfrag{s18}[l][l][\tsize]{\color[rgb]{0,0,0}$\hat{\epsilon}$}%
%
\psfrag{x01}[t][t][\tsize]{0}%
\psfrag{x02}[t][t][\tsize]{0.1}%
\psfrag{x03}[t][t][\tsize]{0.2}%
\psfrag{x04}[t][t][\tsize]{0.3}%
\psfrag{x05}[t][t][\tsize]{0.4}%
\psfrag{x06}[t][t][\tsize]{0.5}%
\psfrag{x07}[t][t][\tsize]{0.6}%
\psfrag{x08}[t][t][\tsize]{0.7}%
\psfrag{x09}[t][t][\tsize]{0.8}%
\psfrag{x10}[t][t][\tsize]{0.9}%
\psfrag{x11}[t][t][\tsize]{1}%
\psfrag{x12}[t][t][\tsize]{0}%
\psfrag{x13}[t][t][\tsize]{2}%
\psfrag{x14}[t][t][\tsize]{4}%
\psfrag{x15}[t][t][\tsize]{6}%
\psfrag{x16}[t][t][\tsize]{8}%
\psfrag{x17}[t][t][\tsize]{10}%
\psfrag{x18}[t][t][\tsize]{12}%
\psfrag{x19}[t][t][\tsize]{14}%
\psfrag{x20}[t][t][\tsize]{16}%
\psfrag{x21}[t][t][\tsize]{18}%
%
\psfrag{v01}[r][r][\tsize]{0}%
\psfrag{v02}[r][r][\tsize]{0.2}%
\psfrag{v03}[r][r][\tsize]{0.4}%
\psfrag{v04}[r][r][\tsize]{0.6}%
\psfrag{v05}[r][r][\tsize]{0.8}%
\psfrag{v06}[r][r][\tsize]{1}%
\psfrag{v07}[r][r][\tsize]{}%
\psfrag{v08}[r][r][\tsize]{0}%
\psfrag{v09}[r][r][\tsize]{}%
\psfrag{v10}[r][r][\tsize]{4}%
\psfrag{v11}[r][r][\tsize]{}%
\psfrag{v12}[r][r][\tsize]{8}%
\psfrag{v13}[r][r][\tsize]{}%
%
\includegraphics[width=\textwidth]{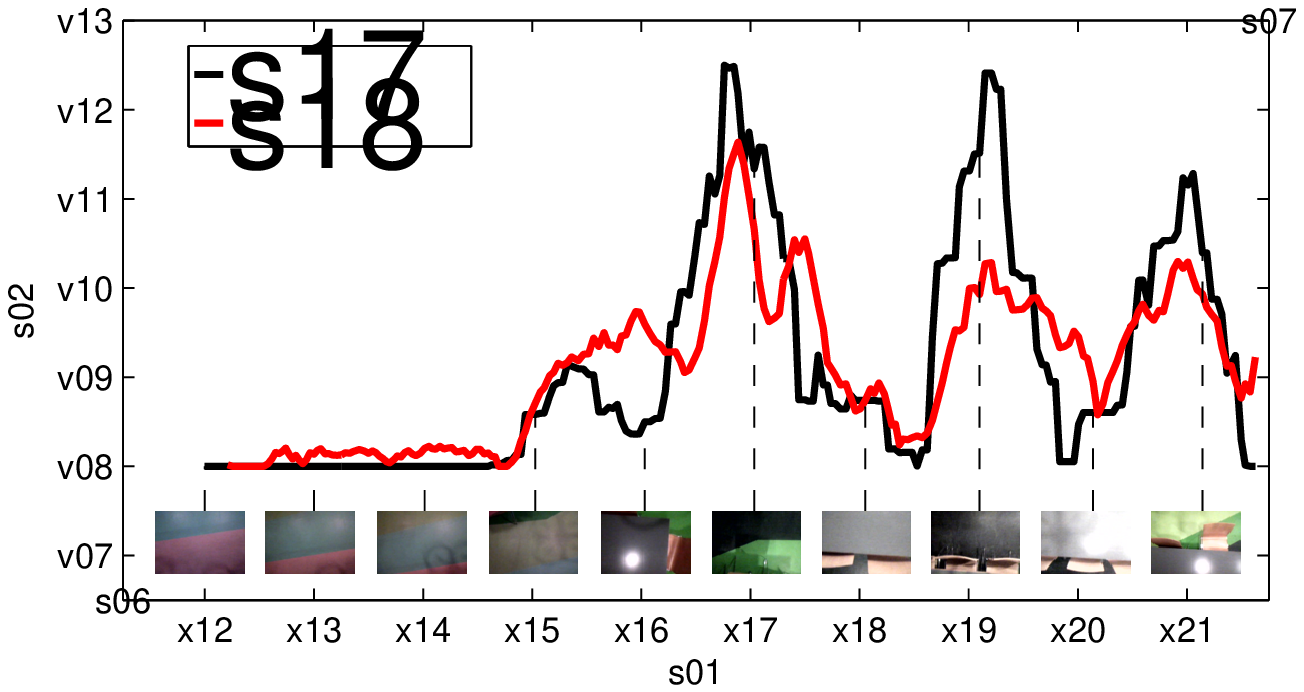}%
\end{psfrags}%
%

		\caption{Indoor scene I (S1)}
		\label{fig:RoughnessCanteen}
	\end{subfigure} \quad
	\begin{subfigure}[thpb]{0.45\textwidth}
%
%
\begin{psfrags}%
\psfragscanon%
\newcommand{\tsize}{0.7}
\newcommand{\tsizeb}{0.65}
%
\psfrag{s05}[t][t][\tsize]{\color[rgb]{0,0,0}\setlength{\tabcolsep}{0pt}\begin{tabular}{c}Time~(s)\end{tabular}}%
\psfrag{s06}[b][b][\tsize]{\color[rgb]{0,0,0}\setlength{\tabcolsep}{0pt}\begin{tabular}{c}\Large$\epsilon^*$,~\Large$\hat{\epsilon}$\end{tabular}}%
\psfrag{s10}[][]{\color[rgb]{0,0,0}\setlength{\tabcolsep}{0pt}\begin{tabular}{c} \end{tabular}}%
\psfrag{s11}[][]{\color[rgb]{0,0,0}\setlength{\tabcolsep}{0pt}\begin{tabular}{c} \end{tabular}}%
\psfrag{s12}[l][l][\tsize]{\color[rgb]{0,0,0}$\hat{\epsilon}$}%
\psfrag{s13}[l][l][\tsize]{\color[rgb]{0,0,0}$\epsilon^*$}%
\psfrag{s14}[l][l][\tsize]{\color[rgb]{0,0,0}$\hat{\epsilon}$}%
%
\psfrag{x01}[t][t][\tsize]{0}%
\psfrag{x02}[t][t][\tsize]{}%
\psfrag{x03}[t][t][\tsize]{4}%
\psfrag{x04}[t][t][\tsize]{}%
\psfrag{x05}[t][t][\tsize]{8}%
\psfrag{x06}[t][t][\tsize]{}%
\psfrag{x07}[t][t][\tsize]{12}%
\psfrag{x08}[t][t][\tsize]{}%
\psfrag{x09}[t][t][\tsize]{16}%
%
\psfrag{v01}[r][r][\tsize]{}%
\psfrag{v02}[r][r][\tsize]{0}%
\psfrag{v03}[r][r][\tsize]{}%
\psfrag{v04}[r][r][\tsize]{4}%
\psfrag{v05}[r][r][\tsize]{}%
\psfrag{v06}[r][r][\tsize]{8}%
\psfrag{v07}[r][r][\tsize]{}%
\psfrag{v08}[r][r][\tsize]{12}%
\psfrag{v09}[r][r][\tsize]{}%
%
\includegraphics[width=\textwidth]{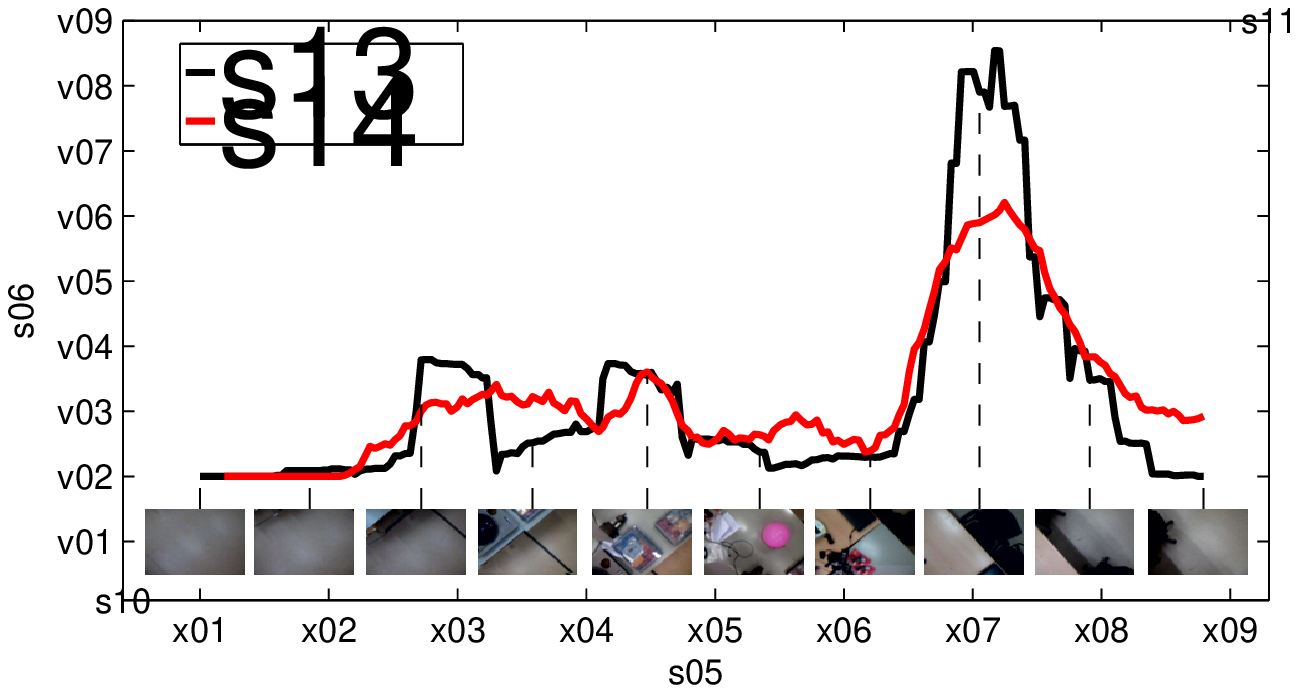}%
\end{psfrags}%
%

		\caption{Indoor scene II (S4)}
		\label{fig:RoughnessLab1}
	\end{subfigure}
	\caption{Obstacle detection using roughness estimates $\epsilon^*$ from optical flow algorithm and $\hat{\epsilon}$ from appearance while navigating.}
	\label{fig:RoughnessNavigating}
\end{figure}

In addition, we analyzed for all 9 scenes in both outdoor and indoor environments datasets the surface roughness, $\epsilon^*$ estimated from the optical flow algorithm (black line) when the MAV flew over areas without and with obstacles. Fig.~\ref{fig:RoughnessNavigating} presents the results of 2 scenes out of 9 for outdoor (S9 and S2) and indoor (S1 and S4) environments. This figure clearly shows that the roughness value is higher when there is an obstacle than when there is no obstacle on the landing surface. 

\begin{figure}[thpb]
	\centering
	\captionsetup{justification=centering}
	\includegraphics[width=0.45\textwidth]{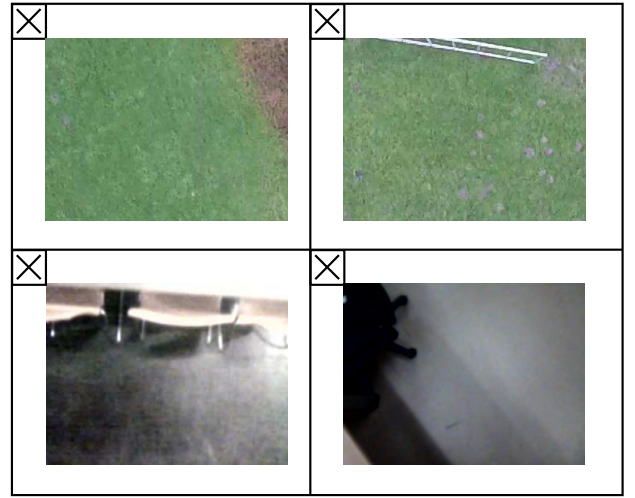} 
	\caption{Undetected obstacles with surface roughness, $\epsilon^*$.}
	\label{fig:MissedImages}
\end{figure}

To evaluate the classification capability using $\epsilon^*$, we labeled the test set using a threshold selected based on Fig.~\ref{fig:threshold_selection_height} and compared to manually labeled results using images. The true positive (TP) rate (the portion of obstacles that are detected) and false positive (FP) rate (the portion of empty landing areas wrongly classified as containing an obstacle) for all scenes shown in Fig.~\ref{fig:RoughnessNavigating} are computed and shown in TABLE~\ref{table:PerformaceClassificationRoughness}. In this table, most of the classification results show high proportions of correct identification of obstacles and non-obstacles images. However, the fence shown in Fig.~\ref{fig:RoughnessObject} can sometimes be missed by the feature point tracker set to only track $25$ points. The undetected obstacles mostly appear at the border of the images as shown in Fig.~\ref{fig:MissedImages}. For example, we can see that from top left to bottom right images in this figure, the undetected obstacles appear either on the top right (trees), top (fence), top (chairs), or left (chair). 	
The result shows that despite the sparseness of feature-based optical flow tracking, this algorithm generally manages to detect the obstacles in the field of view and thus identify safe landing spots. While results can be improved with more computational power, this work will show the applicability of the method even in the case of severe computational limitations. 

\begin{table}[thpb]
	\caption{Performance measure of a classification test using surface roughness, $\epsilon^*$}
	\label{table:PerformaceClassificationRoughness}
	\begin{center}
		\begin{tabular}{|c||c|c|c|c|}
			\hline
			Metrics  & Fig.~\ref{fig:RoughnessTrees} & Fig.~\ref{fig:RoughnessObject} & Fig.~\ref{fig:RoughnessCanteen} & Fig.~\ref{fig:RoughnessLab1} \\
			\hline
			\hline
			TP rate & 0.99 & 0.95 & 0.96 & 0.91\\
			\hline
			FP rate & 0.12 & 0.24 & 0.01 & 0.00\\
			\hline
		\end{tabular}
	\end{center}
\end{table}

\subsection{Obstacle Detection with Visual Appearance}
\label{subsec:ObstacleDetectionAppearance}
\begin{figure}[thpb]
	\captionsetup[subfigure]{justification=centering}
	\centering
	\begin{subfigure}[thpb]{0.335\textwidth}
		\includegraphics[width=\textwidth]{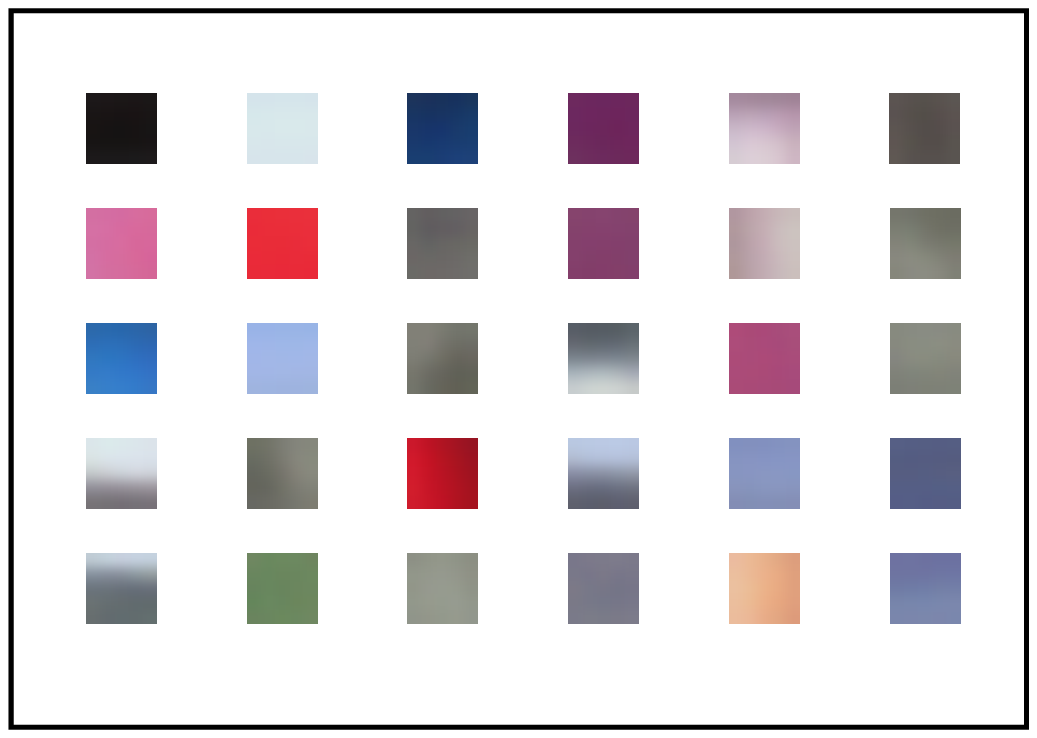}
		\caption{Outdoor dictionary}
		\label{fig:OutdoorDictionary}
	\end{subfigure} \quad
	\begin{subfigure}[thpb]{0.325\textwidth}
		\includegraphics[width=\textwidth]{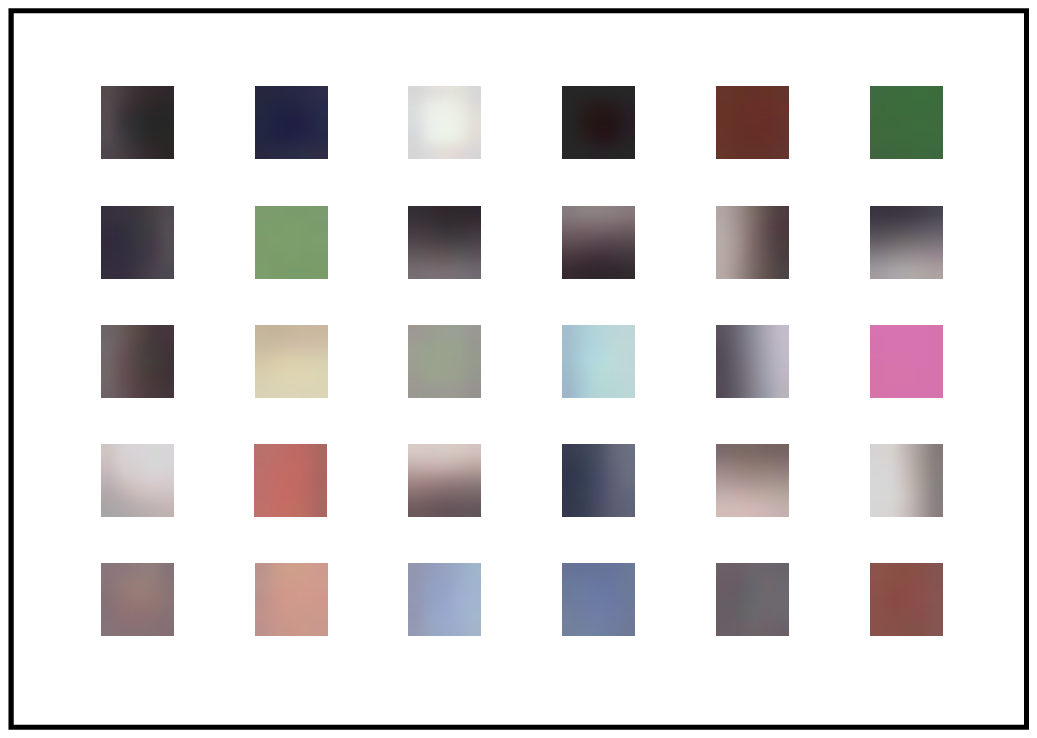}
		\caption{Indoor dictionary}
		\label{fig:IndoorDictionary}
	\end{subfigure}
	\caption{Trained dictionaries used in texton method for scene S5 in outdoor and indoor datasets.}
	\label{fig:Dictionary}
\end{figure}

In this subsection, training and testing is done in the same scene, where training is done on the first 80\% of a dataset and testing on the remaining 20\% of the data set. First, a dictionary was trained, and an example dictionary for scene S5 in each indoor and outdoor dataset is shown in Fig.~\ref{fig:Dictionary}. As can be seen from this figure, some textons represent a specific type of color while other textons represent a specific type of texture, such as horizontal or vertical gradients.

\begin{figure}[h!]
	\centering
	\captionsetup{justification=centering}
	\includegraphics[width=0.45\textwidth]{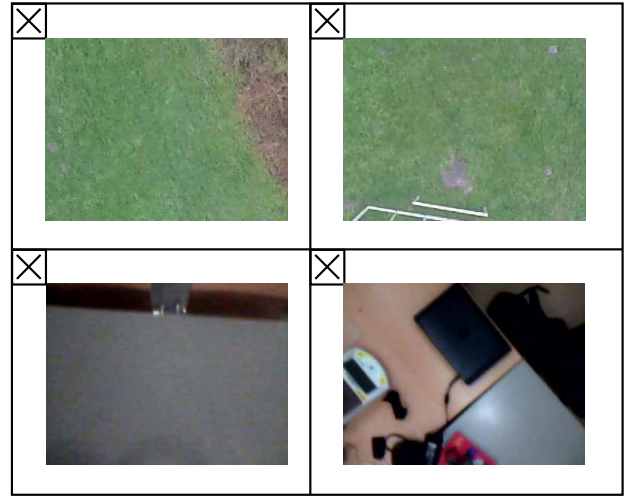} 
	\caption{Undetected obstacles with visual appearance, $\hat{\epsilon}$.}
	\label{fig:MissedImages_SSL}
\end{figure}

Then, a linear regression model was trained to map the texton distribution which represents the appearance of the image to the roughness $\epsilon^*$ estimated with the optical flow algorithm. TABLE~\ref{table:NRMSE_SSL} presents the performance metrics of the learning and classification shown in Fig.~\ref{fig:RoughnessNavigating}. The NRMSE show that it is more challenging to learn the regression model in more complex indoor environments where the appearance variation is larger. Concerning the TP rates, overall results are comparable or slightly better than the classification using $\epsilon^*$. However, for indoor environments, the FP rates are higher due to complexity of the indoor scenes. Fig.~\ref{fig:MissedImages_SSL} shows some undetected obstacles with visual appearance, $\hat{\epsilon}$. Although the TP rates are slightly higher than the TP rates of optical flow approach, obstacles appearing at the border of the image can sometimes be missed. The bottom right image is interesting, since the border of the table is the missed ``obstacle''. The objects on the table (like the closed laptop) are actually too flat to form a real obstacle to landing.	Next, we are going to show that the learned model is able to detect obstacles when there is a movement (while translating) and no movement (while hovering), and also the results of pixel-wise obstacle segmentation.

\begin{table}[thpb]
	\caption{Normalized Root Mean Square Errors (NRMSE) on test sets}
	\label{table:NRMSE_SSL}
	\begin{center}
		\begin{tabular}{|c||c|c|c|c|}
			\hline
			Metrics & Fig.~\ref{fig:RoughnessTrees} & Fig.~\ref{fig:RoughnessObject} & Fig.~\ref{fig:RoughnessCanteen} & Fig.~\ref{fig:RoughnessLab1} \\
			\hline
			\hline
			NRMSE (\%) & 10.8 & 12.7 & 20.7 & 19.1\\
			\hline
			TP rate & 0.96 & 0.96 & 0.97 & 0.96\\
			\hline
			FP rate & 0.00 & 0.20 & 0.51 & 0.18\\
			\hline
		\end{tabular}
	\end{center}
\end{table}

\subsubsection{While Translating} 
\label{subsubsec:translating}
To compare $\hat{\epsilon}$ with $\epsilon^*$ from optical flow algorithm, we also plotted $\hat{\epsilon}$ (red line) on Fig.~\ref{fig:RoughnessNavigating}. This figure clearly shows that both roughness estimates are higher when there is an obstacle than when there is no obstacle on the landing surface. Thus, the results demonstrate that the MAV can detect obstacles using $\hat{\epsilon}$ and thus identify safe spots to land upon. 

\subsubsection{While Hovering}
\label{subsubsec:hovering}
\begin{figure}[thpb]
	\captionsetup[subfigure]{justification=centering}
	\centering
	\begin{subfigure}[thpb]{0.48\textwidth}
%
%
\begin{psfrags}%
\psfragscanon%
\newcommand{\tsize}{0.7}
\newcommand{\tsizeb}{0.65}
%
\psfrag{s03}[b][b][\tsize]{\color[rgb]{0,0,0}\setlength{\tabcolsep}{0pt}\begin{tabular}{c}\end{tabular}}%
\psfrag{s04}[t][t][\tsize]{\color[rgb]{0,0,0}\setlength{\tabcolsep}{0pt}\begin{tabular}{c}x\end{tabular}}%
\psfrag{s05}[b][b][\tsize]{\color[rgb]{0,0,0}\setlength{\tabcolsep}{0pt}\begin{tabular}{c}y\end{tabular}}%
\psfrag{s06}[t][t][\tsize]{\color[rgb]{0,0,0}\setlength{\tabcolsep}{0pt}\begin{tabular}{c}Textons\end{tabular}}%
\psfrag{s07}[b][b][\tsize]{\color[rgb]{0,0,0}\setlength{\tabcolsep}{0pt}\begin{tabular}{c}$\mathbf{q}$\end{tabular}}%
\psfrag{s08}[b][b][\tsize]{\color[rgb]{0,0,0}\setlength{\tabcolsep}{0pt}\begin{tabular}{c}$\hat{\epsilon} = 0.034811$\end{tabular}}%
%
\psfrag{x01}[t][t][\tsize]{0}%
\psfrag{x02}[t][t][\tsize]{5}%
\psfrag{x03}[t][t][\tsize]{10}%
\psfrag{x04}[t][t][\tsize]{15}%
\psfrag{x05}[t][t][\tsize]{20}%
\psfrag{x06}[t][t][\tsize]{25}%
\psfrag{x07}[t][t][\tsize]{30}%
\psfrag{x08}[t][t][\tsize]{}%
\psfrag{x09}[t][t][\tsize]{100}%
\psfrag{x10}[t][t][\tsize]{}%
\psfrag{x11}[t][t][\tsize]{}%
\psfrag{x12}[t][t][\tsize]{200}%
\psfrag{x13}[t][t][\tsize]{}%
\psfrag{x14}[t][t][\tsize]{}%
\psfrag{x15}[t][t][\tsize]{300}%
%
\psfrag{v01}[r][r][\tsize]{0}%
\psfrag{v02}[r][r][\tsize]{0.2}%
\psfrag{v03}[r][r][\tsize]{0.4}%
\psfrag{v04}[r][r][\tsize]{0.6}%
\psfrag{v05}[r][r][\tsize]{0.8}%
\psfrag{v06}[r][r][\tsize]{1}%
\psfrag{v07}[r][r][\tsize]{}%
\psfrag{v08}[r][r][\tsize]{100}%
\psfrag{v09}[r][r][\tsize]{}%
\psfrag{v10}[r][r][\tsize]{}%
\psfrag{v11}[r][r][\tsize]{200}%
\psfrag{v12}[r][r][\tsize]{}%
%
\includegraphics[width=\textwidth]{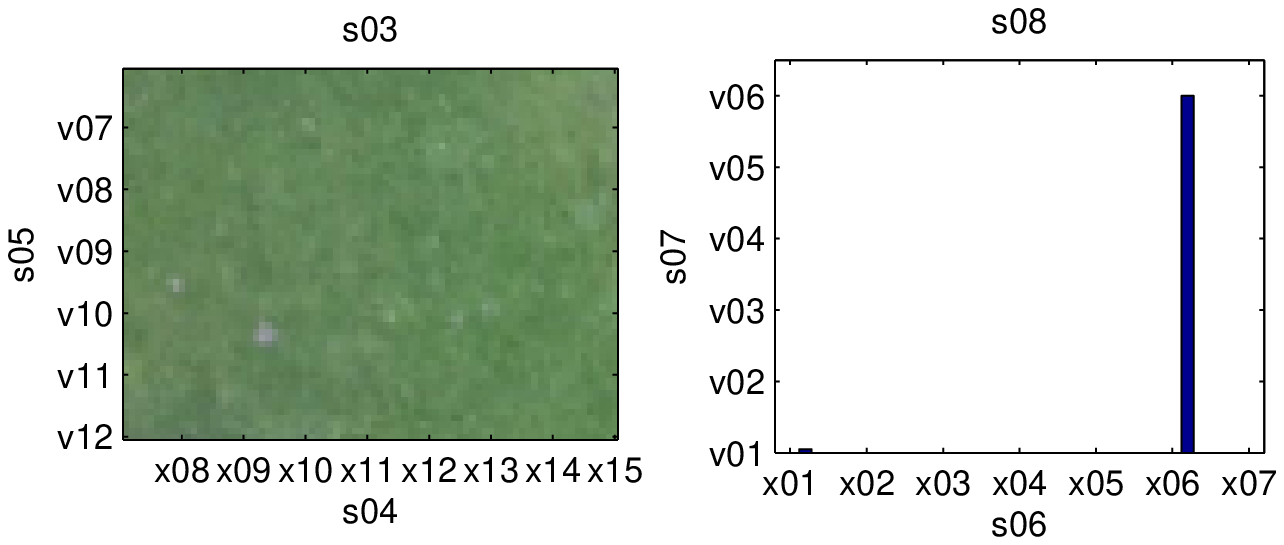}%
\end{psfrags}%
%

		\caption{Grass field}
		\label{fig:GrassHover}
	\end{subfigure} \\
	\begin{subfigure}[thpb]{0.48\textwidth}
%
%
\begin{psfrags}%
\psfragscanon%
\newcommand{\tsize}{0.7}
\newcommand{\tsizeb}{0.65}
%
\psfrag{s03}[b][b][\tsize]{\color[rgb]{0,0,0}\setlength{\tabcolsep}{0pt}\begin{tabular}{c}\end{tabular}}%
\psfrag{s04}[t][t][\tsize]{\color[rgb]{0,0,0}\setlength{\tabcolsep}{0pt}\begin{tabular}{c}x\end{tabular}}%
\psfrag{s05}[b][b][\tsize]{\color[rgb]{0,0,0}\setlength{\tabcolsep}{0pt}\begin{tabular}{c}y\end{tabular}}%
\psfrag{s06}[t][t][\tsize]{\color[rgb]{0,0,0}\setlength{\tabcolsep}{0pt}\begin{tabular}{c}Textons\end{tabular}}%
\psfrag{s07}[b][b][\tsize]{\color[rgb]{0,0,0}\setlength{\tabcolsep}{0pt}\begin{tabular}{c}$\mathbf{q}$\end{tabular}}%
\psfrag{s08}[b][b][\tsize]{\color[rgb]{0,0,0}\setlength{\tabcolsep}{0pt}\begin{tabular}{c}$\hat{\epsilon} = 0.79821$\end{tabular}}%
%
\psfrag{x01}[t][t][\tsize]{0}%
\psfrag{x02}[t][t][\tsize]{5}%
\psfrag{x03}[t][t][\tsize]{10}%
\psfrag{x04}[t][t][\tsize]{15}%
\psfrag{x05}[t][t][\tsize]{20}%
\psfrag{x06}[t][t][\tsize]{25}%
\psfrag{x07}[t][t][\tsize]{30}%
\psfrag{x08}[t][t][\tsize]{}%
\psfrag{x09}[t][t][\tsize]{100}%
\psfrag{x10}[t][t][\tsize]{}%
\psfrag{x11}[t][t][\tsize]{}%
\psfrag{x12}[t][t][\tsize]{200}%
\psfrag{x13}[t][t][\tsize]{}%
\psfrag{x14}[t][t][\tsize]{}%
\psfrag{x15}[t][t][\tsize]{300}%
%
\psfrag{v01}[r][r][\tsize]{0}%
\psfrag{v02}[r][r][\tsize]{0.2}%
\psfrag{v03}[r][r][\tsize]{0.4}%
\psfrag{v04}[r][r][\tsize]{0.6}%
\psfrag{v05}[r][r][\tsize]{0.8}%
\psfrag{v06}[r][r][\tsize]{1}%
\psfrag{v07}[r][r][\tsize]{}%
\psfrag{v08}[r][r][\tsize]{100}%
\psfrag{v09}[r][r][\tsize]{}%
\psfrag{v10}[r][r][\tsize]{}%
\psfrag{v11}[r][r][\tsize]{200}%
\psfrag{v12}[r][r][\tsize]{}%
%
\includegraphics[width=\textwidth]{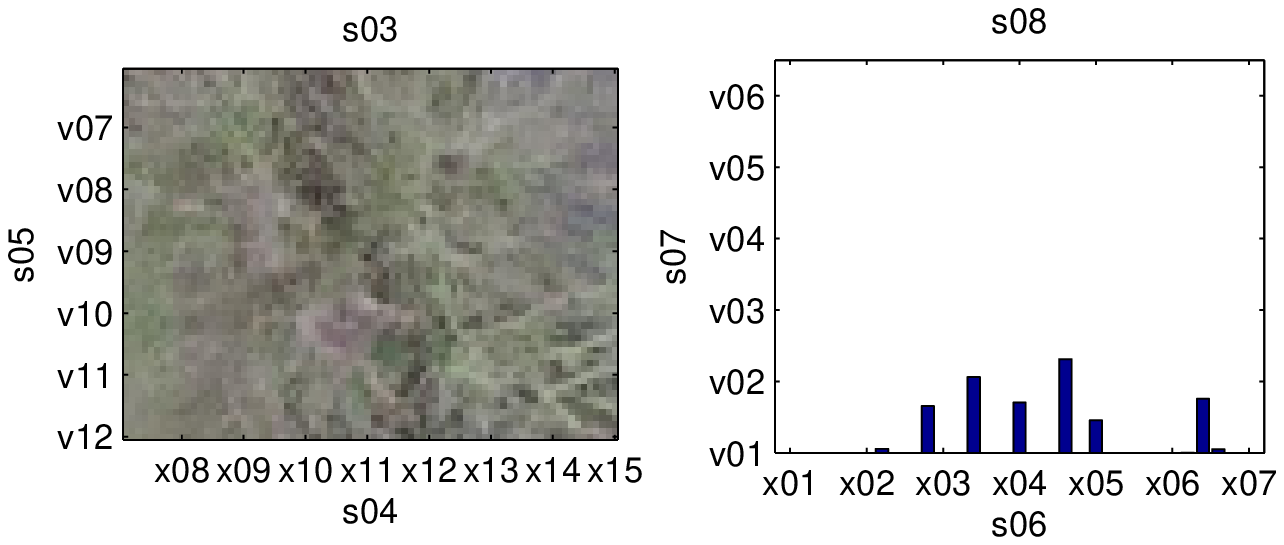}%
\end{psfrags}%
%

		\caption{Trees}
		\label{fig:TreesHover}
	\end{subfigure}
	\caption{Outdoor scene: Obstacle detection using roughness estimates $\hat{\epsilon}$ from appearance while hovering.}
	\label{fig:OutdoorHover}
\end{figure}

\begin{figure}[thpb]
	\captionsetup[subfigure]{justification=centering}
	\centering
	\begin{subfigure}[thpb]{0.48\textwidth}
%
%
\begin{psfrags}%
\psfragscanon%
\newcommand{\tsize}{0.7}
\newcommand{\tsizeb}{0.65}
%
\psfrag{s03}[b][b][\tsize]{\color[rgb]{0,0,0}\setlength{\tabcolsep}{0pt}\begin{tabular}{c}\end{tabular}}%
\psfrag{s04}[t][t][\tsize]{\color[rgb]{0,0,0}\setlength{\tabcolsep}{0pt}\begin{tabular}{c}x\end{tabular}}%
\psfrag{s05}[b][b][\tsize]{\color[rgb]{0,0,0}\setlength{\tabcolsep}{0pt}\begin{tabular}{c}y\end{tabular}}%
\psfrag{s06}[t][t][\tsize]{\color[rgb]{0,0,0}\setlength{\tabcolsep}{0pt}\begin{tabular}{c}Textons\end{tabular}}%
\psfrag{s07}[b][b][\tsize]{\color[rgb]{0,0,0}\setlength{\tabcolsep}{0pt}\begin{tabular}{c}$\mathbf{q}$\end{tabular}}%
\psfrag{s08}[b][b][\tsize]{\color[rgb]{0,0,0}\setlength{\tabcolsep}{0pt}\begin{tabular}{c}$\hat{\epsilon} = 0.017557$\end{tabular}}%
%
\psfrag{x01}[t][t][\tsize]{0}%
\psfrag{x02}[t][t][\tsize]{5}%
\psfrag{x03}[t][t][\tsize]{10}%
\psfrag{x04}[t][t][\tsize]{15}%
\psfrag{x05}[t][t][\tsize]{20}%
\psfrag{x06}[t][t][\tsize]{25}%
\psfrag{x07}[t][t][\tsize]{30}%
\psfrag{x08}[t][t][\tsize]{}%
\psfrag{x09}[t][t][\tsize]{100}%
\psfrag{x10}[t][t][\tsize]{}%
\psfrag{x11}[t][t][\tsize]{}%
\psfrag{x12}[t][t][\tsize]{200}%
\psfrag{x13}[t][t][\tsize]{}%
\psfrag{x14}[t][t][\tsize]{}%
\psfrag{x15}[t][t][\tsize]{300}%
%
\psfrag{v01}[r][r][\tsize]{0}%
\psfrag{v02}[r][r][\tsize]{0.2}%
\psfrag{v03}[r][r][\tsize]{0.4}%
\psfrag{v04}[r][r][\tsize]{0.6}%
\psfrag{v05}[r][r][\tsize]{0.8}%
\psfrag{v06}[r][r][\tsize]{1}%
\psfrag{v07}[r][r][\tsize]{}%
\psfrag{v08}[r][r][\tsize]{100}%
\psfrag{v09}[r][r][\tsize]{}%
\psfrag{v10}[r][r][\tsize]{}%
\psfrag{v11}[r][r][\tsize]{200}%
\psfrag{v12}[r][r][\tsize]{}%
%
\includegraphics[width=\textwidth]{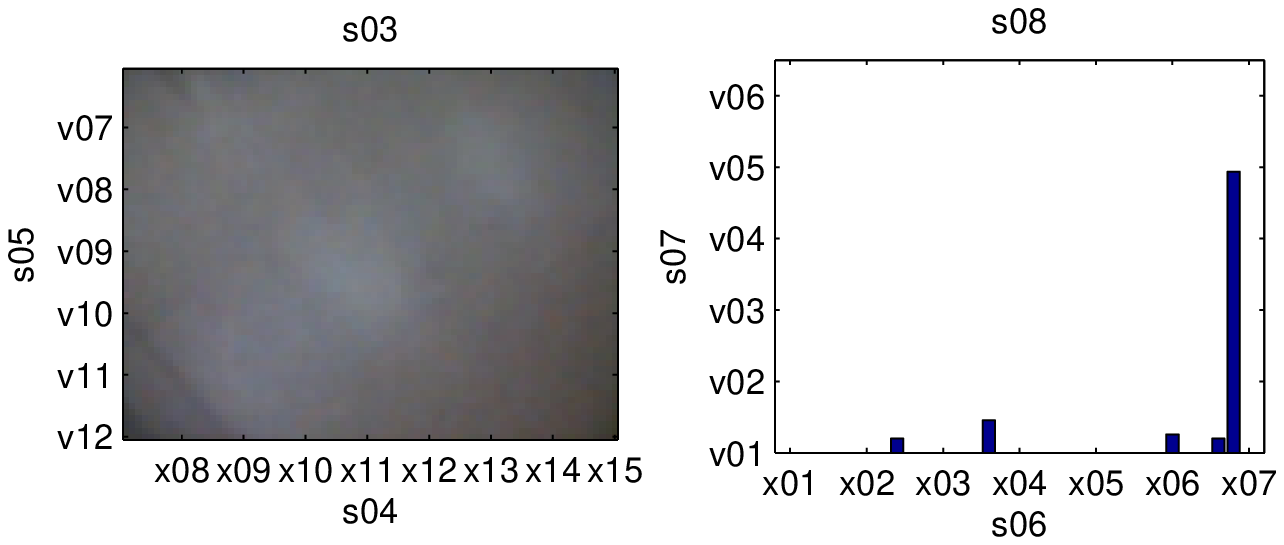}%
\end{psfrags}%
%

		\caption{Floor}
		\label{fig:FloorHover}
	\end{subfigure} \\
	\begin{subfigure}[thpb]{0.48\textwidth}
%
%
\begin{psfrags}%
\psfragscanon%
\newcommand{\tsize}{0.7}
\newcommand{\tsizeb}{0.65}
%
\psfrag{s03}[b][b][\tsize]{\color[rgb]{0,0,0}\setlength{\tabcolsep}{0pt}\begin{tabular}{c}\end{tabular}}%
\psfrag{s04}[t][t][\tsize]{\color[rgb]{0,0,0}\setlength{\tabcolsep}{0pt}\begin{tabular}{c}x\end{tabular}}%
\psfrag{s05}[b][b][\tsize]{\color[rgb]{0,0,0}\setlength{\tabcolsep}{0pt}\begin{tabular}{c}y\end{tabular}}%
\psfrag{s06}[t][t][\tsize]{\color[rgb]{0,0,0}\setlength{\tabcolsep}{0pt}\begin{tabular}{c}Textons\end{tabular}}%
\psfrag{s07}[b][b][\tsize]{\color[rgb]{0,0,0}\setlength{\tabcolsep}{0pt}\begin{tabular}{c}$\mathbf{q}$\end{tabular}}%
\psfrag{s08}[b][b][\tsize]{\color[rgb]{0,0,0}\setlength{\tabcolsep}{0pt}\begin{tabular}{c}$\hat{\epsilon} = 4.6584$\end{tabular}}%
%
\psfrag{x01}[t][t][\tsize]{0}%
\psfrag{x02}[t][t][\tsize]{5}%
\psfrag{x03}[t][t][\tsize]{10}%
\psfrag{x04}[t][t][\tsize]{15}%
\psfrag{x05}[t][t][\tsize]{20}%
\psfrag{x06}[t][t][\tsize]{25}%
\psfrag{x07}[t][t][\tsize]{30}%
\psfrag{x08}[t][t][\tsize]{}%
\psfrag{x09}[t][t][\tsize]{100}%
\psfrag{x10}[t][t][\tsize]{}%
\psfrag{x11}[t][t][\tsize]{}%
\psfrag{x12}[t][t][\tsize]{200}%
\psfrag{x13}[t][t][\tsize]{}%
\psfrag{x14}[t][t][\tsize]{}%
\psfrag{x15}[t][t][\tsize]{300}%
%
\psfrag{v01}[r][r][\tsize]{0}%
\psfrag{v02}[r][r][\tsize]{0.2}%
\psfrag{v03}[r][r][\tsize]{0.4}%
\psfrag{v04}[r][r][\tsize]{0.6}%
\psfrag{v05}[r][r][\tsize]{0.8}%
\psfrag{v06}[r][r][\tsize]{1}%
\psfrag{v07}[r][r][\tsize]{}%
\psfrag{v08}[r][r][\tsize]{100}%
\psfrag{v09}[r][r][\tsize]{}%
\psfrag{v10}[r][r][\tsize]{}%
\psfrag{v11}[r][r][\tsize]{200}%
\psfrag{v12}[r][r][\tsize]{}%
%
\includegraphics[width=\textwidth]{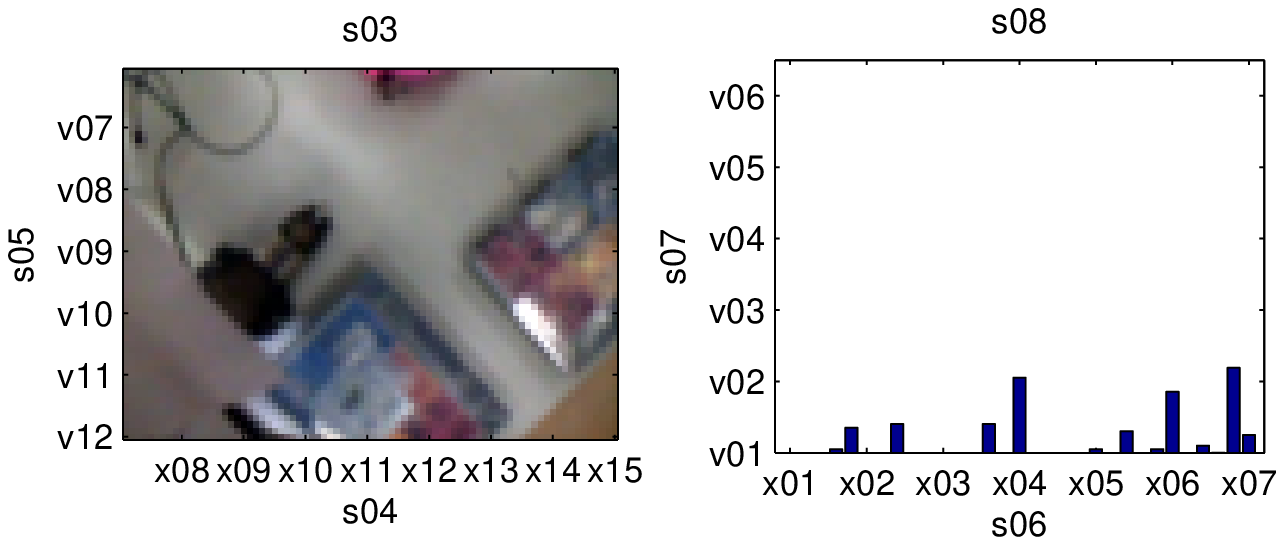}%
\end{psfrags}%
%

		\caption{Table}
		\label{fig:TableHover}
	\end{subfigure}
	\caption{Indoor scene: Obstacle detection using roughness estimates $\hat{\epsilon}$ from appearance while hovering.}
	\label{fig:IndoorHover}
\end{figure}

Having to move close to obstacles in order to detect them takes time and represents a risk. It would be more efficient and safer to detect obstacles without needing to move. Fig.~\ref{fig:OutdoorHover} and Fig.~\ref{fig:IndoorHover} show texton distributions for obstacles and non-obstacle images and their corresponding $\hat{\epsilon}$ for outdoor and indoor scenes, respectively. From these figures, we can observe that the texton distribution is rather different for images with and without obstacles. After learning this results in different surface roughness values $\hat{\epsilon}$. 

\subsubsection{Pixel-wise Obstacle Segmentation}
\label{subsubsec:PixelWiseSegmentation}
The roughness value resulting from the optical flow algorithm is a \emph{global} value for the presence of obstacle in the entire image. This study shows, after SSL, that the MAV will not only be able to detect the presence of obstacles, but will even be capable of pixel-wise segmentation of obstacles.
The basis for this capability is the \emph{local} nature of the image patches involved in the construction of the texton distribution. 

\begin{figure}[thpb]
	\captionsetup[subfigure]{justification=centering}
	\centering
	\begin{subfigure}[thpb]{0.45\textwidth}
		\includegraphics[width=\textwidth]{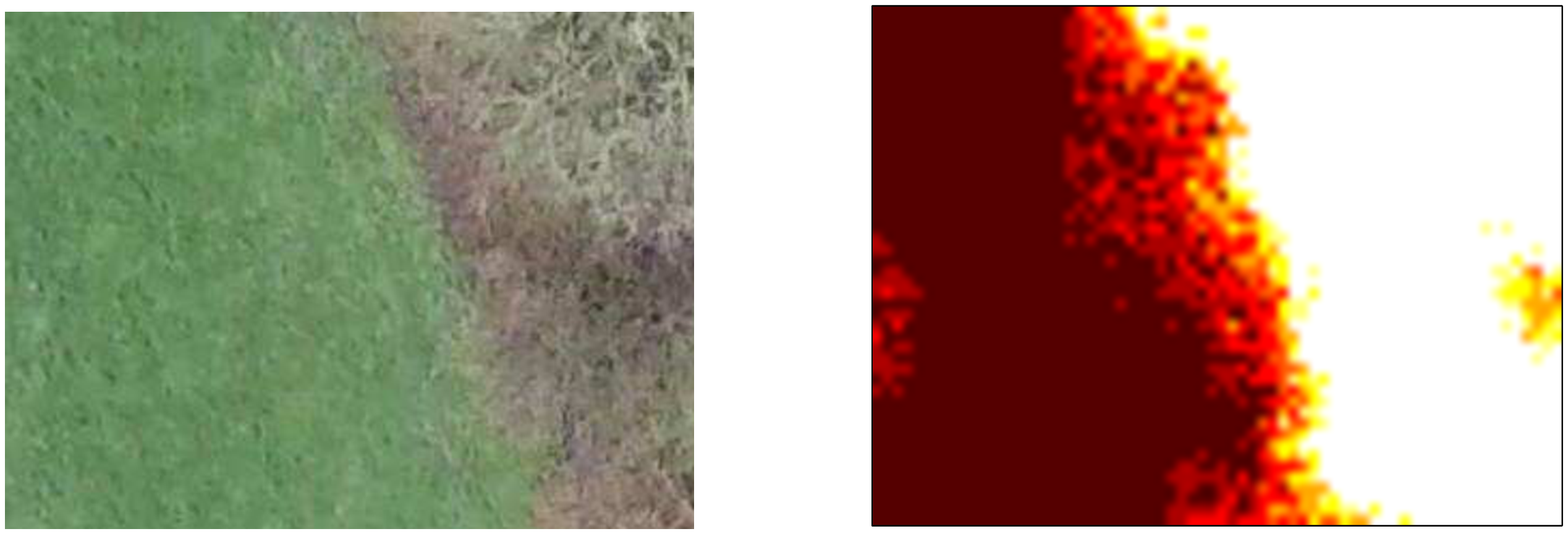}
		\caption{Outdoor Scene I (S9)}
		\label{fig:obstacleLocalizationOutdoor}
	\end{subfigure} \qquad
	\begin{subfigure}[thpb]{0.45\textwidth}
		\includegraphics[width=\textwidth]{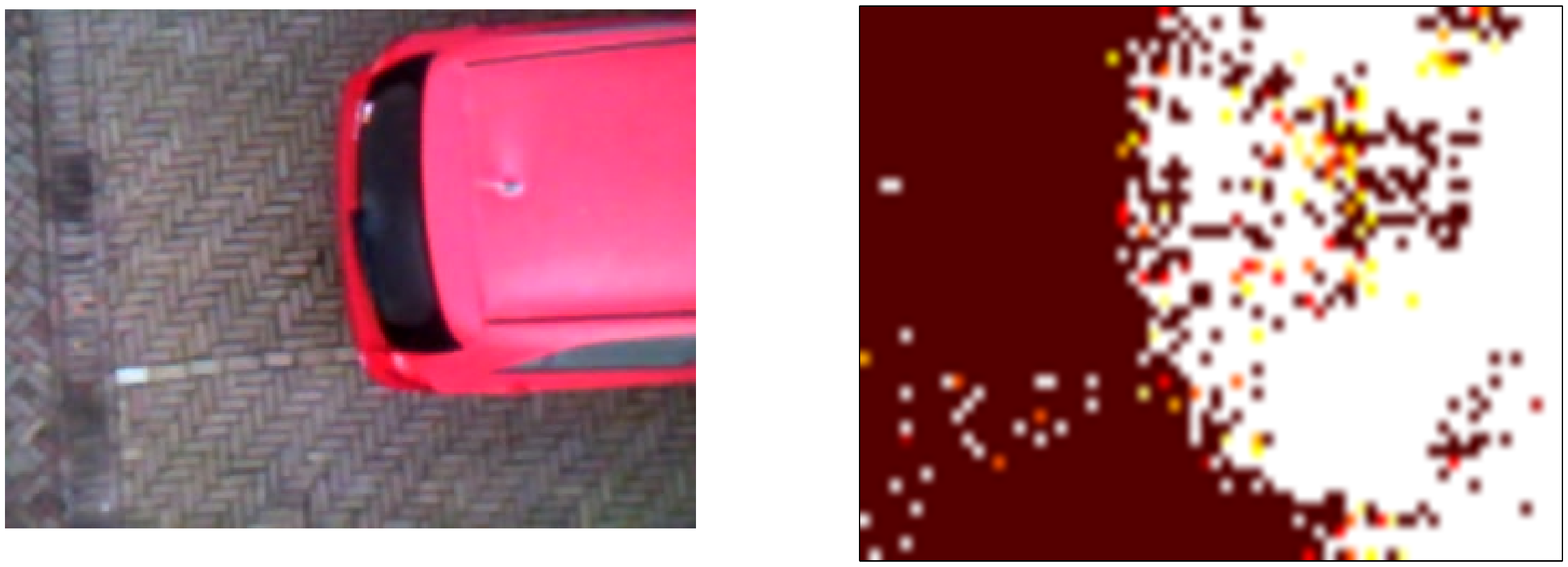}
		\caption{Outdoor Scene II (S6)}
		\label{fig:obstacleLocalizationOutdoor2}
	\end{subfigure} \\
	\begin{subfigure}[thpb]{0.45\textwidth}
		\includegraphics[width=\textwidth]{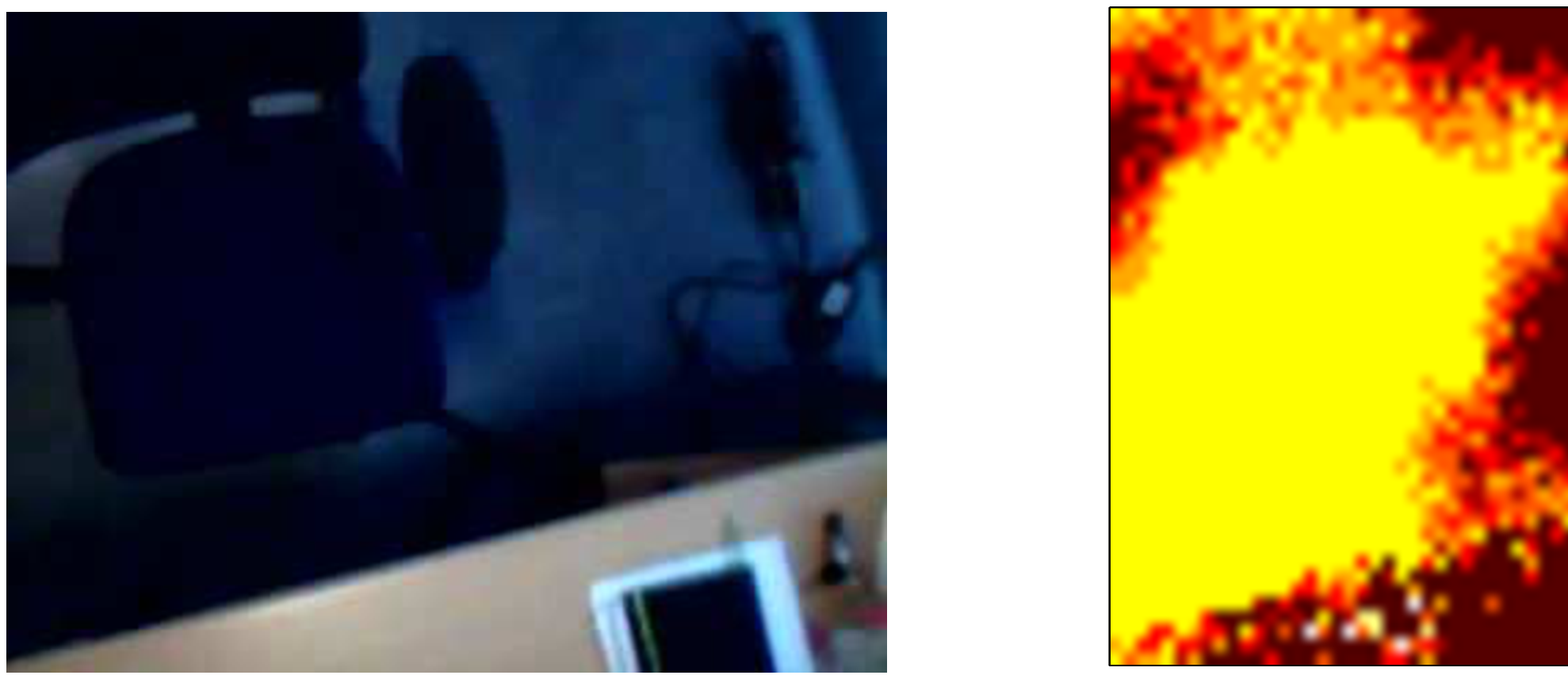}
		\caption{Indoor Scene I (S6)}
		\label{fig:obstacleLocalizationIndoor}
	\end{subfigure} \qquad
	\begin{subfigure}[thpb]{0.45\textwidth}
		\includegraphics[width=\textwidth]{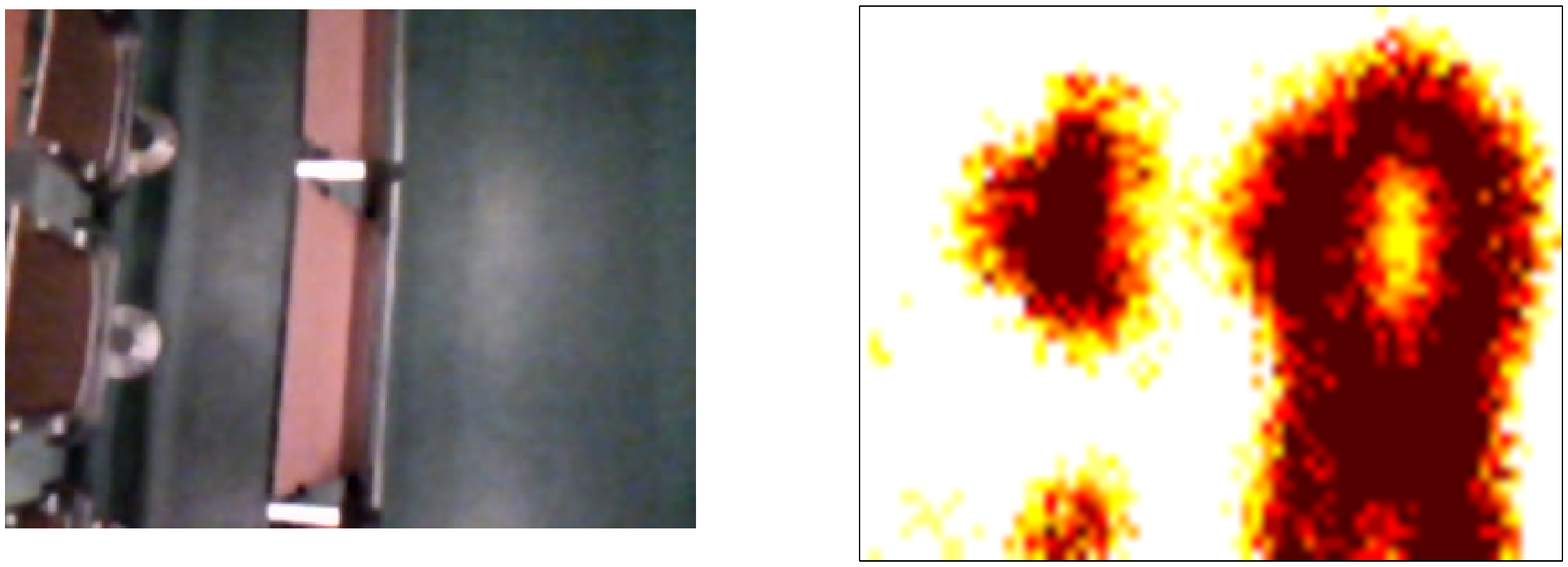}
		\caption{Indoor Scene II (S7)}
		\label{fig:obstacleLocalizationIndoor2}
	\end{subfigure}
	\caption{Obstacle localization using roughness $\hat{\epsilon}$ from SSL method. The light yellow color in roughness map represents the presence of obstacles.}
	\label{fig:obstacleLocalization}
\end{figure}

To show this, a sub-image with a window size of $50\times50$ pixels of an image was moved across $x-$axis of the image for each line in $y-$axis with increment of $4$ pixels until it covered the whole image. For each sub-image, the texton distribution was formed using $50$ image patches and mapped to a roughness value with the regression function discussed above. This creates a new image containing the roughness estimate for each area of the scene. Fig.~\ref{fig:obstacleLocalization} shows two still images from outdoor and indoor scenes and their corresponding regression maps which are color-coded using roughness values. In these figures, the obstacles clearly have a higher value (marked with light yellow color) in the regression maps while the safe landing area have lower value of roughness (marked with dark red color) in the map. In Outdoor Scene 1, it can be seen that the trees in the image on the right are seen as obstacle and marked as yellow. The same holds for the car in the image on the right in Outdoor Scene II are seen as an obstacle and marked as yellow. In Indoor Scene I, the black chair (top left in the image) and the border of the table (bottom right in the image) are detected as obstacles. The table itself is not seen as an obstacle: When having only the table in view, it is considered sufficiently flat and hereby offers an landing place for the MAV. The retractable tables and chairs (center and left in the image) in Indoor Scene II are also detected as obstacles while the floor is found as a safe place for landing. Note that this method with a moving window is used to show that our approach can also segment the obstacles in an image. However, it is computationally expensive since it processes almost every pixel in the image, and we actually do not need all the detailed information unless we need to land on a narrow place. 

\section{Generalization}
\label{sec:Generalization}

\subsection{Generalization of SSL Method to Different Environments}
\label{subsec:Generalization}
There is a main question remaining for the proposed method, i.e., how well the learned mapping from visual appearance to roughness will generalize to different environments. As in any learning scheme, this depends on the training and test distribution and on the learning method. In order to be successful, the training and test distribution should be sufficiently similar. In computer vision, this similarity does not only depend on the environment, but also on the visual features extracted from the images and how invariant they are to for instance rotation, scaling, and lighting changes.

In the context of SSL of obstacle appearance, we expect that features and learning methods can be found that generalize well over different environments and conditions. For instance, when we humans look at a Google maps image, we can discern obstacles such as trees and buildings rather well from areas that are more suitable for landing such as grass fields. Such a classification performance is also within reach of computer vision methods \cite{cesetti2010autonomous}. Of course, the computationally efficient texton distributions and straightforward learning methods used for onboard implementation on small MAVs are quite limited. However, even a limited generalization to a visually very different environment does not have to pose a problem in SSL method. Two strategies are available to deal with this: (1) continuously learn the mapping when the MAV is moving enough with respect to the visual scene, and (2) detecting when the learned mapping is receiving different inputs and hence producing uncertain outputs. If the estimated outputs are uncertain, the MAV can rely again on optical flow and adapt its mapping to the new environment. In this section, we show that the uncertainty of outputs in a visually different environment can be evaluated with a Naive Bayes classifier and Shannon entropy.

\subsubsection{Naive Bayes Classifier}
\label{subsubsec:NaiveBayes}
Given a distribution of $n$ textons ($\mathbf{q} = (q_1, \dots, q_n)$) to be classified, the Naive Bayes classifier for $k$ possible classes is given as in Eq.~(\ref{equation:NaiveBayes}). In this study, we have two classes ($k=2$), i.e., presence and absence of an obstacle, each of which can be represented by a distribution of $n=30$.  

\begin{equation}
p(C_k \vert q_1, \dots, q_n) \varpropto p(C_k) \prod_{i=1}^n p(q_i \vert C_k)
\label{equation:NaiveBayes}
\end{equation}

To create a Naive Bayes classifier, we first classify the distributions based on its corresponding roughness estimate, $\hat{\epsilon}$ into two classes labeled $C_1$ for obstacle or $C_2$ for non-obstacle according to Eq.~(\ref{equation:classlabel}). Based on this dataset, learning of the Naive Bayes classifier was performed using \textit{prtools} \cite{duin2000matlab} in MATLAB. 

\begin{equation}
\mathit{C}_k = \left\{ 
\begin{array}{l l}
C_1 & \quad \text{if $\hat{\epsilon} > \hat{\epsilon}_{th}$}\\
C_2 & \quad \text{if $\hat{\epsilon} < \hat{\epsilon}_{th}$}
\end{array} \right.  
\label{equation:classlabel}
\end{equation}

\subsubsection{Shannon Entropy}
\label{subsubsec:ShannonEntropy}
In information theory, Shannon entropy can be used to provide the amount of \textquoteleft disorder\textquoteright or uncertainty of a system \cite{shannon2001mathematical}. In this study, the entropy, $H$ can also be implemented to detect the change of the environment online based on the outputs of Naive Bayes classifier, as expressed in Eq.~(\ref{equation:entropy}).
\begin{equation}
H=\sum_{k=1}^2p(C_k \vert \mathbf{q})\log_2 (p(C_k \vert \mathbf{q}))
\label{equation:entropy}
\end{equation}

\subsubsection{Analysis on Two Different Environments}
\label{subsubsec:TwoScenes}
\begin{figure}[thpb]
	\captionsetup{justification=centering}
	\centering
	\begin{subfigure}[thpb]{0.5\textwidth}
		\includegraphics[width=\textwidth]{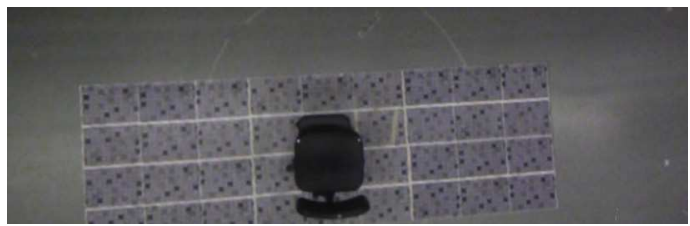}
		\caption{Environment 1 (E1)}
		\label{fig:Scene1}
	\end{subfigure} \\
	\begin{subfigure}[thpb]{0.5\textwidth}
		\includegraphics[width=\textwidth]{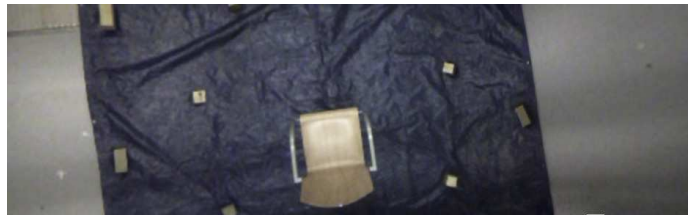}
		\caption{Environment 2 (E2)}
		\label{fig:Scene2}
	\end{subfigure}
	\caption{Images of two different environments stitched using on-board images.}
	\label{fig:TwoEnvironment}
\end{figure}

The MAV was flown in two visually different environments (E1 and E2) in which one different obstacle was placed as shown in Fig.~\ref{fig:TwoEnvironment}. A linear regression model was trained in E1 and then the texton distributions were logged for E1 and E2 by repeatedly flying the MAV over the obstacles. Here, we investigate how well the regression model trained in E1 performs in E2. Please note that despite the use of a similar object (a chair), the environments are visually very different (the chair being dark in E1 and bright in E2, the surface being grey in E1 and dark blue in E2). 

$80\%$ of the distributions in E1 were used to train the Naive Bayes classifier and the rest of the distributions were used for testing purposes. Both test sets from E1 and E2 were tested on the Naive Bayes classifier and the classification error, $Err$ is computed using $Err=(FP+FN)/(n_{test})$,
where $FP$ and $FN$ are the number of false positive and false negative, respectively from a total number of the test data, $n_{test}$. The errors on test sets for E1 and E2 are $4.67\%$ and $11.37\%$, respectively. This error evaluates the dataset by the classifier without considering the class prior. The error is higher for the test set in E2, thus indicating that the generalization to E2 is indeed more difficult than the generalization to E1's test set.

\begin{figure}[thpb]
	\centering
	\captionsetup{justification=centering}
%
%
\begin{psfrags}%
\psfragscanon%
\newcommand{\tsize}{0.7}
\newcommand{\tsizeb}{0.65}
%
\psfrag{s06}[b][b][\tsize]{\color[rgb]{0,0,0}\setlength{\tabcolsep}{0pt}\begin{tabular}{c}\Large$\hat{\epsilon}$\end{tabular}}%
\psfrag{s07}[t][t][\tsize]{\color[rgb]{0,0,0}\setlength{\tabcolsep}{0pt}\begin{tabular}{c}Time~(s)\end{tabular}}%
\psfrag{s08}[b][b][\tsize]{\color[rgb]{0,0,0}\setlength{\tabcolsep}{0pt}\begin{tabular}{c}H\end{tabular}}%
%
\psfrag{x01}[t][t][\tsize]{5}%
\psfrag{x02}[t][t][\tsize]{10}%
\psfrag{x03}[t][t][\tsize]{15}%
\psfrag{x04}[t][t][\tsize]{20}%
\psfrag{x05}[t][t][\tsize]{25}%
\psfrag{x06}[t][t][\tsize]{30}%
\psfrag{x07}[t][t][\tsize]{35}%
\psfrag{x08}[t][t][\tsize]{40}%
\psfrag{x09}[t][t][\tsize]{45}%
\psfrag{x10}[t][t][\tsize]{50}%
\psfrag{x11}[t][t][\tsize]{55}%
%
\psfrag{v01}[r][r][\tsize]{0}%
\psfrag{v02}[r][r][\tsize]{}%
\psfrag{v03}[r][r][\tsize]{0.4}%
\psfrag{v04}[r][r][\tsize]{}%
\psfrag{v05}[r][r][\tsize]{0.8}%
\psfrag{v06}[r][r][\tsize]{}%
\psfrag{v07}[r][r][\tsize]{0}%
\psfrag{v08}[r][r][\tsize]{}%
\psfrag{v09}[r][r][\tsize]{20}%
\psfrag{v10}[r][r][\tsize]{}%
\psfrag{v11}[r][r][\tsize]{40}%
\psfrag{v12}[r][r][\tsize]{}%
\psfrag{v13}[r][r][\tsize]{60}%
\psfrag{v14}[r][r][\tsize]{}%
\psfrag{v15}[r][r][\tsize]{80}%
\psfrag{v16}[r][r][\tsize]{}%
\psfrag{v17}[r][r][\tsize]{100}%
%
\includegraphics[width=0.5\textwidth]{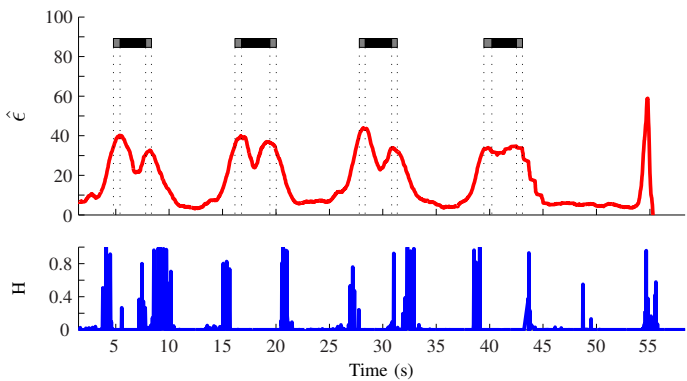}
\end{psfrags}%
%

	\caption{Uncertainty measure H (bottom) and roughness estimate $\hat{\epsilon}$ (top) in E1.}
	\label{fig:UncertaintyEnvironment1}
\end{figure}

\begin{figure}[thpb]
	\centering
	\captionsetup{justification=centering}
%
%
\begin{psfrags}%
\psfragscanon%
\newcommand{\tsize}{0.7}
\newcommand{\tsizeb}{0.65}
%
\psfrag{s06}[b][b][\tsize]{\color[rgb]{0,0,0}\setlength{\tabcolsep}{0pt}\begin{tabular}{c}\Large$\hat{\epsilon}$\end{tabular}}%
\psfrag{s07}[t][t][\tsize]{\color[rgb]{0,0,0}\setlength{\tabcolsep}{0pt}\begin{tabular}{c}Time~(s)\end{tabular}}%
\psfrag{s08}[b][b][\tsize]{\color[rgb]{0,0,0}\setlength{\tabcolsep}{0pt}\begin{tabular}{c}H\end{tabular}}%
%
\psfrag{x01}[t][t][\tsize]{5}%
\psfrag{x02}[t][t][\tsize]{10}%
\psfrag{x03}[t][t][\tsize]{15}%
\psfrag{x04}[t][t][\tsize]{20}%
\psfrag{x05}[t][t][\tsize]{25}%
\psfrag{x06}[t][t][\tsize]{30}%
\psfrag{x07}[t][t][\tsize]{35}%
\psfrag{x08}[t][t][\tsize]{40}%
\psfrag{x09}[t][t][\tsize]{45}%
\psfrag{x10}[t][t][\tsize]{50}%
\psfrag{x11}[t][t][\tsize]{55}%
%
\psfrag{v01}[r][r][\tsize]{0}%
\psfrag{v02}[r][r][\tsize]{}%
\psfrag{v03}[r][r][\tsize]{0.4}%
\psfrag{v04}[r][r][\tsize]{}%
\psfrag{v05}[r][r][\tsize]{0.8}%
\psfrag{v06}[r][r][\tsize]{}%
\psfrag{v07}[r][r][\tsize]{0}%
\psfrag{v08}[r][r][\tsize]{}%
\psfrag{v09}[r][r][\tsize]{20}%
\psfrag{v10}[r][r][\tsize]{}%
\psfrag{v11}[r][r][\tsize]{40}%
\psfrag{v12}[r][r][\tsize]{}%
\psfrag{v13}[r][r][\tsize]{60}%
\psfrag{v14}[r][r][\tsize]{}%
\psfrag{v15}[r][r][\tsize]{80}%
\psfrag{v16}[r][r][\tsize]{}%
\psfrag{v17}[r][r][\tsize]{100}%
%
\includegraphics[width=0.5\textwidth]{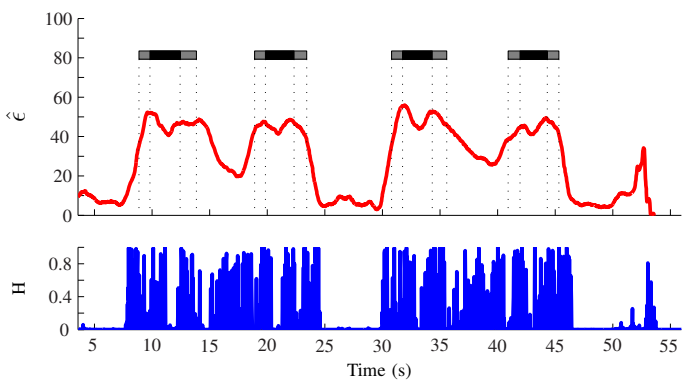}
\end{psfrags}%
%

	\caption{Uncertainty measure H (bottom) and roughness estimate $\hat{\epsilon}$ (top) in E2.}
	\label{fig:UncertaintyEnvironment2}
\end{figure}

In the top part of Fig.~\ref{fig:UncertaintyEnvironment1} and Fig.~\ref{fig:UncertaintyEnvironment2}, the red line is the roughness estimate and square boxes show ground truth position of the obstacle where black areas indicate full visibility of the obstacle in the field of view of the camera while half visibility is shown using gray areas. In the bottom part of these figures, the blue line shows the uncertainty of the outputs from the Naive Bayes classifier with the Shannon entropy. In Fig.~\ref{fig:UncertaintyEnvironment1}, the uncertainty can be observed at the edges of the obstacles. It is reasonable as only a very small part of the obstacle was captured in the image and it can actually give a good checking for the trained environment itself before making the decision. In Fig.~\ref{fig:UncertaintyEnvironment2}, the results of roughness estimate demonstrate that the SSL model actually remains quite effective even when flying in a different environment. Although these results show that the generalization is achieved to some extent, there are some regions which were wrongly classified (e.g. $15-20s$ and $35-40s$). In fact, the entropy gives a more continuous uncertainty of the outputs due to the difference in visual appearance in E2. There is one part (e.g. $25-30s$) where they both agree with their outputs because the field of view of the camera consists of largely the same gray ground on the right side of test field (see Fig.~\ref{fig:TwoEnvironment}). By using this information, the MAV is able to detect the change of environment and trigger the optical flow algorithm to re-train the linear regression model so that it can adapt itself to the new environment.


\subsubsection{K-fold Test on Multiple Real Scenes}
\label{subsubsec:KfoldTest}
\begin{figure}[thpb]
	\captionsetup[subfigure]{justification=centering}
	\centering
	\begin{subfigure}[thpb]{0.45\textwidth}
%
%
\begin{psfrags}%
\psfragscanon%
\newcommand{\tsize}{0.8}
\newcommand{\tsizeb}{0.5}
%
\psfrag{A}[t][t][\tsize]{\color[rgb]{0,0,0}\setlength{\tabcolsep}{0pt}\begin{tabular}{c}Scene~1\end{tabular}}%
\psfrag{B}[t][t][\tsize]{\color[rgb]{0,0,0}\setlength{\tabcolsep}{0pt}\begin{tabular}{c}Scene~2\end{tabular}}%
\psfrag{C}[t][t][\tsize]{\color[rgb]{0,0,0}\setlength{\tabcolsep}{0pt}\begin{tabular}{c}Scene~9\end{tabular}}%
\psfrag{J}[t][t][\tsize]{\color[rgb]{0,0,0}\setlength{\tabcolsep}{0pt}\begin{tabular}{c}$L_1$\end{tabular}}%
\psfrag{L}[t][t][\tsize]{\color[rgb]{0,0,0}\setlength{\tabcolsep}{0pt}\begin{tabular}{c}$T_1$\end{tabular}}%
\psfrag{M}[t][t][\tsize]{\color[rgb]{0,0,0}\setlength{\tabcolsep}{0pt}\begin{tabular}{c}$L_2$\end{tabular}}%
\psfrag{N}[t][t][\tsize]{\color[rgb]{0,0,0}\setlength{\tabcolsep}{0pt}\begin{tabular}{c}$T_2$\end{tabular}}%
\psfrag{O}[t][t][\tsize]{\color[rgb]{0,0,0}\setlength{\tabcolsep}{0pt}\begin{tabular}{c}$L_9$\end{tabular}}%
\psfrag{P}[t][t][\tsize]{\color[rgb]{0,0,0}\setlength{\tabcolsep}{0pt}\begin{tabular}{c}$T_9$\end{tabular}}%
%
\includegraphics[width=\textwidth]{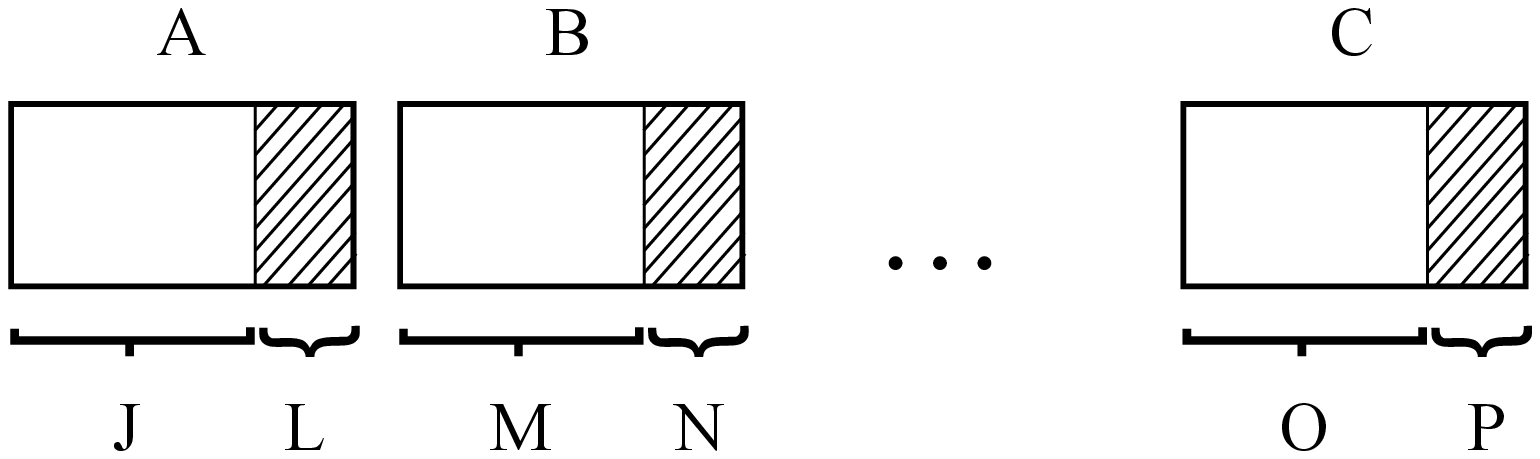}
\end{psfrags}%
%

		\caption{K-fold cross-validation 1}
		\label{fig:KFold1}
	\end{subfigure} \\
	\begin{subfigure}[thpb]{0.45\textwidth}
%
%
\begin{psfrags}%
\psfragscanon%
\newcommand{\tsize}{0.8}
\newcommand{\tsizeb}{0.5}
%
\psfrag{A}[t][t][\tsize]{\color[rgb]{0,0,0}\setlength{\tabcolsep}{0pt}\begin{tabular}{c}Scene~1\end{tabular}}%
\psfrag{B}[t][t][\tsize]{\color[rgb]{0,0,0}\setlength{\tabcolsep}{0pt}\begin{tabular}{c}Scene~8\end{tabular}}%
\psfrag{C}[t][t][\tsize]{\color[rgb]{0,0,0}\setlength{\tabcolsep}{0pt}\begin{tabular}{c}Scene~9\end{tabular}}%
\psfrag{J}[t][t][\tsize]{\color[rgb]{0,0,0}\setlength{\tabcolsep}{0pt}\begin{tabular}{c}$L_1$\end{tabular}}%
\psfrag{L}[t][t][\tsize]{\color[rgb]{0,0,0}\setlength{\tabcolsep}{0pt}\begin{tabular}{c}$L_8$\end{tabular}}%
\psfrag{M}[t][t][\tsize]{\color[rgb]{0,0,0}\setlength{\tabcolsep}{0pt}\begin{tabular}{c}$T$\end{tabular}}%
%
\includegraphics[width=\textwidth]{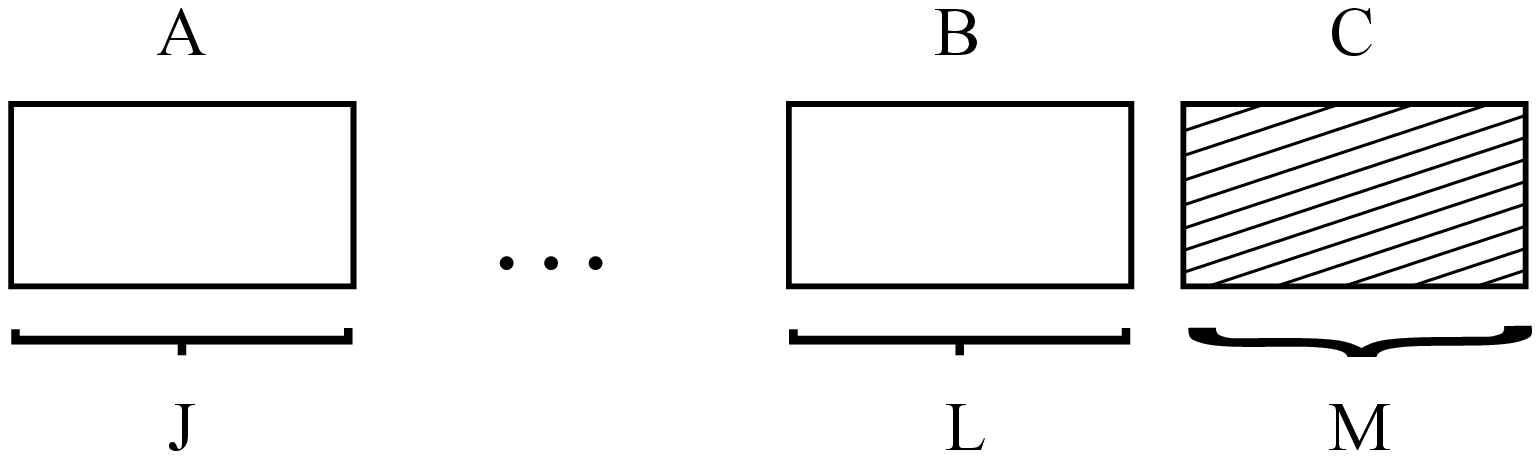}
\end{psfrags}%
%

		\caption{K-fold cross-validation 2}
		\label{fig:KFold2}
	\end{subfigure}
	\caption{Two K-fold cross-validations for evaluation of learning and exploration capabilities of the classifier.}
	\label{fig:KFoldMethods}
\end{figure}

In order to evaluate the performance of our classification models in an unknown scene or for an untrained dataset, we perform two K-fold cross-validations on multiple real scenes in both outdoor and indoor environments presented in Fig.~\ref{fig:Dataset}. The border color of each scene in the figure was used in plotting the results to distinguish different scenes. Fig.~\ref{fig:KFoldMethods} shows the K-fold tests where $L$ indicates the learning sets and $T$ represents the test sets. 

In the first K-fold validation, $8/9$ ratio of samples from each scene were randomly selected and used for training ($L_1+L_2+\ldots+L_9$) while the remaining $1/9$ ratio of samples were tested ($T_1,T_2,\ldots,T_9$). We iterated this process 9 times so that all data has been used as test set. We performed this test to evaluate the learning capability of the proposed SSL setup in which multiple scenes with different variation of appearance are included in the training set. 

In the second K-fold validation, we tested the exploration performance of our SSL setup. This can be done by using datasets from 8 out of 9 scenes for training ($L_1+L_2+\ldots+L_8$) and testing on the remaining dataset ($T_9$). The remaining dataset can be completely different or slightly similar to the training set. We iterated the process until all scenes were tested. Therefore, in total we performed 9 tests for each K-fold cross-validation in both outdoor and indoor environment and surface roughness $\epsilon^*$ was used as the ground truth in the validation. 

\begin{figure}[thpb]
	\captionsetup[subfigure]{justification=centering}
	\centering
	\begin{subfigure}[thpb]{0.45\textwidth}
%
%
\begin{psfrags}%
\psfragscanon%
\newcommand{\tsize}{0.7}
\newcommand{\tsizeb}{0.65}
%
\psfrag{s02}[b][b][\tsize]{\color[rgb]{0,0,0}\setlength{\tabcolsep}{0pt}\begin{tabular}{c}\end{tabular}}%
\psfrag{s03}[t][t][\tsize]{\color[rgb]{0,0,0}\setlength{\tabcolsep}{0pt}\begin{tabular}{c}False~Positive~Rate\end{tabular}}%
\psfrag{s04}[b][b][\tsize]{\color[rgb]{0,0,0}\setlength{\tabcolsep}{0pt}\begin{tabular}{c}True~Positive~Rate\end{tabular}}%
%
\psfrag{x01}[t][t][\tsize]{0}%
\psfrag{x02}[t][t][\tsize]{0.1}%
\psfrag{x03}[t][t][\tsize]{0.2}%
\psfrag{x04}[t][t][\tsize]{0.3}%
\psfrag{x05}[t][t][\tsize]{0.4}%
\psfrag{x06}[t][t][\tsize]{0.5}%
\psfrag{x07}[t][t][\tsize]{0.6}%
\psfrag{x08}[t][t][\tsize]{0.7}%
\psfrag{x09}[t][t][\tsize]{0.8}%
\psfrag{x10}[t][t][\tsize]{0.9}%
\psfrag{x11}[t][t][\tsize]{1}%
%
\psfrag{v01}[r][r][\tsize]{0}%
\psfrag{v02}[r][r][\tsize]{0.2}%
\psfrag{v03}[r][r][\tsize]{0.4}%
\psfrag{v04}[r][r][\tsize]{0.6}%
\psfrag{v05}[r][r][\tsize]{0.8}%
\psfrag{v06}[r][r][\tsize]{1}%
%
\includegraphics[width=\textwidth]{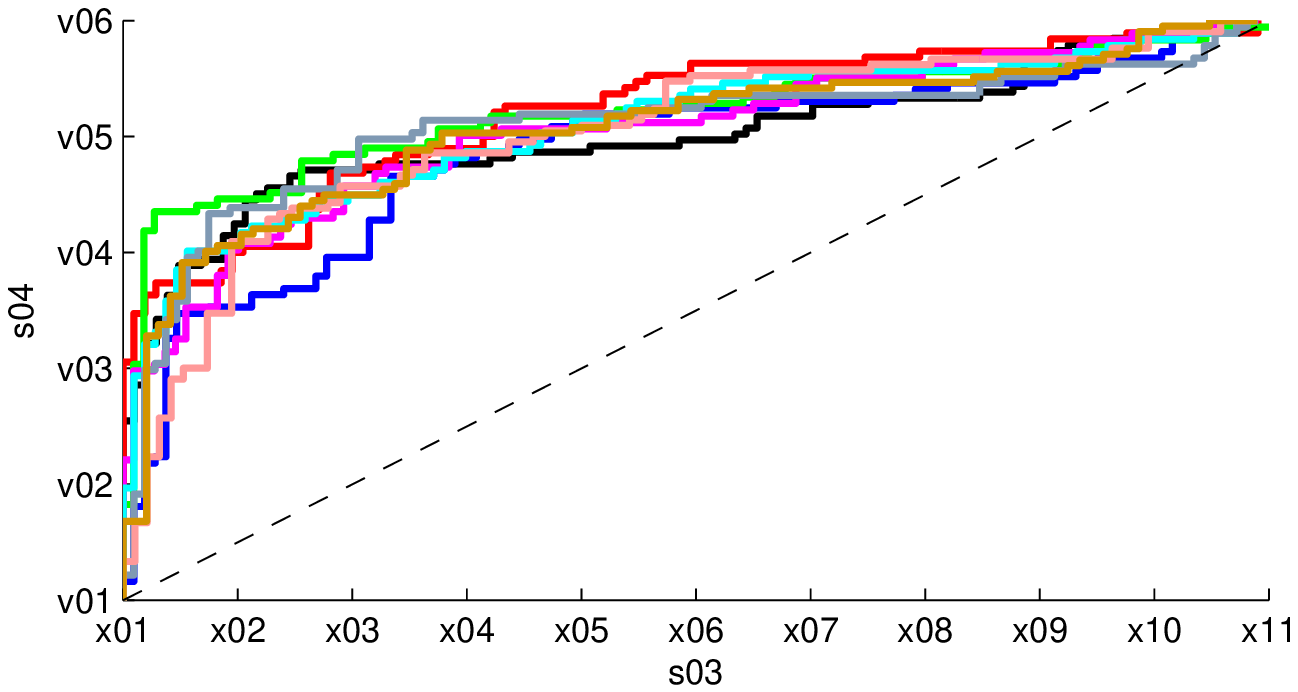}
\end{psfrags}%
%

		\label{fig:ROCOutdoorRand}
	\end{subfigure} \quad
	\begin{subfigure}[thpb]{0.45\textwidth}
%
%
\begin{psfrags}%
\psfragscanon%
\newcommand{\tsize}{0.7}
\newcommand{\tsizeb}{0.65}
%
\psfrag{s02}[b][b][\tsize]{\color[rgb]{0,0,0}\setlength{\tabcolsep}{0pt}\begin{tabular}{c}\end{tabular}}%
\psfrag{s03}[t][t][\tsize]{\color[rgb]{0,0,0}\setlength{\tabcolsep}{0pt}\begin{tabular}{c}False~Positive~Rate\end{tabular}}%
\psfrag{s04}[b][b][\tsize]{\color[rgb]{0,0,0}\setlength{\tabcolsep}{0pt}\begin{tabular}{c}True~Positive~Rate\end{tabular}}%
%
\psfrag{x01}[t][t][\tsize]{0}%
\psfrag{x02}[t][t][\tsize]{0.1}%
\psfrag{x03}[t][t][\tsize]{0.2}%
\psfrag{x04}[t][t][\tsize]{0.3}%
\psfrag{x05}[t][t][\tsize]{0.4}%
\psfrag{x06}[t][t][\tsize]{0.5}%
\psfrag{x07}[t][t][\tsize]{0.6}%
\psfrag{x08}[t][t][\tsize]{0.7}%
\psfrag{x09}[t][t][\tsize]{0.8}%
\psfrag{x10}[t][t][\tsize]{0.9}%
\psfrag{x11}[t][t][\tsize]{1}%
%
\psfrag{v01}[r][r][\tsize]{0}%
\psfrag{v02}[r][r][\tsize]{0.2}%
\psfrag{v03}[r][r][\tsize]{0.4}%
\psfrag{v04}[r][r][\tsize]{0.6}%
\psfrag{v05}[r][r][\tsize]{0.8}%
\psfrag{v06}[r][r][\tsize]{1}%
%
\includegraphics[width=\textwidth]{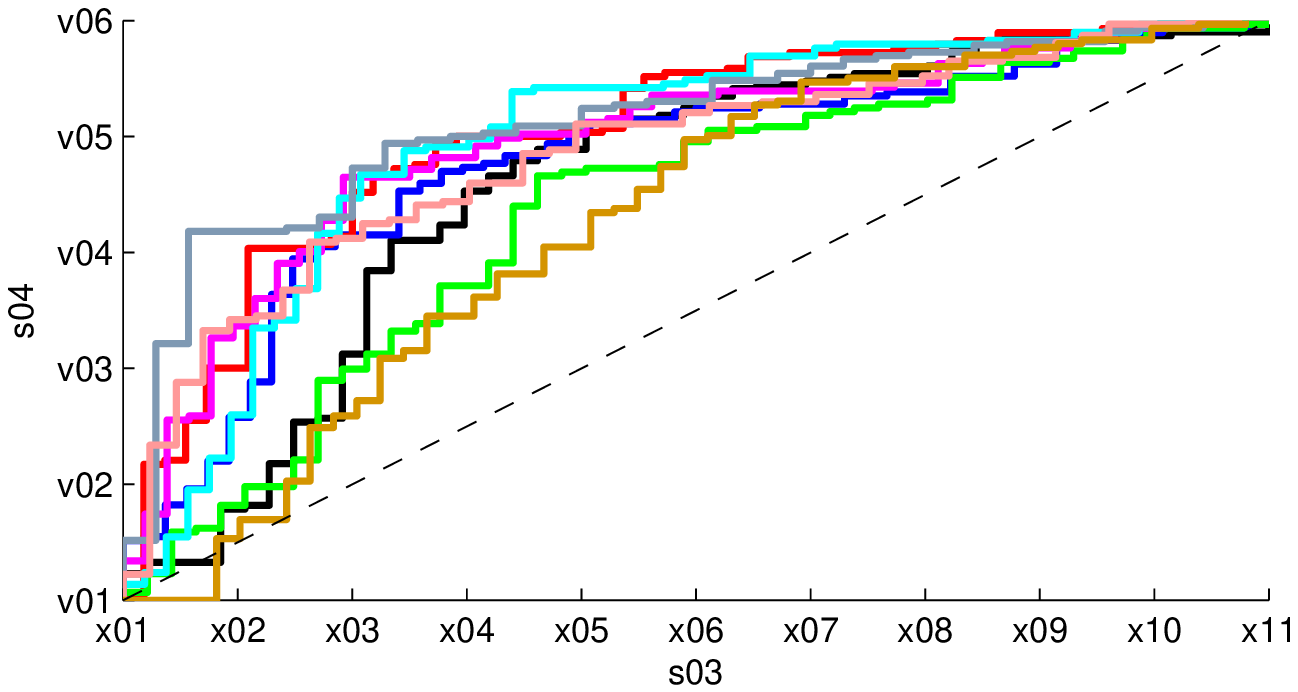}
\end{psfrags}%
%

		\label{fig:ROCIndoorRand}
	\end{subfigure} \\
	\begin{subfigure}[thpb]{0.45\textwidth}
%
%
\begin{psfrags}%
\psfragscanon%
\newcommand{\tsize}{0.7}
\newcommand{\tsizeb}{0.65}
%
\psfrag{H}[b][b][\tsize]{\color[rgb]{0,0,0}\setlength{\tabcolsep}{0pt}\begin{tabular}{c}H\end{tabular}}%
%
\psfrag{S1}[t][c][\tsize]{T1}%
\psfrag{S2}[t][c][\tsize]{T2}%
\psfrag{S3}[t][c][\tsize]{T3}%
\psfrag{S4}[t][c][\tsize]{T4}%
\psfrag{S5}[t][c][\tsize]{T5}%
\psfrag{S6}[t][c][\tsize]{T6}%
\psfrag{S7}[t][c][\tsize]{T7}%
\psfrag{S8}[t][c][\tsize]{T8}%
\psfrag{S9}[t][c][\tsize]{T9}%
%
\psfrag{0}[r][r][\tsize]{0}%
\psfrag{0.1}[r][r][\tsize]{}%
\psfrag{0.2}[r][r][\tsize]{0.2}%
\psfrag{0.3}[r][r][\tsize]{}%
\psfrag{0.4}[r][r][\tsize]{0.4}%
\psfrag{0.5}[r][r][\tsize]{}%
\psfrag{0.6}[r][r][\tsize]{0.6}%
\psfrag{0.7}[r][r][\tsize]{}%
\psfrag{0.8}[r][r][\tsize]{0.8}%
\psfrag{0.9}[r][r][\tsize]{}%
\psfrag{1}[r][r][\tsize]{1}%

%
\includegraphics[width=\textwidth]{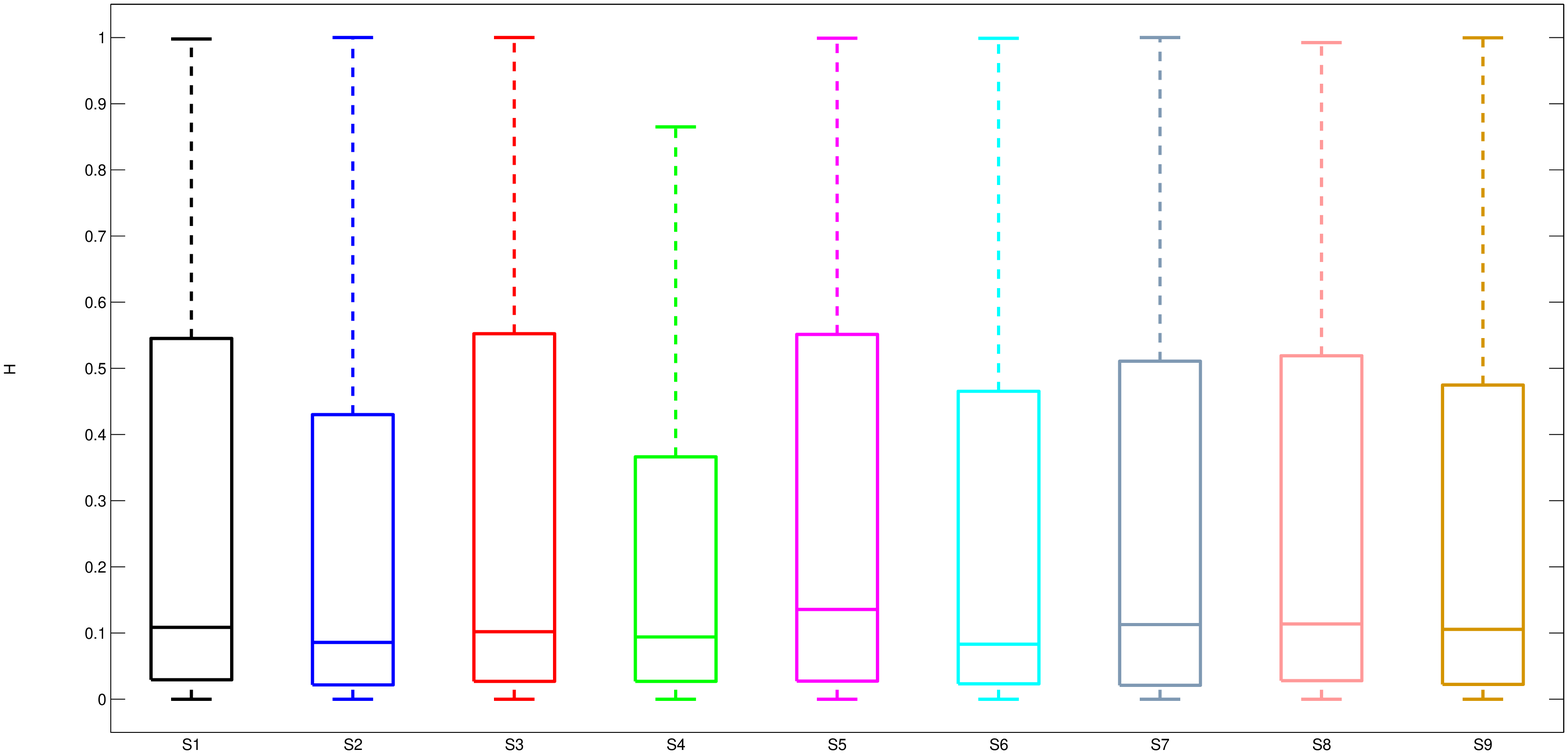}
\end{psfrags}%
%

		\caption{Outdoor scenes.}
		\label{fig:EntropyOutdoorRand}
	\end{subfigure} \quad
	\begin{subfigure}[thpb]{0.45\textwidth}
%
%
\begin{psfrags}%
\psfragscanon%
\newcommand{\tsize}{0.7}
\newcommand{\tsizeb}{0.65}
%
\psfrag{H}[b][b][\tsize]{\color[rgb]{0,0,0}\setlength{\tabcolsep}{0pt}\begin{tabular}{c}H\end{tabular}}%
%
\psfrag{S1}[t][c][\tsize]{T1}%
\psfrag{S2}[t][c][\tsize]{T2}%
\psfrag{S3}[t][c][\tsize]{T3}%
\psfrag{S4}[t][c][\tsize]{T4}%
\psfrag{S5}[t][c][\tsize]{T5}%
\psfrag{S6}[t][c][\tsize]{T6}%
\psfrag{S7}[t][c][\tsize]{T7}%
\psfrag{S8}[t][c][\tsize]{T8}%
\psfrag{S9}[t][c][\tsize]{T9}%
%
\psfrag{0}[r][r][\tsize]{0}%
\psfrag{0.1}[r][r][\tsize]{}%
\psfrag{0.2}[r][r][\tsize]{0.2}%
\psfrag{0.3}[r][r][\tsize]{}%
\psfrag{0.4}[r][r][\tsize]{0.4}%
\psfrag{0.5}[r][r][\tsize]{}%
\psfrag{0.6}[r][r][\tsize]{0.6}%
\psfrag{0.7}[r][r][\tsize]{}%
\psfrag{0.8}[r][r][\tsize]{0.8}%
\psfrag{0.9}[r][r][\tsize]{}%
\psfrag{1}[r][r][\tsize]{1}%

%
\includegraphics[width=\textwidth]{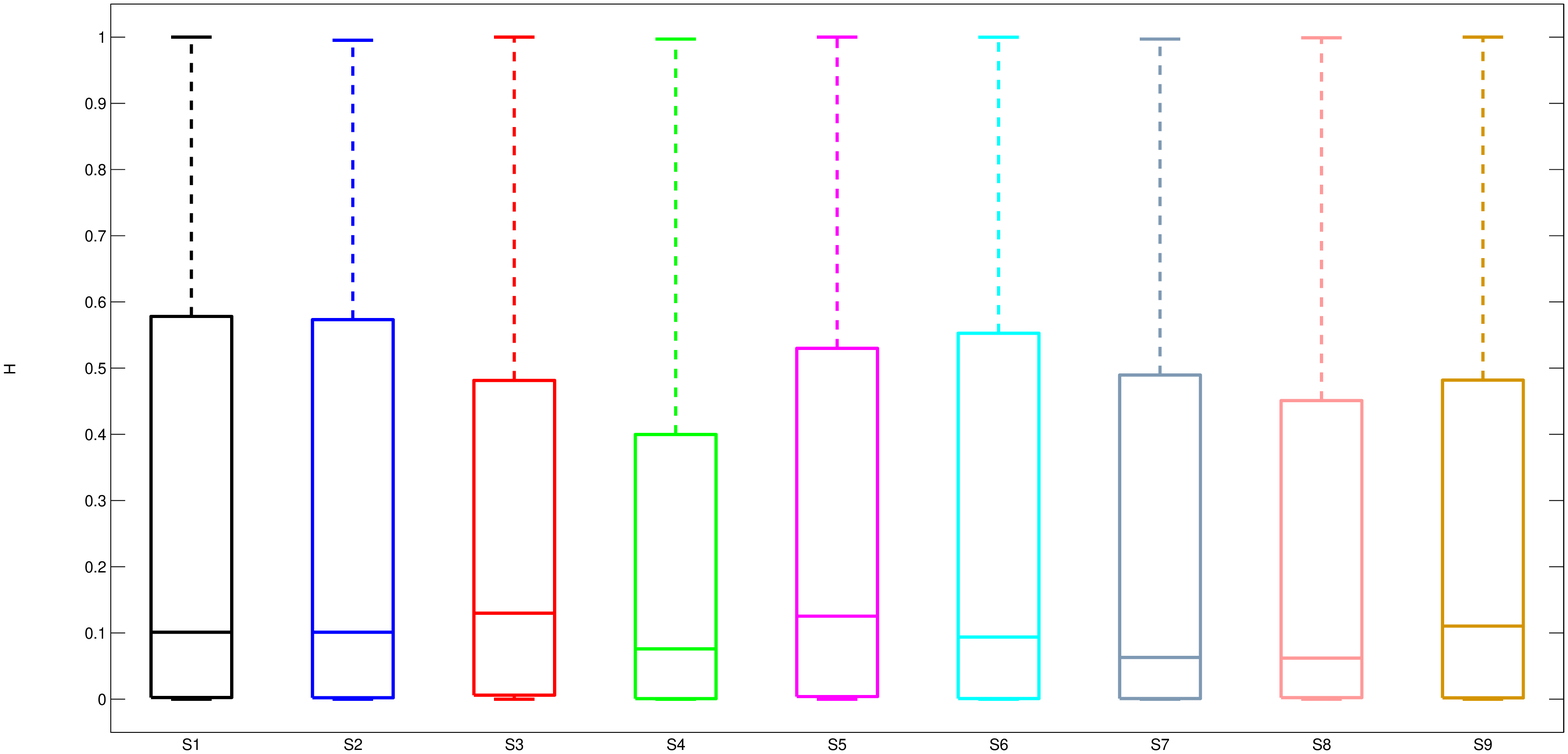}
\end{psfrags}%
%

		\caption{Indoor scenes.}
		\label{fig:EntropyIndoorRand}
	\end{subfigure}			
	\caption{ROC curves and uncertainty measures for K-fold cross-validations 1. The results are given the color of the test set.}
	\label{fig:ROCRand}
\end{figure}

\begin{figure}[thpb]
	\captionsetup[subfigure]{justification=centering}
	\centering
	\begin{subfigure}[thpb]{0.45\textwidth}
%
%
\begin{psfrags}%
\psfragscanon%
\newcommand{\tsize}{0.7}
\newcommand{\tsizeb}{0.65}
%
\psfrag{s02}[b][b][\tsize]{\color[rgb]{0,0,0}\setlength{\tabcolsep}{0pt}\begin{tabular}{c}\end{tabular}}%
\psfrag{s03}[t][t][\tsize]{\color[rgb]{0,0,0}\setlength{\tabcolsep}{0pt}\begin{tabular}{c}False~Positive~Rate\end{tabular}}%
\psfrag{s04}[b][b][\tsize]{\color[rgb]{0,0,0}\setlength{\tabcolsep}{0pt}\begin{tabular}{c}True~Positive~Rate\end{tabular}}%
%
\psfrag{x01}[t][t][\tsize]{0}%
\psfrag{x02}[t][t][\tsize]{0.1}%
\psfrag{x03}[t][t][\tsize]{0.2}%
\psfrag{x04}[t][t][\tsize]{0.3}%
\psfrag{x05}[t][t][\tsize]{0.4}%
\psfrag{x06}[t][t][\tsize]{0.5}%
\psfrag{x07}[t][t][\tsize]{0.6}%
\psfrag{x08}[t][t][\tsize]{0.7}%
\psfrag{x09}[t][t][\tsize]{0.8}%
\psfrag{x10}[t][t][\tsize]{0.9}%
\psfrag{x11}[t][t][\tsize]{1}%
%
\psfrag{v01}[r][r][\tsize]{0}%
\psfrag{v02}[r][r][\tsize]{0.2}%
\psfrag{v03}[r][r][\tsize]{0.4}%
\psfrag{v04}[r][r][\tsize]{0.6}%
\psfrag{v05}[r][r][\tsize]{0.8}%
\psfrag{v06}[r][r][\tsize]{1}%
%
\includegraphics[width=\textwidth]{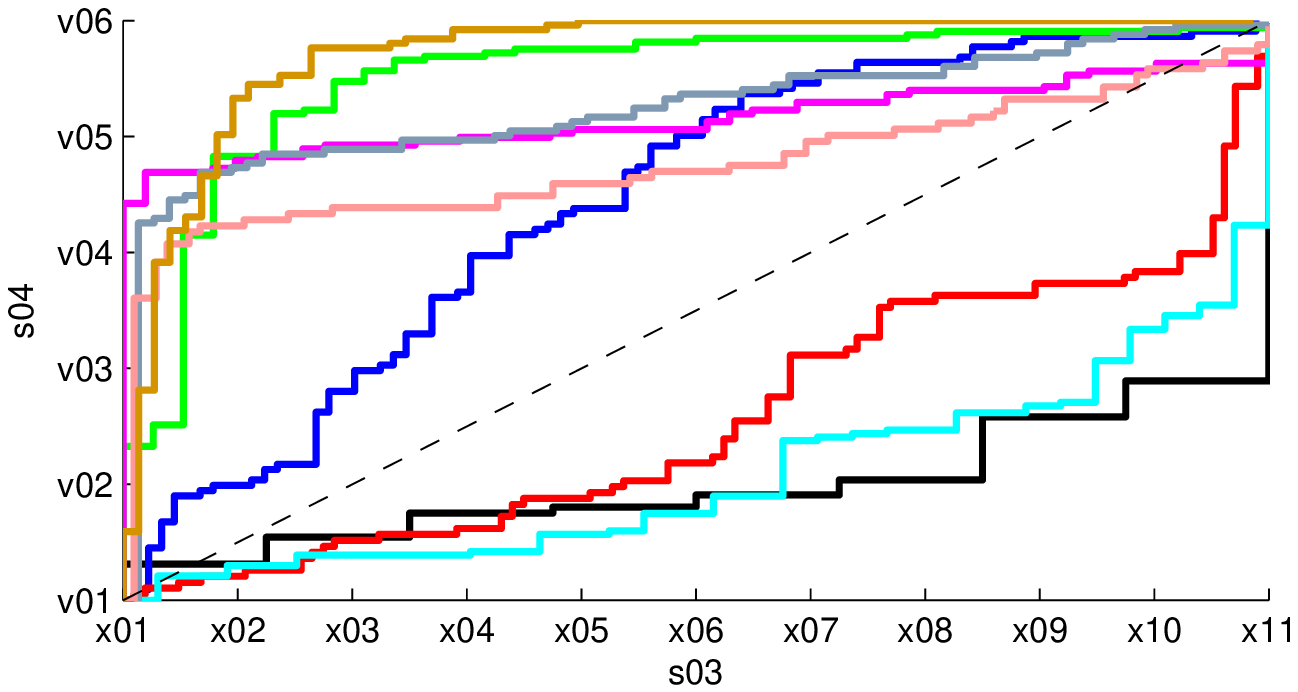}
\end{psfrags}%
%

		\label{fig:ROCOutdoor}
	\end{subfigure} \quad
	\begin{subfigure}[thpb]{0.45\textwidth}
%
%
\begin{psfrags}%
\psfragscanon%
\newcommand{\tsize}{0.7}
\newcommand{\tsizeb}{0.65}
%
\psfrag{s02}[b][b][\tsize]{\color[rgb]{0,0,0}\setlength{\tabcolsep}{0pt}\begin{tabular}{c}\end{tabular}}%
\psfrag{s03}[t][t][\tsize][\tsize]{\color[rgb]{0,0,0}\setlength{\tabcolsep}{0pt}\begin{tabular}{c}False~Positive~Rate\end{tabular}}%
\psfrag{s04}[b][b][\tsize]{\color[rgb]{0,0,0}\setlength{\tabcolsep}{0pt}\begin{tabular}{c}True~Positive~Rate\end{tabular}}%
%
\psfrag{x01}[t][t][\tsize]{0}%
\psfrag{x02}[t][t][\tsize]{0.1}%
\psfrag{x03}[t][t][\tsize]{0.2}%
\psfrag{x04}[t][t][\tsize]{0.3}%
\psfrag{x05}[t][t][\tsize]{0.4}%
\psfrag{x06}[t][t][\tsize]{0.5}%
\psfrag{x07}[t][t][\tsize]{0.6}%
\psfrag{x08}[t][t][\tsize]{0.7}%
\psfrag{x09}[t][t][\tsize]{0.8}%
\psfrag{x10}[t][t][\tsize]{0.9}%
\psfrag{x11}[t][t][\tsize]{1}%
%
\psfrag{v01}[r][r][\tsize]{0}%
\psfrag{v02}[r][r][\tsize]{0.2}%
\psfrag{v03}[r][r][\tsize]{0.4}%
\psfrag{v04}[r][r][\tsize]{0.6}%
\psfrag{v05}[r][r][\tsize]{0.8}%
\psfrag{v06}[r][r][\tsize]{1}%
%
\includegraphics[width=\textwidth]{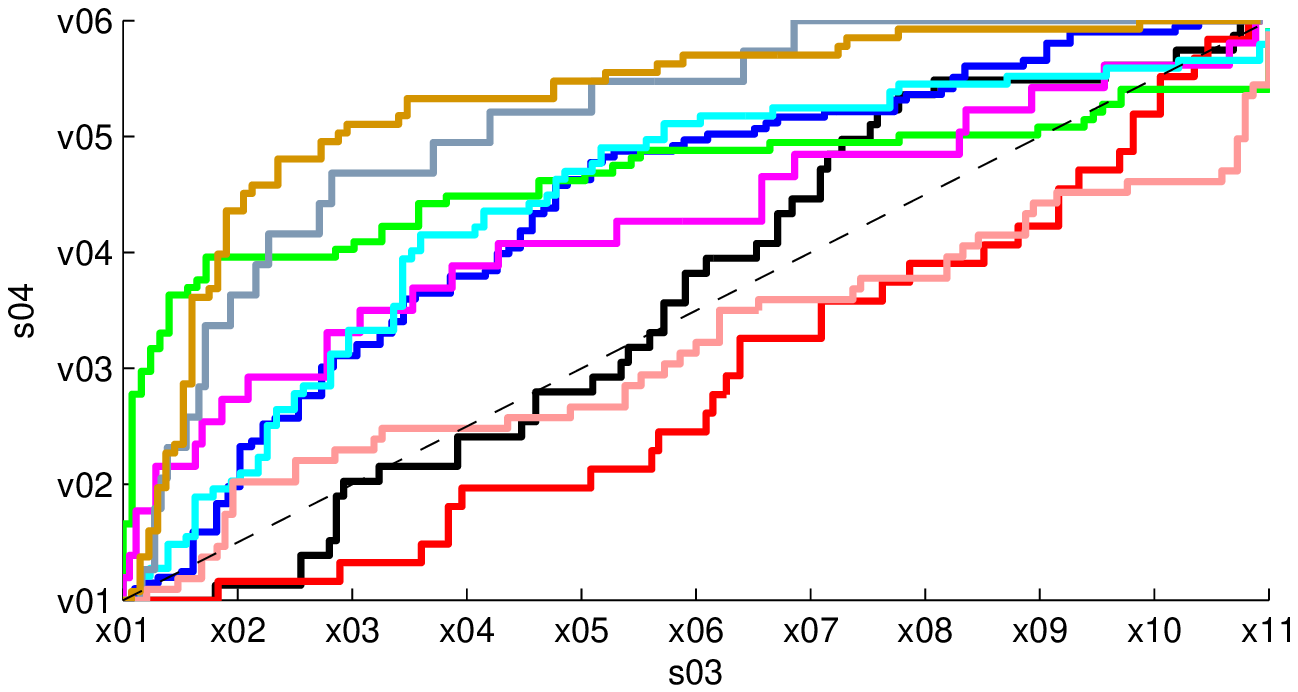}
\end{psfrags}%
%

		\label{fig:ROCIndoor}
	\end{subfigure}\\
	\begin{subfigure}[thpb]{0.45\textwidth}
%
%
\begin{psfrags}%
\psfragscanon%
\newcommand{\tsize}{0.7}
\newcommand{\tsizeb}{0.65}
%
\psfrag{H}[b][b][\tsize]{\color[rgb]{0,0,0}\setlength{\tabcolsep}{0pt}\begin{tabular}{c}H\end{tabular}}%
%
\psfrag{S1}[t][c][\tsize]{T1}%
\psfrag{S2}[t][c][\tsize]{T2}%
\psfrag{S3}[t][c][\tsize]{T3}%
\psfrag{S4}[t][c][\tsize]{T4}%
\psfrag{S5}[t][c][\tsize]{T5}%
\psfrag{S6}[t][c][\tsize]{T6}%
\psfrag{S7}[t][c][\tsize]{T7}%
\psfrag{S8}[t][c][\tsize]{T8}%
\psfrag{S9}[t][c][\tsize]{T9}%
%
\psfrag{0}[r][r][\tsize]{0}%
\psfrag{0.1}[r][r][\tsize]{}%
\psfrag{0.2}[r][r][\tsize]{0.2}%
\psfrag{0.3}[r][r][\tsize]{}%
\psfrag{0.4}[r][r][\tsize]{0.4}%
\psfrag{0.5}[r][r][\tsize]{}%
\psfrag{0.6}[r][r][\tsize]{0.6}%
\psfrag{0.7}[r][r][\tsize]{}%
\psfrag{0.8}[r][r][\tsize]{0.8}%
\psfrag{0.9}[r][r][\tsize]{}%
\psfrag{1}[r][r][\tsize]{1}%

%
\includegraphics[width=\textwidth]{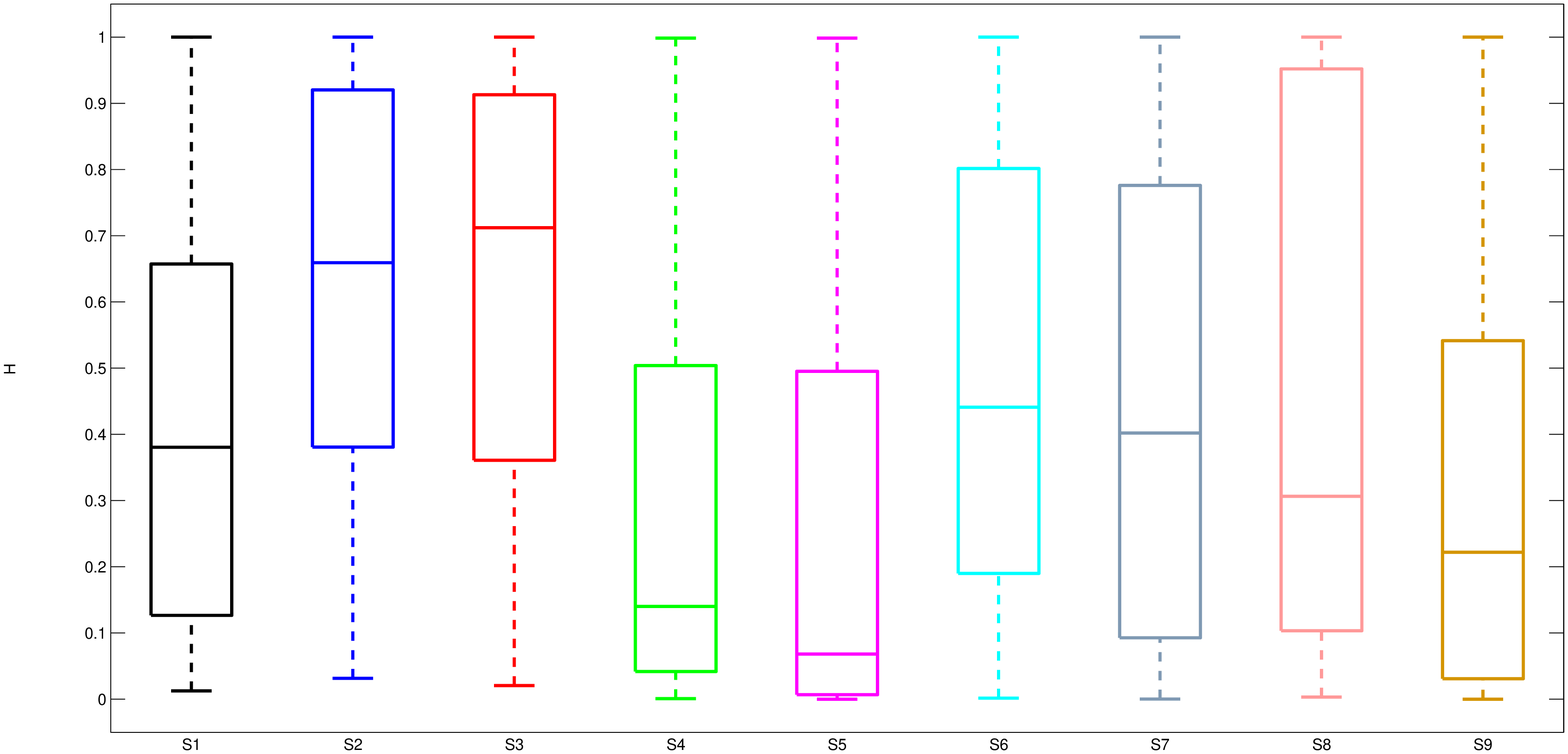}
\end{psfrags}%
%

		\caption{Outdoor scenes.}
		\label{fig:EntropyOutdoor}
	\end{subfigure} \quad
	\begin{subfigure}[thpb]{0.45\textwidth}
%
%
\begin{psfrags}%
\psfragscanon%
\newcommand{\tsize}{0.7}
\newcommand{\tsizeb}{0.65}
%
\psfrag{H}[b][b][\tsize]{\color[rgb]{0,0,0}\setlength{\tabcolsep}{0pt}\begin{tabular}{c}H\end{tabular}}%
%
\psfrag{S1}[t][c][\tsize]{T1}%
\psfrag{S2}[t][c][\tsize]{T2}%
\psfrag{S3}[t][c][\tsize]{T3}%
\psfrag{S4}[t][c][\tsize]{T4}%
\psfrag{S5}[t][c][\tsize]{T5}%
\psfrag{S6}[t][c][\tsize]{T6}%
\psfrag{S7}[t][c][\tsize]{T7}%
\psfrag{S8}[t][c][\tsize]{T8}%
\psfrag{S9}[t][c][\tsize]{T9}%
%
\psfrag{0}[r][r][\tsize]{0}%
\psfrag{0.1}[r][r][\tsize]{}%
\psfrag{0.2}[r][r][\tsize]{0.2}%
\psfrag{0.3}[r][r][\tsize]{}%
\psfrag{0.4}[r][r][\tsize]{0.4}%
\psfrag{0.5}[r][r][\tsize]{}%
\psfrag{0.6}[r][r][\tsize]{0.6}%
\psfrag{0.7}[r][r][\tsize]{}%
\psfrag{0.8}[r][r][\tsize]{0.8}%
\psfrag{0.9}[r][r][\tsize]{}%
\psfrag{1}[r][r][\tsize]{1}%

%
\includegraphics[width=\textwidth]{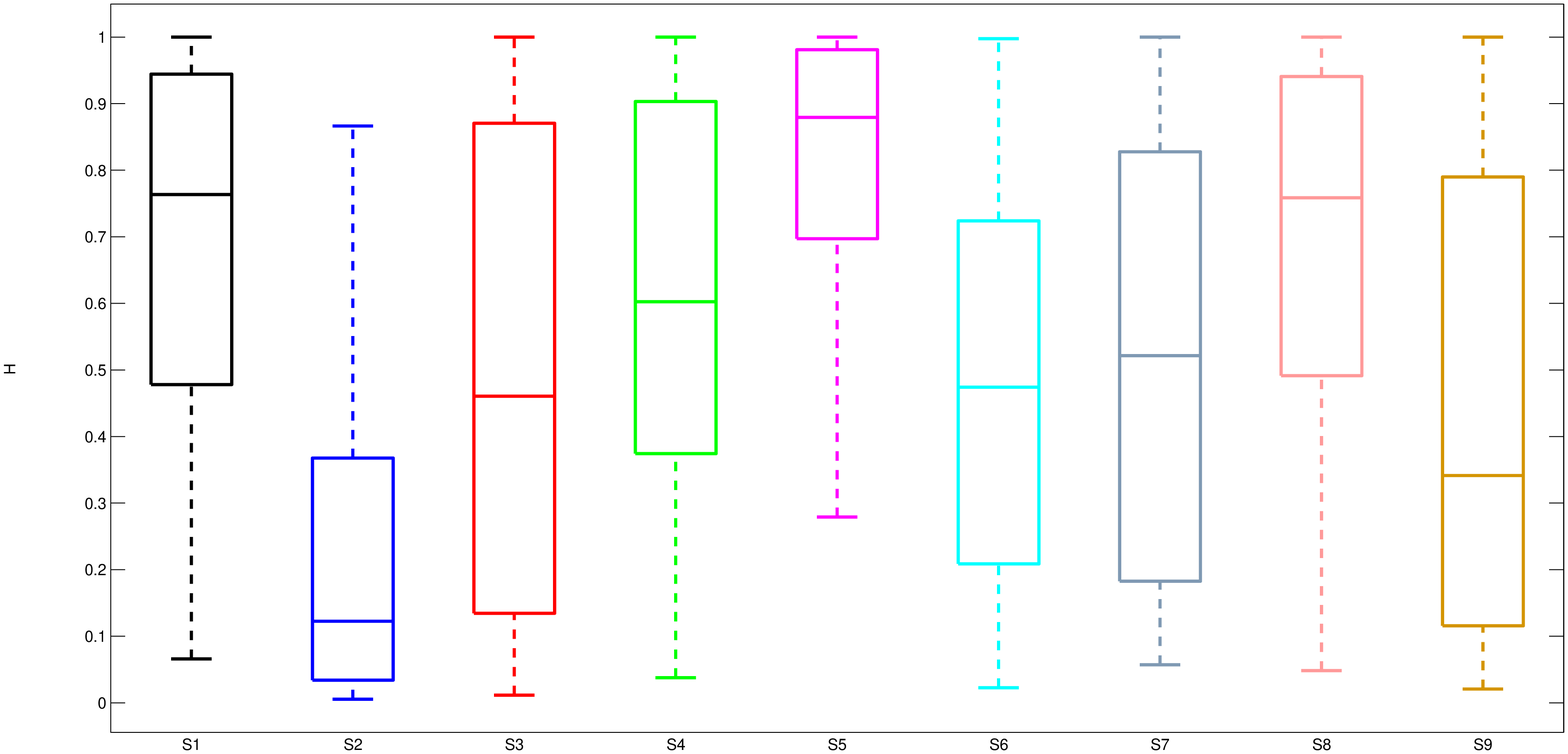}
\end{psfrags}%
%

		\caption{Indoor scenes.}
		\label{fig:EntropyIndoor}
	\end{subfigure}
	\caption{ROC curves and uncertainty measures for K-fold cross-validations 2. The results are given the color of the test scene (see Fig.~\ref{fig:Dataset}).}
	\label{fig:ROC}
\end{figure}

To illustrate the performance of the classifier, we plot a receiver operating characteristic (ROC) curve for each K-fold test in Fig.~\ref{fig:ROCRand} and Fig.~\ref{fig:ROC}. In addition, the uncertainty measure from Shannon entropy for the corresponding test is also presented. TABLE~\ref{table:AUC} shows the Area Under ROC Curve ($AUC$) and NRMSE for each test which is a common metric used to evaluate the ROC curves and regression tests, respectively. 

\begin{table*}[thpb]
	\caption{Performance measure of ROC curves using Area Under the Curve (AUC - higher is better) and NRMSE (\% - lower is better)}
	\label{table:AUC}
	\begin{center}
		\resizebox{\textwidth}{!}{%
			\begin{tabular}{|c|c|c||c|c|c|c|c|c|c|c|c|}
				\hline
				K-fold  & Environment & Metrics & T1 & T2 & T3 & T4 & T5 & T6 & T7 & T8 & T9 \\
				\hline
				\hline
				1 & Outdoor & AUC & 0.79 & 0.77 & 0.83 & 0.84 & 0.80 & 0.81 & 0.81 & 0.80 & 0.80 \\
				\hline
				&  & NRMSE & 15.4 & 17.5 & 16.4 & 16.1 & 16.2 & 18.9 & 14.0 & 14.4 & 12.9\\
				\hline
				& Indoor & AUC  & 0.72 & 0.75 & 0.81 & 0.67 & 0.77 & 0.78 & 0.80 & 0.75 & 0.64 \\
				\hline
				&  & NRMSE & 13.0 & 13.6 & 17.7 & 14.8 & 12.8 & 15.2 & 15.9 & 13.9 & 13.4\\
				\hline
				\hline
				2 & Outdoor & AUC & 0.23 & 0.68 & 0.31 & 0.90 & 0.83 & 0.22 & 0.85 & 0.75 & 0.93\\
				\hline
				&  & NRMSE & 84.2 & 87.2 & 43.8 & 22.9 & 33.2 & 27.0 & 21.8 & 52.9 & 18.1\\
				\hline
				& Indoor & AUC & 0.52 & 0.68 & 0.37 & 0.73 & 0.65 & 0.69 & 0.82 & 0.44 & 0.85 \\
				\hline
				&  & NRMSE & 30.0 & 29.1 & 34.8 & 22.2 & 83.3 & 24.1 & 28.9 & 72.1 & 27.6\\
				\hline
		\end{tabular}}
	\end{center}
\end{table*}

\begin{figure}[!h]
	\centering
	\captionsetup{justification=centering}
%
%
\begin{psfrags}%
	\psfragscanon%
	\newcommand{\tsize}{0.7}
	\newcommand{\tsizeb}{0.65}
	%
	\psfrag{s13}[b][b][\tsize]{\color[rgb]{0,0,0}\setlength{\tabcolsep}{0pt}\begin{tabular}{c}\Large$\epsilon^*$,~\Large$\hat{\epsilon}$\end{tabular}}%
	\psfrag{s18}[][]{\color[rgb]{0,0,0}\setlength{\tabcolsep}{0pt}\begin{tabular}{c} \end{tabular}}%
	\psfrag{s19}[][]{\color[rgb]{0,0,0}\setlength{\tabcolsep}{0pt}\begin{tabular}{c} \end{tabular}}%
	\psfrag{s20}[l][l][\tsize]{\color[rgb]{0,0,0}$\hat{\epsilon}$}%
	\psfrag{s21}[l][l][\tsize]{\color[rgb]{0,0,0}$\epsilon^*$}%
	\psfrag{s22}[l][l][\tsize]{\color[rgb]{0,0,0}$\hat{\epsilon}$}%
	\psfrag{s23}[b][b][\tsize]{\color[rgb]{0,0,0}\setlength{\tabcolsep}{0pt}\begin{tabular}{c}H\end{tabular}}%
	\psfrag{s24}[b][b][\tsize]{\color[rgb]{0,0,0}\setlength{\tabcolsep}{0pt}\begin{tabular}{c}$l_1$\end{tabular}}%
	\psfrag{s25}[t][t][\tsize]{\color[rgb]{0,0,0}\setlength{\tabcolsep}{0pt}\begin{tabular}{c}Time~(s)\end{tabular}}%
	\psfrag{s26}[b][b][\tsize]{\color[rgb]{0,0,0}\setlength{\tabcolsep}{0pt}\begin{tabular}{c}$l_2$\end{tabular}}%
	%
	\psfrag{x01}[t][t][\tsize]{0}%
	\psfrag{x02}[t][t][\tsize]{2}%
	\psfrag{x03}[t][t][\tsize]{4}%
	\psfrag{x04}[t][t][\tsize]{6}%
	\psfrag{x05}[t][t][\tsize]{8}%
	\psfrag{x06}[t][t][\tsize]{10}%
	\psfrag{x07}[t][t][\tsize]{12}%
	\psfrag{x08}[t][t][\tsize]{14}%
	\psfrag{x09}[t][t][\tsize]{16}%
	\psfrag{x10}[t][t][\tsize]{18}%
	\psfrag{x11}[t][t][\tsize]{20}%
	%
	\psfrag{v01}[r][r][\tsize]{}%
	\psfrag{v02}[r][r][\tsize]{}%
	\psfrag{v03}[r][r][\tsize]{}%
	\psfrag{v04}[r][r][\tsize]{}%
	\psfrag{v05}[r][r][\tsize]{}%
	\psfrag{v06}[r][r][\tsize]{}%
	\psfrag{v07}[r][r][\tsize]{0}%
	\psfrag{v08}[r][r][\tsize]{}%
	\psfrag{v09}[r][r][\tsize]{1}%
	\psfrag{v10}[r][r][\tsize]{}%
	\psfrag{v11}[r][r][\tsize]{}%
	\psfrag{v12}[r][r][\tsize]{}%
	\psfrag{v13}[r][r][\tsize]{0}%
	\psfrag{v14}[r][r][\tsize]{}%
	\psfrag{v15}[r][r][\tsize]{2}%
	\psfrag{v16}[r][r][\tsize]{}%
	\psfrag{v17}[r][r][\tsize]{4}%
	\psfrag{v18}[r][r][\tsize]{}%
	\psfrag{v19}[r][r][\tsize]{6}%
	%
	\includegraphics[trim = 0mm 0mm 3mm 0mm, clip, width=0.48\textwidth]{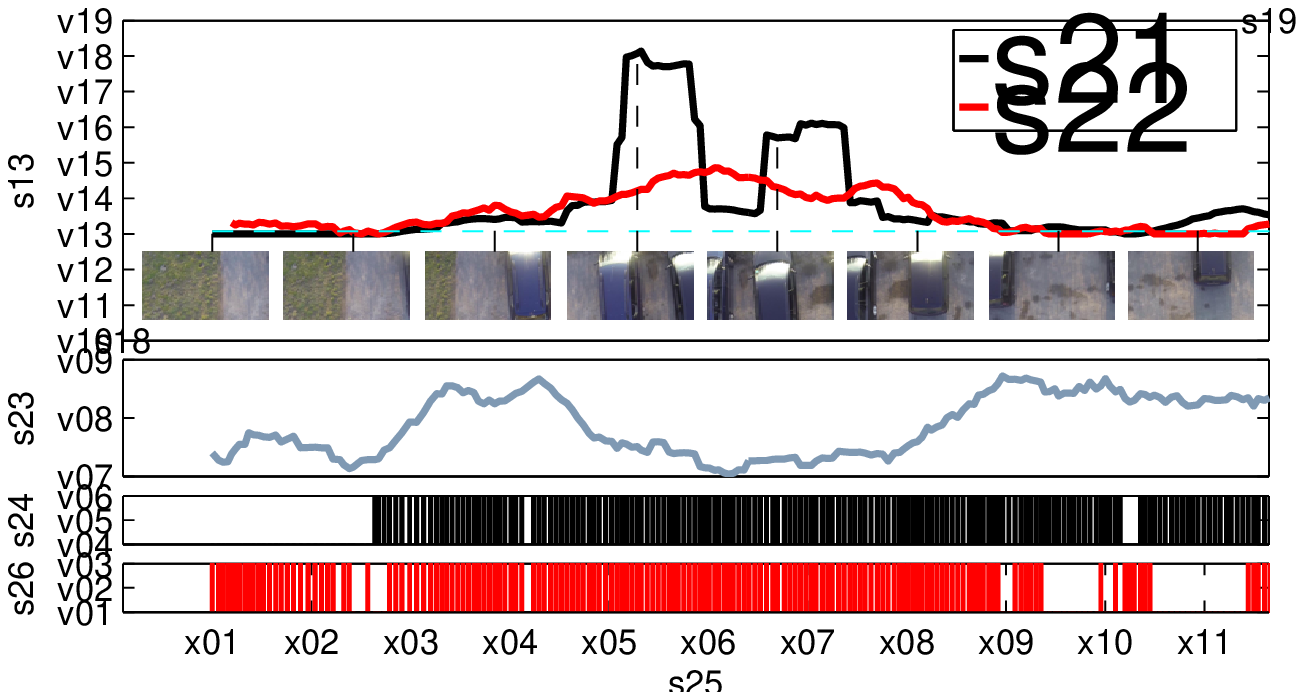}
\end{psfrags}%
%

	\caption{Classification of obstacles with $\epsilon^*$ ($l_1$) and $\hat{\epsilon}$ ($l_2$) in an untrained outdoor scene (S3), and the corresponding uncertainty measure, H. This example with a particular threshold of $\hat{\epsilon}_{th} = 0.076$ is chosen in the K-fold validation 2 which resulted in TP rate of $0.9$ and FP rate of $0.5$.}
	\label{fig:TP9_FP5_S3}
\end{figure}

\begin{figure}[!h]
	\centering
	\captionsetup{justification=centering}
%
%
\begin{psfrags}%
	\psfragscanon%
	\newcommand{\tsize}{0.7}
	\newcommand{\tsizeb}{0.65}
	%
	\psfrag{s13}[b][b][\tsize]{\color[rgb]{0,0,0}\setlength{\tabcolsep}{0pt}\begin{tabular}{c}\Large$\epsilon^*$,~\Large$\hat{\epsilon}$\end{tabular}}%
	\psfrag{s18}[][]{\color[rgb]{0,0,0}\setlength{\tabcolsep}{0pt}\begin{tabular}{c} \end{tabular}}%
	\psfrag{s19}[][]{\color[rgb]{0,0,0}\setlength{\tabcolsep}{0pt}\begin{tabular}{c} \end{tabular}}%
	\psfrag{s20}[l][l][\tsize]{\color[rgb]{0,0,0}$\hat{\epsilon}$}%
	\psfrag{s21}[l][l][\tsize]{\color[rgb]{0,0,0}$\epsilon^*$}%
	\psfrag{s22}[l][l][\tsize]{\color[rgb]{0,0,0}$\hat{\epsilon}$}%
	\psfrag{s23}[b][b][\tsize]{\color[rgb]{0,0,0}\setlength{\tabcolsep}{0pt}\begin{tabular}{c}H\end{tabular}}%
	\psfrag{s24}[b][b][\tsize]{\color[rgb]{0,0,0}\setlength{\tabcolsep}{0pt}\begin{tabular}{c}$l_1$\end{tabular}}%
	\psfrag{s25}[t][t][\tsize]{\color[rgb]{0,0,0}\setlength{\tabcolsep}{0pt}\begin{tabular}{c}Time~(s)\end{tabular}}%
	\psfrag{s26}[b][b][\tsize]{\color[rgb]{0,0,0}\setlength{\tabcolsep}{0pt}\begin{tabular}{c}$l_2$\end{tabular}}%
	%
	\psfrag{x01}[t][t][\tsize]{0}%
	\psfrag{x02}[t][t][\tsize]{2}%
	\psfrag{x03}[t][t][\tsize]{4}%
	\psfrag{x04}[t][t][\tsize]{6}%
	\psfrag{x05}[t][t][\tsize]{8}%
	\psfrag{x06}[t][t][\tsize]{10}%
	\psfrag{x07}[t][t][\tsize]{12}%
	\psfrag{x08}[t][t][\tsize]{14}%
	%
	\psfrag{v01}[r][r][\tsize]{}%
	\psfrag{v02}[r][r][\tsize]{}%
	\psfrag{v03}[r][r][\tsize]{}%
	\psfrag{v04}[r][r][\tsize]{}%
	\psfrag{v05}[r][r][\tsize]{}%
	\psfrag{v06}[r][r][\tsize]{}%
	\psfrag{v07}[r][r][\tsize]{0}%
	\psfrag{v08}[r][r][\tsize]{}%
	\psfrag{v09}[r][r][\tsize]{}%
	\psfrag{v10}[r][r][\tsize]{}%
	\psfrag{v11}[r][r][\tsize]{1}%
	\psfrag{v12}[r][r][\tsize]{}%
	\psfrag{v13}[r][r][\tsize]{}%
	\psfrag{v14}[r][r][\tsize]{}%
	\psfrag{v15}[r][r][\tsize]{0}%
	\psfrag{v16}[r][r][\tsize]{}%
	\psfrag{v17}[r][r][\tsize]{2}%
	\psfrag{v18}[r][r][\tsize]{}%
	\psfrag{v19}[r][r][\tsize]{4}%
	\psfrag{v20}[r][r][\tsize]{}%
	\psfrag{v21}[r][r][\tsize]{6}%
	%
	\includegraphics[trim = 0mm 0mm 3mm 0mm, clip, width=0.48\textwidth]{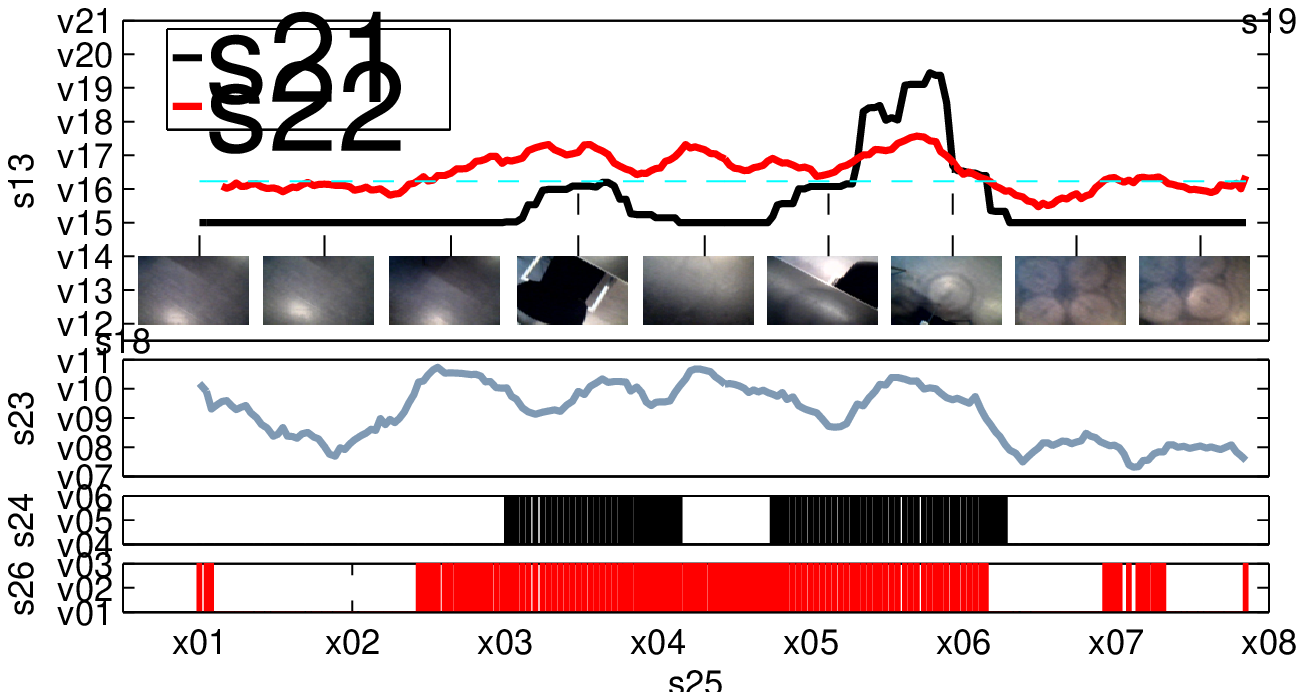}
\end{psfrags}%
%

	\caption{Classification of obstacles with $\epsilon^*$ ($l_1$) and $\hat{\epsilon}$ ($l_2$) in an untrained indoor scene (S3), and the corresponding uncertainty measure, H. This example with a particular threshold of $\hat{\epsilon}_{th} = 1.225$ is chosen in the K-fold validation 2 which resulted in TP rate of $0.9$ and FP rate of $0.5$.}
	\label{fig:TP9_FP5_S3_indoor}
\end{figure}

For landing, we would like to have a rather high true positive rate ($\approx0.9$) in order not to miss obstacles, while a rather high false positive rate may be acceptable - this will just reduce the available landing locations but will not endanger the MAVs. In Fig.~\ref{fig:ROCRand}, we can observe that to achieve a true positive rate of $\approx0.9$, we need to compromise with $\approx0.5$ false positive rate. Two examples in the second validation of untrained outdoor and indoor test sets with a chosen threshold result in TP rate of $\approx0.9$ and FP rate of $\approx0.5$ are shown in Fig.~\ref{fig:TP9_FP5_S3} and \ref{fig:TP9_FP5_S3_indoor}, respectively. In these figures, $l_1$ and $l_2$ indicate the time steps where the presence of obstacles are predicted using $\epsilon^*$ and $\hat{\epsilon}$, and $H$ is the uncertainty measure of $\hat{\epsilon}$. The results in these figures show that the obstacles can be detected, but many of the places considered safe by the optical flow algorithm are being classified by $\hat{\epsilon}$ as containing obstacles. In Fig.~\ref{fig:TP9_FP5_S3}, we can observe that the false positive cases happen mostly at at the beginning ($\approx0s-3s$) and also at images in which obstacles appear partly in the images ($\approx4s-6s$). If we look at the uncertainty measure, it shows an uncertain moment at the beginning and at the end of the image sequence. Similarly, in Fig.~\ref{fig:TP9_FP5_S3_indoor}, false positive happens at the beginning with high uncertainty measure. If we aim for a low $\hat{\epsilon}$ estimate and $H$, we can still find a suitable landing place though. 

The average $AUC$ for outdoor and indoor scenes in the first validation are $0.81$ and $0.74$, respectively. These results are rather good and indicate that the SSL method does not have difficulty of learning multiple scenes together, and it can predict scene with obstacles quite accurately if similar or part of the scene was learned. In Fig.~\ref{fig:EntropyIndoorRand}, it can be seen that the uncertainty measures are relatively small ($\approx 0.1$) indicating that the test environment is not too different from the training environment. Comparing outdoor and indoor environments, we can observe that learning of the indoor scenes is more difficult. This is due to the fact that indoor scenes are typically more complex than outdoor scenes. For instance, in an indoor scene, we can find many man-made objects and floors which look very different in terms of colors and textures, and thus the representation of these visual appearances can be more difficult. 


In the second validation, for completely unknown scenes, the performance of the classifier becomes worse. If we compare the results of NRMSE in TABLE~\ref{table:AUC}, we can observe that test errors are mostly larger in validation 2. Although some ROC curves are still acceptable (T4, T5, T7, and T9 outdoors for example), there are also ROC curves that are even poorer than random (e.g. T1, T3, T6 in outdoor scenes, and T1, T3, T8 in indoor scenes). However, the main observation is that the uncertainty measures for this K-fold test are much higher than for the first one, especially for those which have ROC curves worse than random. This indication allows the MAV to know when and where to re-train the regression model to adapt to new changes in its surroundings. 

\section{Flight Tests}
\label{sec:FlightTests}
\subsection{Experiment Platform}
\label{subsec:ExperimentPlatform}
\begin{figure}[thpb]
	\centering
	\captionsetup{justification=centering}
%
%
\begin{psfrags}%
\psfragscanon%
\newcommand{\tsize}{0.6}
\newcommand{\tsizeb}{0.65}
%
\psfrag{a}[t][t][\tsizeb]{\color[rgb]{0,0,0}\setlength{\tabcolsep}{0pt}\begin{tabular}{c}\textbf{MAV}\end{tabular}}%
\psfrag{b}[t][t][\tsizeb]{\color[rgb]{0,0,0}\setlength{\tabcolsep}{0pt}\begin{tabular}{c}\textbf{Hardware}\end{tabular}}%
\psfrag{c}[t][t][\tsizeb]{\color[rgb]{0,0,0}\setlength{\tabcolsep}{0pt}\begin{tabular}{c}\textbf{Software (Paparazzi Autopilot)}\end{tabular}}%
\psfrag{d}[t][t][\tsize]{\color[rgb]{0,0,0}\setlength{\tabcolsep}{0pt}\begin{tabular}{c}\textit{Vision Module}\end{tabular}}%
\psfrag{e}[t][t][\tsize]{\color[rgb]{0,0,0}\setlength{\tabcolsep}{0pt}\begin{tabular}{c}\textit{Control Loop}\\ \textit{(Rotorcraft)}\end{tabular}}%
\psfrag{f}[t][t][\tsize]{\color[rgb]{0,0,0}\setlength{\tabcolsep}{0pt}\begin{tabular}{c}Camera\end{tabular}}%
\psfrag{g}[t][t][\tsize]{\color[rgb]{0,0,0}\setlength{\tabcolsep}{0pt}\begin{tabular}{c}IMU\end{tabular}}%
\psfrag{h}[t][t][\tsize]{\color[rgb]{0,0,0}\setlength{\tabcolsep}{0pt}\begin{tabular}{c}GPS\end{tabular}}%
\psfrag{i}[t][t][\tsize]{\color[rgb]{0,0,0}\setlength{\tabcolsep}{0pt}\begin{tabular}{c}Sonar\end{tabular}}%
\psfrag{j}[t][t][\tsize]{\color[rgb]{0,0,0}\setlength{\tabcolsep}{0pt}\begin{tabular}{c}Actuator\end{tabular}}%
\psfrag{k}[t][t][\tsize]{\color[rgb]{0,0,0}\setlength{\tabcolsep}{0pt}\begin{tabular}{c}YUV Image\\Capturing\end{tabular}}%
\psfrag{l}[t][t][\tsize]{\color[rgb]{0,0,0}\setlength{\tabcolsep}{0pt}\begin{tabular}{c}Optical Flow\\Computation\end{tabular}}%
\psfrag{m}[t][t][\tsize]{\color[rgb]{0,0,0}\setlength{\tabcolsep}{0pt}\begin{tabular}{c}Flow Field\\Fitting\end{tabular}}%
\psfrag{n}[t][t][\tsize]{\color[rgb]{0,0,0}\setlength{\tabcolsep}{0pt}\begin{tabular}{c}Texton\\Extraction\end{tabular}}%
\psfrag{o}[t][t][\tsize]{\color[rgb]{0,0,0}\setlength{\tabcolsep}{0pt}\begin{tabular}{c}SSL $f(\mathbf{q})$\end{tabular}}%
\psfrag{p}[t][t][\tsize]{\color[rgb]{0,0,0}\setlength{\tabcolsep}{0pt}\begin{tabular}{c}Attitude\\Stabilization\end{tabular}}%
\psfrag{q}[t][t][\tsize]{\color[rgb]{0,0,0}\setlength{\tabcolsep}{0pt}\begin{tabular}{c}Guidance\end{tabular}}%
\psfrag{r}[t][t][\tsize]{\color[rgb]{0,0,0}\setlength{\tabcolsep}{0pt}\begin{tabular}{c}$p,q,r$\end{tabular}}%
\psfrag{s}[t][t][\tsize]{\color[rgb]{0,0,0}\setlength{\tabcolsep}{0pt}\begin{tabular}{c}$x,y$\end{tabular}}%
\psfrag{t}[t][t][\tsize]{\color[rgb]{0,0,0}\setlength{\tabcolsep}{0pt}\begin{tabular}{c}$u,v$\end{tabular}}%
\psfrag{u}[t][t][\tsize]{\color[rgb]{0,0,0}\setlength{\tabcolsep}{0pt}\begin{tabular}{c}$\epsilon^*$\end{tabular}}%
\psfrag{v}[t][t][\tsize]{\color[rgb]{0,0,0}\setlength{\tabcolsep}{0pt}\begin{tabular}{c}$\epsilon^*$\end{tabular}}%
\psfrag{w}[t][t][\tsize]{\color[rgb]{0,0,0}\setlength{\tabcolsep}{0pt}\begin{tabular}{c}$\mathbf{q}$\end{tabular}}%
\psfrag{x}[t][t][\tsize]{\color[rgb]{0,0,0}\setlength{\tabcolsep}{0pt}\begin{tabular}{c}$X,Y$\end{tabular}}%
\psfrag{y}[t][t][\tsize]{\color[rgb]{0,0,0}\setlength{\tabcolsep}{0pt}\begin{tabular}{c}$\widehat{\epsilon}$\end{tabular}}%
\psfrag{z}[t][t][\tsize]{\color[rgb]{0,0,0}\setlength{\tabcolsep}{0pt}\begin{tabular}{c}$Z$\end{tabular}}%
\psfrag{z1}[t][t][\tsize]{\color[rgb]{0,0,0}\setlength{\tabcolsep}{0pt}\begin{tabular}{c}$\phi_{ref},\theta_{ref},\psi_{ref}$\end{tabular}}%
\psfrag{z2}[t][t][\tsize]{\color[rgb]{0,0,0}\setlength{\tabcolsep}{0pt}\begin{tabular}{c}$\mu$\end{tabular}}%
%
\includegraphics[trim = 0mm 0mm 0mm 0mm, clip, width=0.5\textwidth]{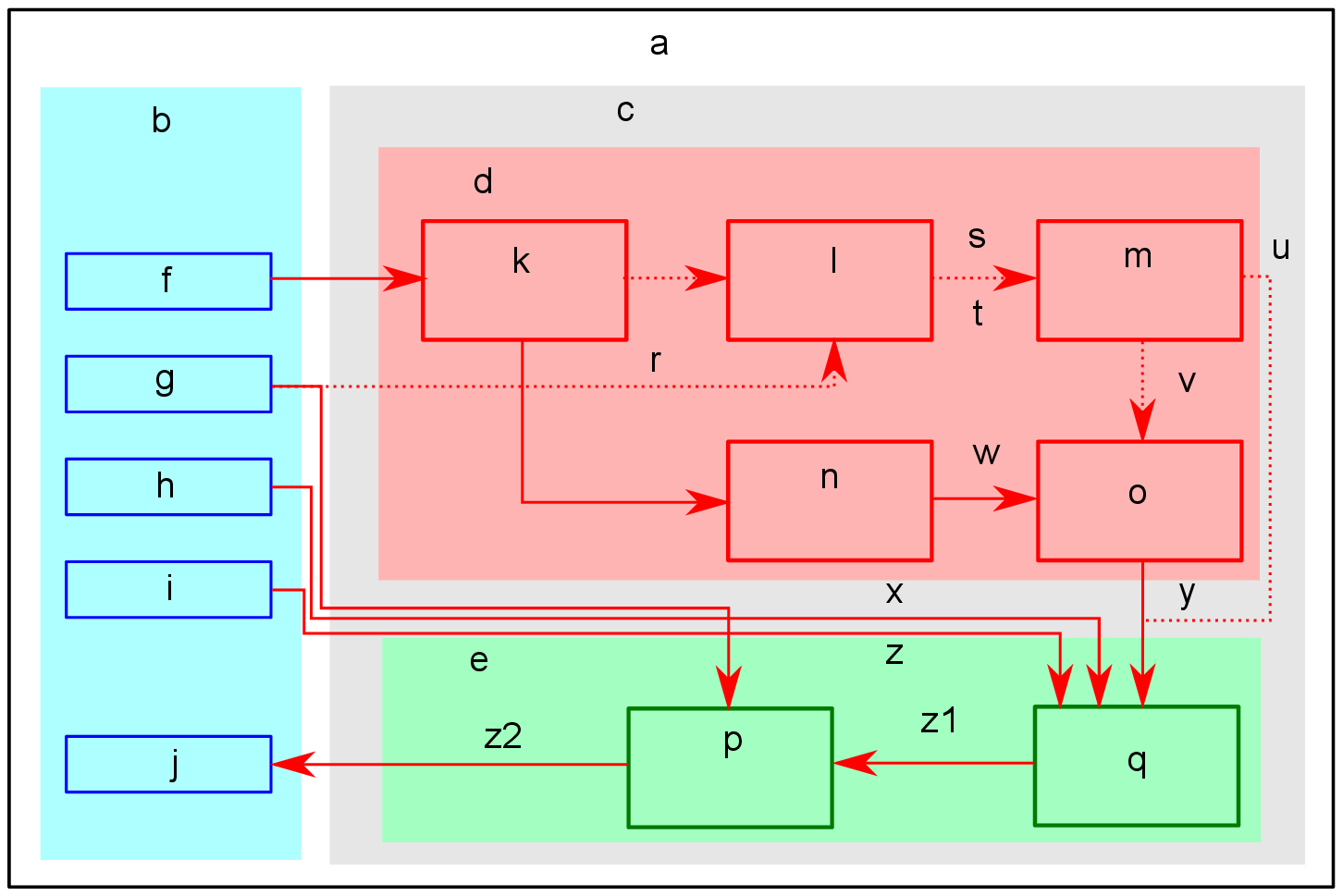}
\end{psfrags}%
%

	\caption{Integration of computer vision module in Paparazzi Autopilot. The left blue box shows the hardware of the MAV while the right gray box indicates the software architecture used in the MAV. The process flow guided by the dashed line can be neglected after the learning is complete.}
	\label{fig:Paparazzi_Autopilot}
\end{figure}

A Parrot AR.Drone 2.0\footnote{\url{http://ardrone2.parrot.com}} and Bebop\footnote{\url{http://www.parrot.com/products/bebop-drone/}} are used as a testing platform for indoor and outdoor experiments, respectively in this study. They are equipped with a downward-looking camera which runs up to $60~FPS$ and is of particular interest to us for the landing purpose. Instead of using the original Parrot AR.Drone program, an open-source autopilot software, Paparazzi Autopilot\footnote{\url{http://wiki.paparazziuav.org}} is used because it allows us to have direct access to the sensors and control the MAV \cite{remes2013paparazzi,hattenberger2014using}. We created a computer vision module in Paparazzi Autopilot to capture and process images from the camera and test our proposed algorithm in flight tests. Fig.~\ref{fig:Paparazzi_Autopilot} shows the overview of the control architecture of Paparazzi and how it integrates the vision module. All the computer processing tasks in this study are performed using the on-board processor of the MAV so that the MAV does not rely on the ground control station (GCS). This can avoid mission failure due to loss or delay of data transmission between MAV and GCS. 

In Fig.~\ref{fig:Paparazzi_Autopilot}, images are captured from the downward-looking camera in the vision module. These images are processed using the computer vision algorithms (presented in Section~\ref{sec:Method}), such as the optical flow and the texton methods, to detect obstacles and find a suitable landing spot. The IMU from the MAV is used in the optical flow algorithm to reduce the effect of MAV rotation on optical flow measurements. The output from the vision module (e.g. the position where it is considered safe to land upon) is fed into the control loop in Paparazzi autopilot to control the MAV. A GPS or motion tracking system provides position measurements of the MAV for outdoor or indoor navigation purposes. In this study, vision is purely used for the detection of obstacles on the landing surface.  

\subsection{Processing Time of Computer Vision Algorithms}
\label{subsec:ProcessingTime}
To examine the computational efficiency of the optical flow and SSL algorithms, we measured the times taken by each process of the algorithms. TABLE~\ref{table:ProcessingTime} shows the average processing time required for each stage of both algorithms. In our experiments, the maximum number of corners in optical flow algorithm and the number of samples used in SSL are both set to 25. Note that this value can be tuned to include more or less information from the images, however, a higher value requires more computational time. The total processing times (see TABLE~\ref{table:ProcessingTime}) show that both algorithms are computationally efficient and can be executed on-board the MAV. For instance, when both methods are running during data acquisition for learning, it costs roughly $35~ms$ which is approximately $28~FPS$. The processing time is even faster after learning as the SSL method runs at the frame rate of $\approx60~Hz$. This frame rate is sufficient to capture surrounding information for fast moving MAVs.

\begin{table}[thpb]
	\caption{Average processing time for each stage of the vision algorithms}
	\label{table:ProcessingTime}
	\begin{center}
		\begin{tabular}{|c||c|c|c|c|}
			\hline
			Optical  & Corner & Corner & Flow & Total \\
			Flow & Detection & Tracking & Fitting &\\
			\hline
			Time (ms) & 14.04 & 4.01 & 2.01 & \textbf{20.06}\\
			\hline
			\hline
			SSL & \multicolumn{2}{c|}{Distribution}& Regression & Total\\
			& \multicolumn{2}{c|}{Extraction}& Function &\\
			\hline
			Time (ms) & \multicolumn{2}{c|}{15.13}& 0.01 & \textbf{15.14}\\
			\hline
		\end{tabular}
	\end{center}
\end{table}	

\subsection{Autonomous Landing Strategy Using $\epsilon^*$ from Optical Flow}
\label{subsec:LandingStrategyRoughness}
In this subsection, we first explain a landing strategy using $\epsilon^*$ to guide the MAV to land on a safe landing place. Then, we demonstrate the experiment results from the flight tests we performed with the landing strategy.

\subsubsection{Landing Strategy}
\label{subsubsec:LandingStrategyRoughness}
Here, we propose a straightforward landing strategy which allows the MAV to decide where to land safely using $\epsilon^*$ from optical flow algorithm. Since this method requires movement of the MAV to detect obstacles, the MAV needs to fly over the potential landing area in order to search for a suitable landing spot/ waypoint. In this strategy, we define the safest landing spot as a waypoint in which it covers the largest area without obstacles underneath the MAV. To determine this waypoint, we first classify on-board images into safe ($\mathit{SF}=1$) and unsafe ($\mathit{SF}=0$) classes by thresholding $\epsilon^*$ with $\epsilon_{th}$ as shown in Eq.~(\ref{equation:safety}).

\begin{equation}
\mathit{SF}_i = \left\{ 
\begin{array}{l l}
1 & \quad \text{if $\epsilon_i^* < \epsilon_{th}$}\\
0 & \quad \text{if $\epsilon_i^* > \epsilon_{th}$}
\end{array} \right.  
\label{equation:safety}
\end{equation}

\noindent Then, the largest area without obstacle can be found by choosing the largest value of $\mathit{A_{SF}} = \sum^{i=L}_{i=1}{SF_i}$ where $i=1:L$ are a set of continuous images which are classified as safe images. Once an obstacle is detected ($\mathit{SF_{i=L+1}}=0$), $\mathit{A_{SF}}$ is reset to $0$ and $i=1$. Furthermore, the middle waypoint of the safest area, $\mathit{P}_{\mathit{land}}$ is considered as the desired landing spot and it is continuously updated when the largest $\mathit{A_{SF}}$ is found.

\subsubsection{Experiment Results}
\label{subsubsec:ExperimentResultsRoughness}
\begin{figure}[thpb]
	\centering
	\captionsetup{justification=centering}
%
%
\begin{psfrags}%
\psfragscanon%
\newcommand{\tsize}{0.8}
\newcommand{\tsizeb}{0.5}
%
\psfrag{HOME}[t][t][\tsize]{\color[rgb]{0,0,0}\setlength{\tabcolsep}{0pt}\begin{tabular}{c}Image\\Patch\end{tabular}}%
\psfrag{1}[t][t][\tsize]{\color[rgb]{0,0,0}\setlength{\tabcolsep}{0pt}\begin{tabular}{c}Dictionary\end{tabular}}%
\psfrag{c}[t][t][\tsize]{\color[rgb]{0,0,0}\setlength{\tabcolsep}{0pt}\begin{tabular}{c}Image\end{tabular}}%
\psfrag{d}[t][t][\tsize]{\color[rgb]{0,0,0}\setlength{\tabcolsep}{0pt}\begin{tabular}{c}Textons\end{tabular}}%
\psfrag{e}[t][t][\tsize]{\color[rgb]{0,0,0}\setlength{\tabcolsep}{0pt}\begin{tabular}{c}Probability\\Distribution\end{tabular}}%
%
\includegraphics[trim = 0mm 0mm 0mm 0mm, clip, width=0.5\textwidth]{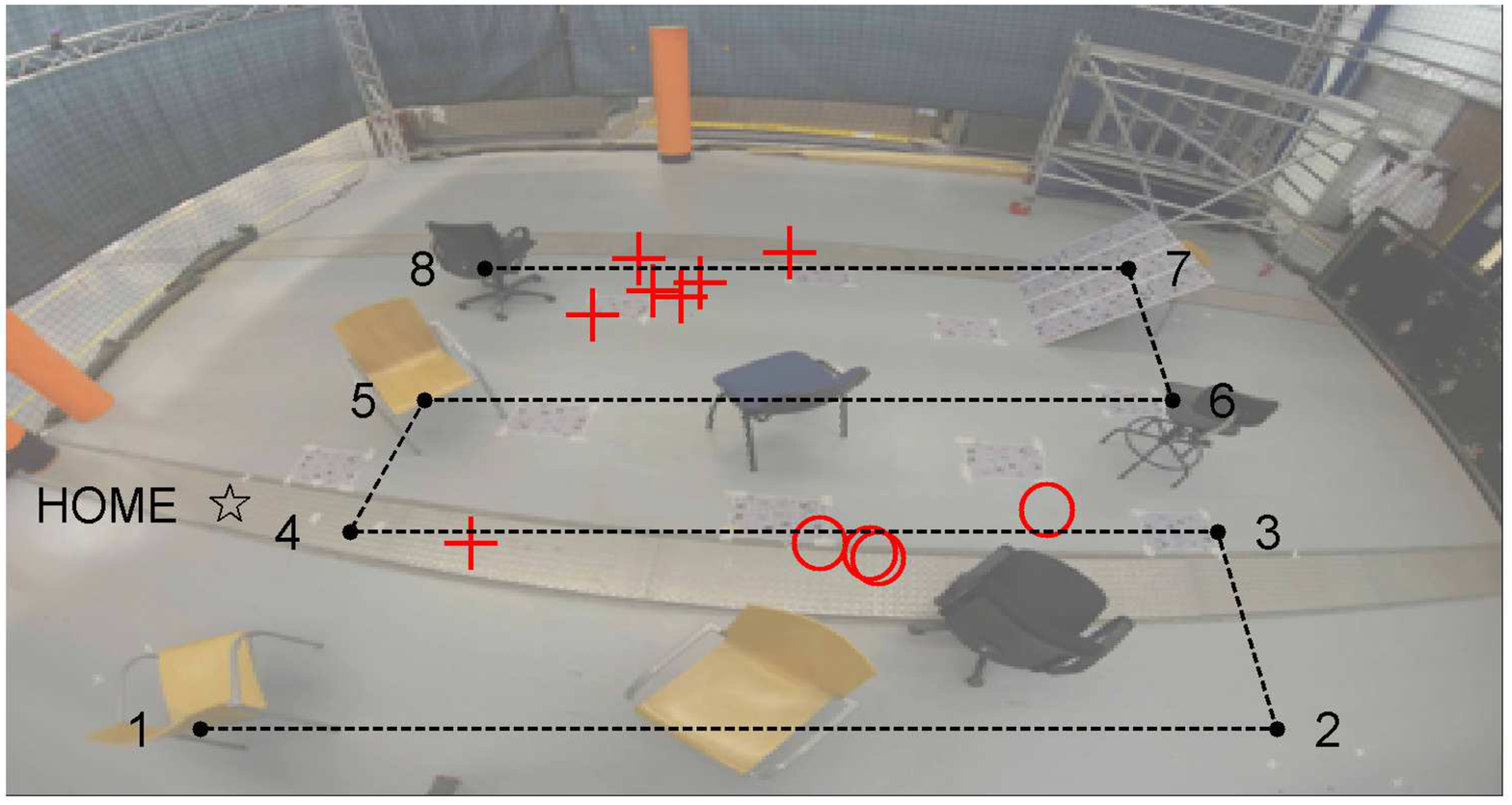}
\end{psfrags}%
%

	\caption{Flight plan and landing waypoints of the MAV based on roughness estimate $\epsilon^*$ from optical flow algorithm. Black dashed line represents the flight path following waypoints 1 to 8 and HOME is the position where the MAV take-off. Red markers indicate the landing spots $\mathit{P}_{\mathit{land}}$ directed using the landing strategy.}
	\label{fig:LandingFlatness}
\end{figure}

The optical flow algorithm and landing strategy were implemented in a computer vision module in Paparazzi Autopilot and ran in real-time on a Parrot AR drone as described in Subsection~\ref{subsec:ExperimentPlatform}. A motion tracking system, OptiTrack was used to serve as an indoor GPS to allow the MAV to fly autonomously in the arena according to a simple flight plan (see Fig.~\ref{fig:LandingFlatness}). The MAV took off from its HOME position, and followed a route starting from waypoint 1 and ending at waypoint 8. After scanning the whole landing site using its on-board camera, it navigated to a safe landing waypoint using the proposed landing strategy and landed there.

\begin{figure}[thpb]
	\centering
	\captionsetup{justification=centering}
%
%
\begin{psfrags}%
\psfragscanon%
\newcommand{\tsize}{0.7}
\newcommand{\tsizeb}{0.65}
%
\psfrag{s03}[t][t][\tsize]{\color[rgb]{0,0,0}\setlength{\tabcolsep}{0pt}\begin{tabular}{c}x~(m)\end{tabular}}%
\psfrag{s04}[b][b][\tsize]{\color[rgb]{0,0,0}\setlength{\tabcolsep}{0pt}\begin{tabular}{c}y~(m)\end{tabular}}%
\psfrag{s05}[l][l][\tsize]{\color[rgb]{0,0,0}\setlength{\tabcolsep}{0pt}\begin{tabular}{l}1\end{tabular}}%
\psfrag{s06}[l][l][\tsize]{\color[rgb]{0,0,0}\setlength{\tabcolsep}{0pt}\begin{tabular}{l}2\end{tabular}}%
\psfrag{s07}[l][l][\tsize]{\color[rgb]{0,0,0}\setlength{\tabcolsep}{0pt}\begin{tabular}{l}3\end{tabular}}%
\psfrag{s08}[b][b][\tsize]{\color[rgb]{0,0,0}\setlength{\tabcolsep}{0pt}\begin{tabular}{l}4\end{tabular}}%
\psfrag{s09}[l][l][\tsize]{\color[rgb]{0,0,0}\setlength{\tabcolsep}{0pt}\begin{tabular}{l}5\end{tabular}}%
\psfrag{s10}[l][l][\tsize]{\color[rgb]{0,0,0}\setlength{\tabcolsep}{0pt}\begin{tabular}{l}6\end{tabular}}%
\psfrag{s11}[l][l][\tsize]{\color[rgb]{0,0,0}\setlength{\tabcolsep}{0pt}\begin{tabular}{l}7\end{tabular}}%
\psfrag{s12}[l][l][\tsize]{\color[rgb]{0,0,0}\setlength{\tabcolsep}{0pt}\begin{tabular}{l}8\end{tabular}}%
\psfrag{s13}[l][l][\tsize]{\color[rgb]{0,0,0}\setlength{\tabcolsep}{0pt}\begin{tabular}{l}Land\end{tabular}}%
%
\psfrag{x01}[t][t][\tsize]{-1}%
\psfrag{x02}[t][t][\tsize]{0}%
\psfrag{x03}[t][t][\tsize]{1}%
\psfrag{x04}[t][t][\tsize]{2}%
\psfrag{x05}[t][t][\tsize]{3}%
\psfrag{x06}[t][t][\tsize]{4}%
\psfrag{x07}[t][t][\tsize]{5}%
%
\psfrag{v01}[r][r][\tsize]{-2}%
\psfrag{v02}[r][r][\tsize]{-1}%
\psfrag{v03}[r][r][\tsize]{0}%
\psfrag{v04}[r][r][\tsize]{1}%
\psfrag{v05}[r][r][\tsize]{2}%
\psfrag{v06}[r][r][\tsize]{3}%
%
\includegraphics[trim=0mm 0mm 0mm 0mm,clip,width=0.5\textwidth]{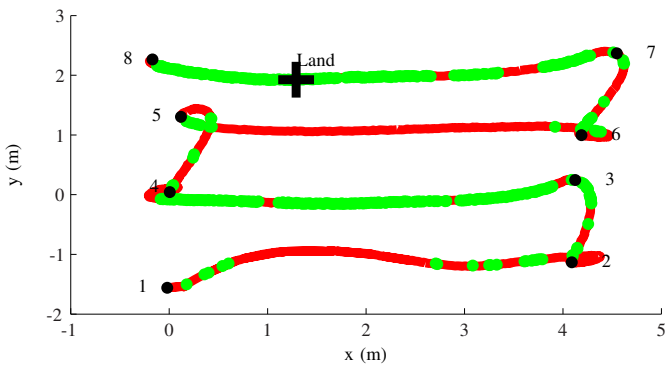}
\end{psfrags}%
%

	\caption{Safety value $\mathit{SF}$ along the flight path in one of the experiments. Green markers indicate the value of $\mathit{SF}$ equals to 1. Black cross shows where the landing spot is.}
	\label{fig:SFColorMap}
\end{figure}

We performed 11 landing tests with the Parrot AR drone. In each test, it first flew the entire flight plan and then returned to the point it considered safest. Fig.~\ref{fig:SFColorMap} shows the safety value $\mathit{SF}$ along the flight path in one of the landing experiments. In this figure, the longest stretch $A_{SF}$ is located between waypoints 7 and 8, and the landing spot $\mathit{P}_{\mathit{land}}$ (indicated as black cross) is selected in this area. The landing spots for all the experiments are plotted with red markers in Fig.~\ref{fig:LandingFlatness}. All landings were successful and on flat/ non-obstacle ground. Still, we have indicated with circle markers in the figure $4$ landing spots that were rather close to obstacles in the environment. We expect that measurement noise is the cause of landing on these locations rather than the safer area higher up in the image.

\subsection{Autonomous Landing Strategy Using $\hat{\epsilon}$ from Appearance}
\label{subsec:LandingStrategyAppearance}
To utilize the advantage of SSL method, we switch to detecting obstacles using only the appearance while the MAV is hovering. 	

\subsubsection{Landing Strategy}
\label{subsubsec:LandingStrategyAppearance}
In the previous subsections, we used the whole image to determine whether the region underneath the MAV was a safe landing place. However, it did not tell us where to go if we found the obstacle(s). Therefore, we propose a straightforward method for allowing the MAV to decide where to land using $\hat{\epsilon}$. In this strategy, the image is divided into nine regions (grid of $3\times3$). Using the SSL method, we estimate $\hat{\epsilon}$ in each region and choose the region with minimum $\hat{\epsilon}$ as the place we are going to land or reject them if they are all higher than a threshold value (indicating the presence of an obstacle). To compute the movement, we take the distance $d_c$ from the image center to the center of the region where it is considered safe and use the height $h$ measured from the sonar sensor to project this distance in pixels to physical distance in meters $d_p$ using Eq.~(\ref{equation:DistanceProjection}). 

\begin{equation}
d_p = \frac{d_c\times h}{\tilde{f}}
\label{equation:DistanceProjection}
\end{equation}
\noindent where $\tilde{f}$ is the focal length of the camera in pixels.

\subsubsection{Experiment Results}
\label{subsubsec:ExperimentResultsSSL}
\begin{figure}[h!]
	\centering
	\captionsetup{justification=centering}
	\input{image_ob_SSL1.tex}\hspace{9.4mm}\input{image_ob_SSL2.tex}\\ \vspace{2mm}
%
%
\begin{psfrags}%
\psfragscanon%
\newcommand{\tsize}{0.8}
\newcommand{\tsizeb}{0.5}
%
\psfrag{s04}[c][c][\tsize]{\color[rgb]{1,1,1}\setlength{\tabcolsep}{0pt}\begin{tabular}{l}0.25\end{tabular}}%
\psfrag{s05}[c][c][\tsize]{\color[rgb]{1,1,1}\setlength{\tabcolsep}{0pt}\begin{tabular}{l}0.25\end{tabular}}%
\psfrag{s06}[c][c][\tsize]{\color[rgb]{1,1,1}\setlength{\tabcolsep}{0pt}\begin{tabular}{l}0.38\end{tabular}}%
\psfrag{s07}[c][c][\tsize]{\color[rgb]{0,0,0}\setlength{\tabcolsep}{0pt}\begin{tabular}{l}0.57\end{tabular}}%
\psfrag{s08}[c][c][\tsize]{\color[rgb]{0,0,0}\setlength{\tabcolsep}{0pt}\begin{tabular}{l}0.53\end{tabular}}%
\psfrag{s09}[c][c][\tsize]{\color[rgb]{1,1,1}\setlength{\tabcolsep}{0pt}\begin{tabular}{l}0.32\end{tabular}}%
\psfrag{s10}[c][c][\tsize]{\color[rgb]{0,0,0}\setlength{\tabcolsep}{0pt}\begin{tabular}{l}0.58\end{tabular}}%
\psfrag{s11}[c][c][\tsize]{\color[rgb]{0,0,0}\setlength{\tabcolsep}{0pt}\begin{tabular}{l}0.62\end{tabular}}%
\psfrag{s12}[c][c][\tsize]{\color[rgb]{1,1,1}\setlength{\tabcolsep}{0pt}\begin{tabular}{l}0.32\end{tabular}}%
%
\psfrag{x01}[t][t][\tsize]{0.5}%
\psfrag{x02}[t][t][\tsize]{1}%
\psfrag{x03}[t][t][\tsize]{1.5}%
\psfrag{x04}[t][t][\tsize]{2}%
\psfrag{x05}[t][t][\tsize]{2.5}%
\psfrag{x06}[t][t][\tsize]{3}%
\psfrag{x07}[t][t][\tsize]{3.5}%
%
\psfrag{v01}[r][r][\tsize]{0.5}%
\psfrag{v02}[r][r][\tsize]{1}%
\psfrag{v03}[r][r][\tsize]{1.5}%
\psfrag{v04}[r][r][\tsize]{2}%
\psfrag{v05}[r][r][\tsize]{2.5}%
\psfrag{v06}[r][r][\tsize]{3}%
\psfrag{v07}[r][r][\tsize]{3.5}%
%
\includegraphics[trim = 1mm 0mm 1mm 0mm, clip, width=0.2\textwidth]{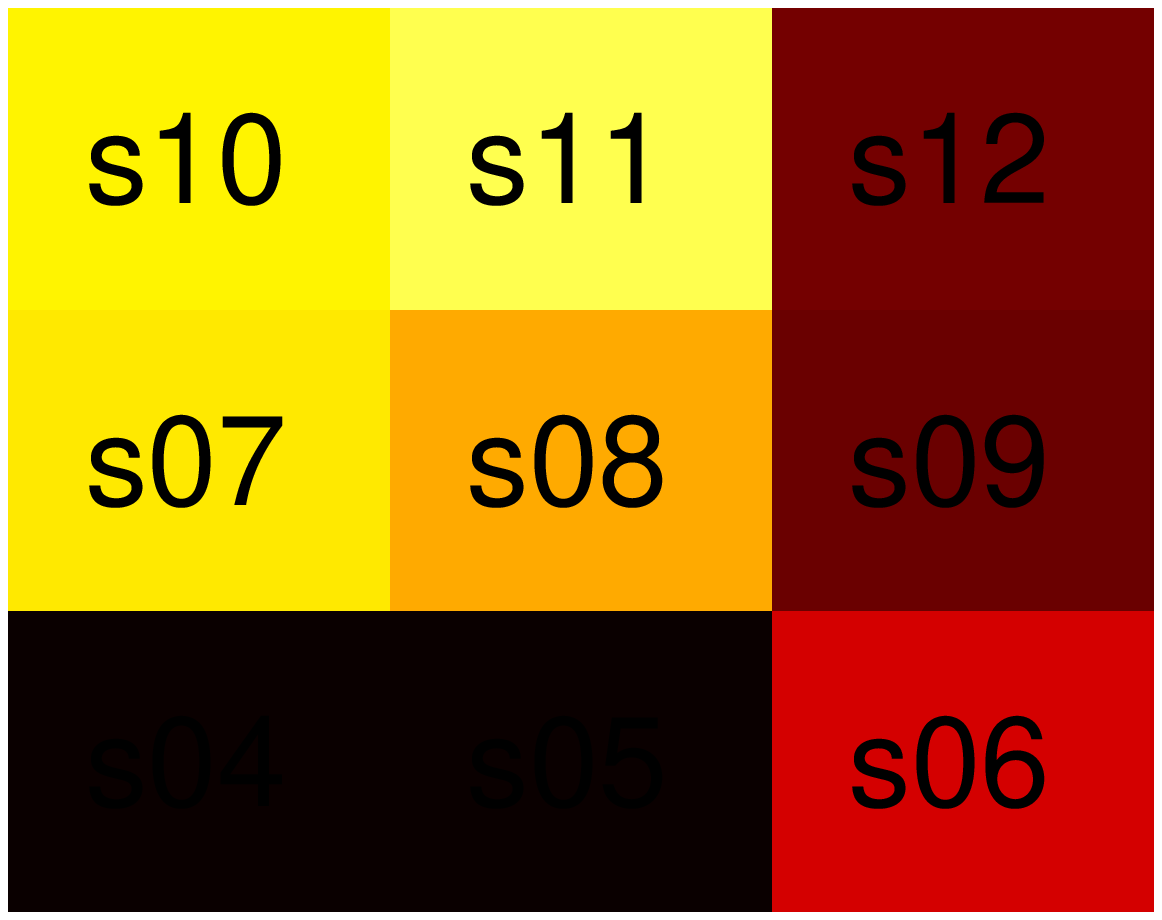}
\end{psfrags}%
%
\qquad
%
%
\begin{psfrags}%
\psfragscanon%
\newcommand{\tsize}{0.8}
\newcommand{\tsizeb}{0.5}
%
\psfrag{s04}[c][c][\tsize]{\color[rgb]{0,0,0}\setlength{\tabcolsep}{0pt}\begin{tabular}{l}0.57\end{tabular}}%
\psfrag{s05}[c][c][\tsize]{\color[rgb]{0,0,0}\setlength{\tabcolsep}{0pt}\begin{tabular}{l}0.47\end{tabular}}%
\psfrag{s06}[c][c][\tsize]{\color[rgb]{0,0,0}\setlength{\tabcolsep}{0pt}\begin{tabular}{l}0.47\end{tabular}}%
\psfrag{s07}[c][c][\tsize]{\color[rgb]{0,0,0}\setlength{\tabcolsep}{0pt}\begin{tabular}{l}0.63\end{tabular}}%
\psfrag{s08}[c][c][\tsize]{\color[rgb]{0,0,0}\setlength{\tabcolsep}{0pt}\begin{tabular}{l}0.40\end{tabular}}%
\psfrag{s09}[c][c][\tsize]{\color[rgb]{1,1,1}\setlength{\tabcolsep}{0pt}\begin{tabular}{l}0.31\end{tabular}}%
\psfrag{s10}[c][c][\tsize]{\color[rgb]{0,0,0}\setlength{\tabcolsep}{0pt}\begin{tabular}{l}0.55\end{tabular}}%
\psfrag{s11}[c][c][\tsize]{\color[rgb]{1,1,1}\setlength{\tabcolsep}{0pt}\begin{tabular}{l}0.31\end{tabular}}%
\psfrag{s12}[c][c][\tsize]{\color[rgb]{1,1,1}\setlength{\tabcolsep}{0pt}\begin{tabular}{l}0.26\end{tabular}}%
%
\psfrag{x01}[t][t][\tsize]{0.5}%
\psfrag{x02}[t][t][\tsize]{1}%
\psfrag{x03}[t][t][\tsize]{1.5}%
\psfrag{x04}[t][t][\tsize]{2}%
\psfrag{x05}[t][t][\tsize]{2.5}%
\psfrag{x06}[t][t][\tsize]{3}%
\psfrag{x07}[t][t][\tsize]{3.5}%
%
\psfrag{v01}[r][r][\tsize]{0.5}%
\psfrag{v02}[r][r][\tsize]{1}%
\psfrag{v03}[r][r][\tsize]{1.5}%
\psfrag{v04}[r][r][\tsize]{2}%
\psfrag{v05}[r][r][\tsize]{2.5}%
\psfrag{v06}[r][r][\tsize]{3}%
\psfrag{v07}[r][r][\tsize]{3.5}%
%
\includegraphics[trim = 1mm 0mm 1mm 0mm, clip, width=0.2\textwidth]{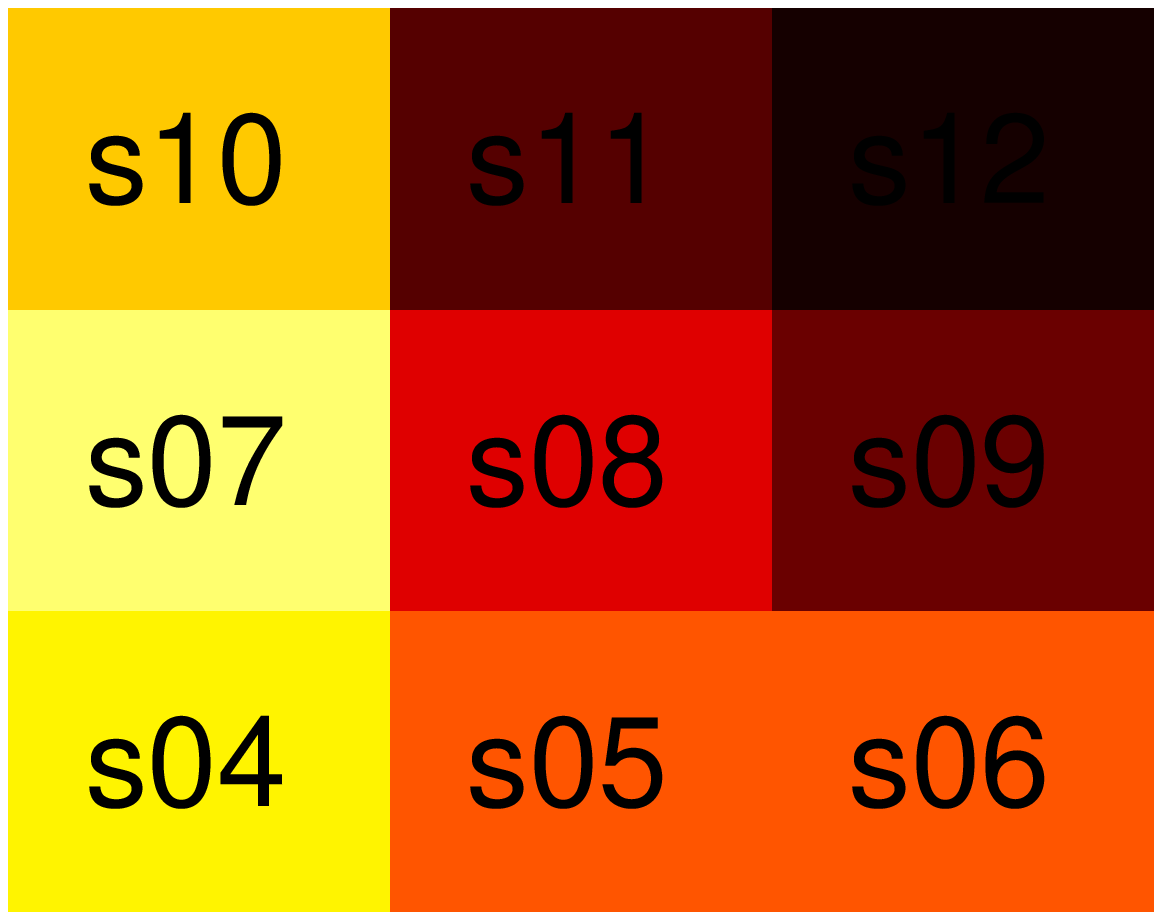}
\end{psfrags}%
%
\\ \vspace{2mm}
%
%
\begin{psfrags}%
\psfragscanon%
\newcommand{\tsize}{0.7}
\newcommand{\tsizeb}{0.65}
%
\psfrag{s01}[l][l][\tsize]{\color[rgb]{0,0,0}\setlength{\tabcolsep}{0pt}\begin{tabular}{l}Start\end{tabular}}%
\psfrag{s02}[b][b][\tsize]{\color[rgb]{0.15,0.15,0.15}\setlength{\tabcolsep}{0pt}\begin{tabular}{c}$X_{b}~(m)$\end{tabular}}%
\psfrag{s04}[t][t][\tsize][\tsize]{\color[rgb]{0.15,0.15,0.15}\setlength{\tabcolsep}{0pt}\begin{tabular}{c}$Y_{b}~(m)$\end{tabular}}%
\psfrag{s05}[l][l][\tsize]{\color[rgb]{0,0,0}\setlength{\tabcolsep}{0pt}\begin{tabular}{l}Land\end{tabular}}%
%
\color[rgb]{0.15,0.15,0.15}%
%
\psfrag{x01}[t][t][\tsize]{}%
\psfrag{x02}[t][t][\tsize]{-1}%
\psfrag{x03}[t][t][\tsize]{}%
\psfrag{x04}[t][t][\tsize]{0}%
\psfrag{x05}[t][t][\tsize]{}%
\psfrag{x06}[t][t][\tsize]{1}%
\psfrag{x07}[t][t][\tsize]{}%
\psfrag{x08}[t][t][\tsize]{2}%
%
\psfrag{v01}[r][r][\tsize]{0}%
\psfrag{v02}[r][r][\tsize]{}%
\psfrag{v03}[r][r][\tsize]{1}%
\psfrag{v04}[r][r][\tsize]{}%
\psfrag{v05}[r][r][\tsize]{2}%
\psfrag{v06}[r][r][\tsize]{}%
%
\includegraphics[trim = 1mm 0mm 1mm 0mm, clip, width=0.2\textwidth]{land_path_SSL1.eps}
\end{psfrags}%
%
\qquad
%
%
\begin{psfrags}%
\psfragscanon%
\newcommand{\tsize}{0.7}
\newcommand{\tsizeb}{0.65}
%
\psfrag{s02}[l][l][\tsize]{\color[rgb]{0,0,0}\setlength{\tabcolsep}{0pt}\begin{tabular}{l}Land\end{tabular}}%
\psfrag{s03}[l][l][\tsize]{\color[rgb]{0,0,0}\setlength{\tabcolsep}{0pt}\begin{tabular}{l}Start\end{tabular}}%
\psfrag{s05}[b][b][\tsize]{\color[rgb]{0.15,0.15,0.15}\setlength{\tabcolsep}{0pt}\begin{tabular}{c}$X_{b}~(m)$\end{tabular}}%
\psfrag{s06}[t][t][\tsize]{\color[rgb]{0.15,0.15,0.15}\setlength{\tabcolsep}{0pt}\begin{tabular}{c}$Y_{b}~(m)$\end{tabular}}%
%
\color[rgb]{0.15,0.15,0.15}%
%
\psfrag{x01}[t][t][\tsize]{}%
\psfrag{x02}[t][t][\tsize]{-1}%
\psfrag{x03}[t][t][\tsize]{}%
\psfrag{x04}[t][t][\tsize]{0}%
\psfrag{x05}[t][t][\tsize]{}%
\psfrag{x06}[t][t][\tsize]{1}%
\psfrag{x07}[t][t][\tsize]{}%
\psfrag{x08}[t][t][\tsize]{2}%
%
\psfrag{v01}[r][r][\tsize]{}%
\psfrag{v02}[r][r][\tsize]{1}%
\psfrag{v03}[r][r][\tsize]{}%
\psfrag{v04}[r][r][\tsize]{2}%
\psfrag{v05}[r][r][\tsize]{}%
\psfrag{v06}[r][r][\tsize]{3}%
\psfrag{v07}[r][r][\tsize]{}%
%
\includegraphics[trim = 1mm 0mm 1mm 0mm, clip, width=0.2\textwidth]{land_path_SSL2.eps}
\end{psfrags}%
%
\\ \vspace{2mm}
	\caption{Indoor Experiment: Cross markers in the top row images show the chosen landing spot. Center row presents $\hat{\epsilon}$ for all regions with color scaled according $\hat{\epsilon}$ (Yellow = obstacle; Black = non-obstacle). Bottom row presents the flight path of the MAV from the time when the landing strategy activated until the time when it landed.}
	\label{fig:AppearanceLandingStrategy}
\end{figure}

The landing experiments were conducted in indoor and outdoor environments at the heights above the ground of $\approx3.5~m$ and $\approx7~m$, respectively. In the experiments, the MAV hovered at waypoints where one or more obstacles were underneath it. Once the landing strategy was activated, the MAV autonomously moved and landed at a place it considered safe. Fig.~\ref{fig:AppearanceLandingStrategy} shows the results of the flight tests in an indoor environment with the presence of single (first column) and multiple obstacles (second column). The top row of this figure presents the images taken when the landing strategy was activated. These images were divided into nine regions (separated by red lines) for $\hat{\epsilon}$ computation and the minimum $\hat{\epsilon}$ (indicated with a green cross) was chosen as the landing spot. The center row of the figure shows the $\hat{\epsilon}$ values for each region with colors representing the degree of visibility of the obstacle scaled using $\hat{\epsilon}$. The yellow color indicates the presence of an obstacle whereas the black color shows the absence of an obstacle. The bottom row of the figure illustrates the flight path of the MAV from the position where the landing strategy began to the position where the MAV landed. The results clearly demonstrate that the proposed strategy manages to lead the MAV to a safe landing place. \footnote{Experiment video: \url{https://goo.gl/Le5HT2}}

\begin{figure}[h!]
	\centering
	\captionsetup{justification=centering}
	\input{image_ob_SSL_outdoor_arxiv.tex}\qquad
%
%
\begin{psfrags}%
\psfragscanon%
\newcommand{\tsize}{0.7}
\newcommand{\tsizeb}{0.65}
%
\psfrag{s03}[t][c][\tsize]{\color[rgb]{0,0,0}\setlength{\tabcolsep}{0pt}\begin{tabular}{l}Start\end{tabular}}%
\psfrag{s04}[l][c][\tsize]{\color[rgb]{0,0,0}\setlength{\tabcolsep}{0pt}\begin{tabular}{l}Land\end{tabular}}%
\psfrag{s05}[t][t][\tsize]{\color[rgb]{0,0,0}\setlength{\tabcolsep}{0pt}\begin{tabular}{c}\small$Y_{b}~(m)$\end{tabular}}%
\psfrag{s06}[b][b][\tsize]{\color[rgb]{0,0,0}\setlength{\tabcolsep}{0pt}\begin{tabular}{c}\small$X_{b}~(m)$\end{tabular}}%
%
\psfrag{x01}[t][t][\tsize]{\small-8}%
\psfrag{x02}[t][t][\tsize]{}%
\psfrag{x03}[t][t][\tsize]{\small-4}%
\psfrag{x04}[t][t][\tsize]{}%
\psfrag{x05}[t][t][\tsize]{\small0}%
\psfrag{x06}[t][t][\tsize]{}%
\psfrag{x07}[t][t][\tsize]{\small4}%
\psfrag{x08}[t][t][\tsize]{}%
%
\psfrag{v01}[r][r][\tsize]{\small0}%
\psfrag{v02}[r][r][\tsize]{}%
\psfrag{v03}[r][r][\tsize]{\small4}%
\psfrag{v04}[r][r][\tsize]{}%
\psfrag{v05}[r][r][\tsize]{\small8}%
\psfrag{v06}[r][r][\tsize]{}%
\psfrag{v07}[r][r][\tsize]{\small12}%
\psfrag{v08}[r][r][\tsize]{}%
%
\includegraphics[trim = 1mm 0mm 1mm 0mm, clip, width=0.2\textwidth]{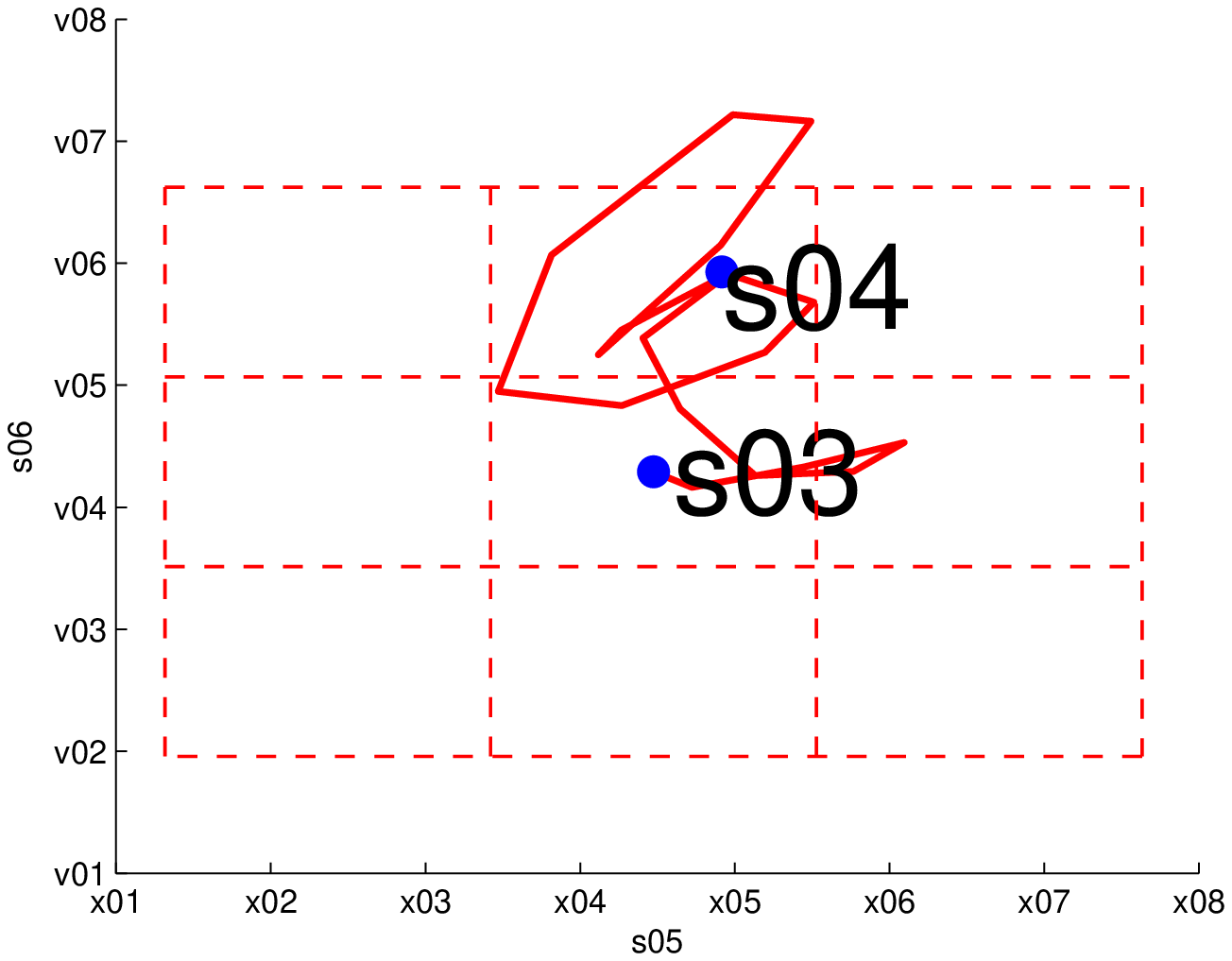}
\end{psfrags}%
%
\\ \vspace{2mm}
%
%
\begin{psfrags}%
\psfragscanon%
\newcommand{\tsize}{0.8}
\newcommand{\tsizeb}{0.5}
%
\psfrag{s04}[c][c][\tsize]{\color[rgb]{0,0,0}\setlength{\tabcolsep}{0pt}\begin{tabular}{l}0.23\end{tabular}}%
\psfrag{s05}[c][c][\tsize]{\color[rgb]{0,0,0}\setlength{\tabcolsep}{0pt}\begin{tabular}{l}0.28\end{tabular}}%
\psfrag{s06}[c][c][\tsize]{\color[rgb]{0,0,0}\setlength{\tabcolsep}{0pt}\begin{tabular}{l}0.26\end{tabular}}%
\psfrag{s07}[c][c][\tsize]{\color[rgb]{0,0,0}\setlength{\tabcolsep}{0pt}\begin{tabular}{l}0.26\end{tabular}}%
\psfrag{s08}[c][c][\tsize]{\color[rgb]{0,0,0}\setlength{\tabcolsep}{0pt}\begin{tabular}{l}0.26\end{tabular}}%
\psfrag{s09}[c][c][\tsize]{\color[rgb]{0,0,0}\setlength{\tabcolsep}{0pt}\begin{tabular}{l}0.25\end{tabular}}%
\psfrag{s10}[c][c][\tsize]{\color[rgb]{1,1,1}\setlength{\tabcolsep}{0pt}\begin{tabular}{l}0.22\end{tabular}}%
\psfrag{s11}[c][c][\tsize]{\color[rgb]{1,1,1}\setlength{\tabcolsep}{0pt}\begin{tabular}{l}0.11\end{tabular}}%
\psfrag{s12}[c][c][\tsize]{\color[rgb]{1,1,1}\setlength{\tabcolsep}{0pt}\begin{tabular}{l}0.16\end{tabular}}%
%
\psfrag{x01}[t][t][\tsize]{0.5}%
\psfrag{x02}[t][t][\tsize]{1}%
\psfrag{x03}[t][t][\tsize]{1.5}%
\psfrag{x04}[t][t][\tsize]{2}%
\psfrag{x05}[t][t][\tsize]{2.5}%
\psfrag{x06}[t][t][\tsize]{3}%
\psfrag{x07}[t][t][\tsize]{3.5}%
%
\psfrag{v01}[r][r][\tsize]{0.5}%
\psfrag{v02}[r][r][\tsize]{1}%
\psfrag{v03}[r][r][\tsize]{1.5}%
\psfrag{v04}[r][r][\tsize]{2}%
\psfrag{v05}[r][r][\tsize]{2.5}%
\psfrag{v06}[r][r][\tsize]{3}%
\psfrag{v07}[r][r][\tsize]{3.5}%
%
\includegraphics[trim = 1mm 0mm 1mm 0mm, clip, width=0.2\textwidth]{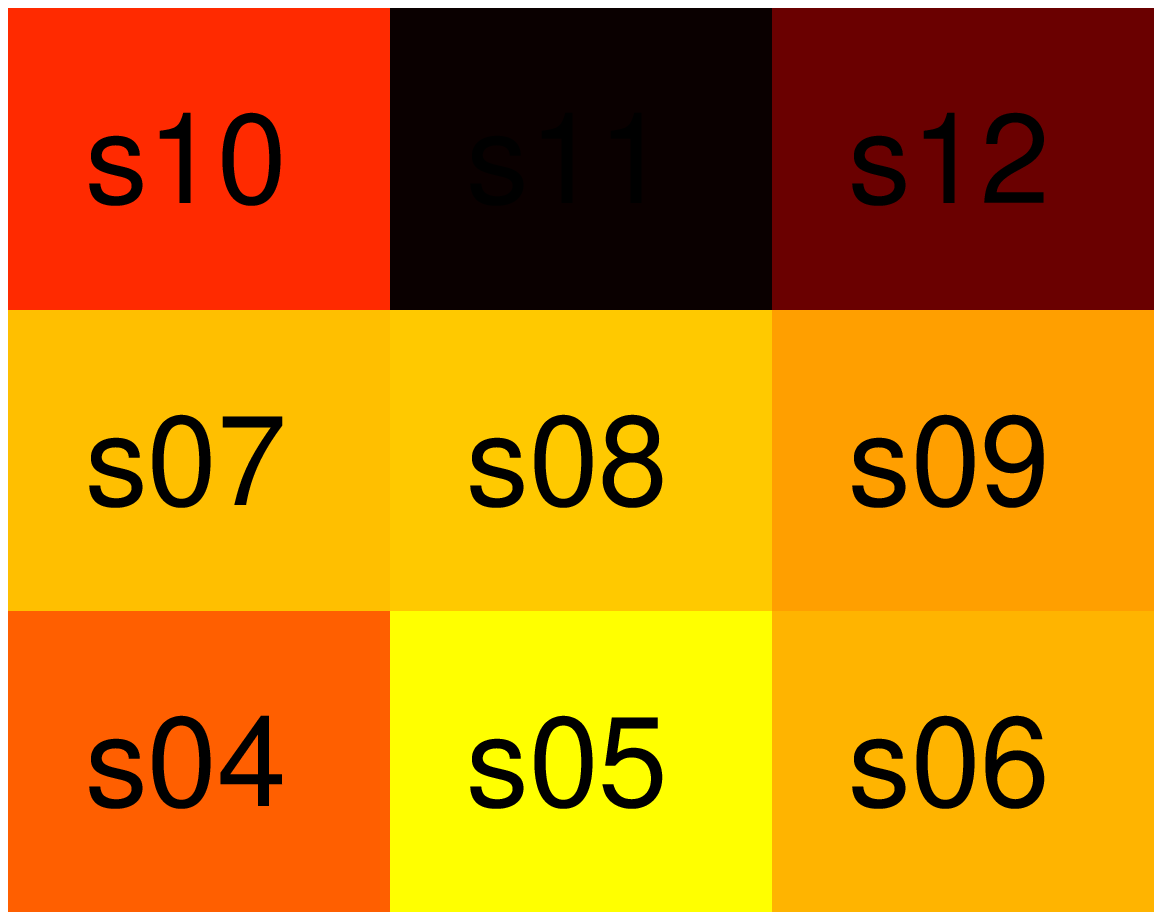}
\end{psfrags}%
%
\qquad
%
%
\begin{psfrags}%
\psfragscanon%
\newcommand{\tsize}{0.7}
\newcommand{\tsizeb}{0.65}
%
\psfrag{s06}[b][c][\tsize]{\color[rgb]{0,0,0}\setlength{\tabcolsep}{0pt}\begin{tabular}{c}$\hat{\epsilon}_m$\end{tabular}}%
\psfrag{s07}[t][c][\tsize]{\color[rgb]{0,0,0}\setlength{\tabcolsep}{0pt}\begin{tabular}{c}\small time~(s)\end{tabular}}%
\psfrag{s08}[b][c][\tsize]{\color[rgb]{0,0,0}\setlength{\tabcolsep}{0pt}\begin{tabular}{c}\scriptsize$h~(m)$\end{tabular}}%
%
\psfrag{x01}[t][t][\tsize]{\small170}%
\psfrag{x02}[t][t][\tsize]{}%
\psfrag{x03}[t][t][\tsize]{\small180}%
\psfrag{x04}[t][t][\tsize]{}%
\psfrag{x05}[t][t][\tsize]{\small190}%
\psfrag{x06}[t][t][\tsize]{}%
\psfrag{x07}[t][t][\tsize]{\small170}%
\psfrag{x08}[t][t][\tsize]{}%
\psfrag{x09}[t][t][\tsize]{\small180}%
\psfrag{x10}[t][t][\tsize]{}%
\psfrag{x11}[t][t][\tsize]{\small190}%
\psfrag{x12}[t][t][\tsize]{}%
%
\psfrag{v01}[r][r][\tsize]{}%
\psfrag{v02}[r][r][\tsize]{\small0}%
\psfrag{v03}[r][r][\tsize]{}%
\psfrag{v04}[r][r][\tsize]{\small10}%
\psfrag{v05}[r][r][\tsize]{}%
\psfrag{v06}[r][r][\tsize]{\small0.1}%
\psfrag{v07}[r][r][\tsize]{}%
\psfrag{v08}[r][r][\tsize]{\small0.2}%
\psfrag{v09}[r][r][\tsize]{}%
%
\includegraphics[trim = 1mm 0mm 1mm 0mm, clip, width=0.2\textwidth]{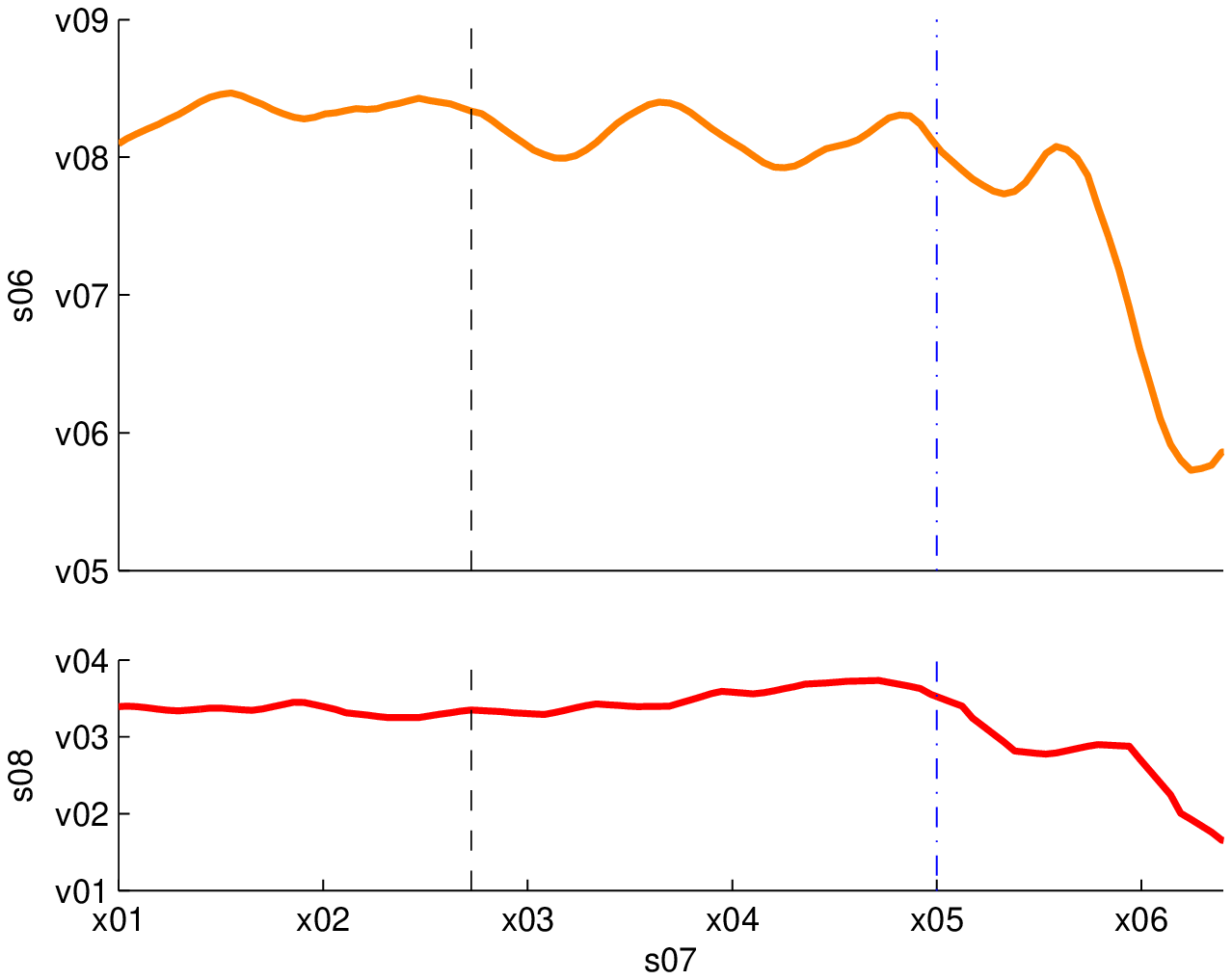}
\end{psfrags}%
%
\\ \vspace{2mm}
	\caption{Outdoor Experiment: Cross marker in the top left images shows the chosen landing spot. Bottom left presents $\hat{\epsilon}$ for all regions with color scaled according $\hat{\epsilon}$ (Yellow = obstacle; Black = non-obstacle). Top right presents the flight path of the MAV from the time when the landing strategy activated until the time when it landed. Bottom right shows the average of the roughness in 9 regions, $\hat{\epsilon}_m$ and height, $h$}
	\label{fig:AppearanceLandingStrategyOutdoor}
\end{figure}

Fig.~\ref{fig:AppearanceLandingStrategyOutdoor} shows the results of the flight test in a windy outdoor environment (wind speed $\approx$ 11 knots). The obstacles are a camp (center bottom of the image) and trees surrounding it. In this figure, the top left shows the chosen landing spot according to the position of the minimum value of the 9 $\hat{\epsilon}$ presented in the bottom left using SSL landing strategy. The top right illustrates the flight path of the MAV while bottom right represents the mean value of the 9 $\hat{\epsilon}$ and the height of the MAV. Also, the instances when the landing strategy was activated and when the landing was performed are indicated as the black and blue vertical lines, respectively. Note that in this experiment, the training was not done in this scene but in another, similar outdoor environment consisting of trees and grass field. This illustrates that the learned appearance can allow MAVs to select a landing spot from hover even in a different environment than the trained one.

\subsection{Applications}
\label{subsec:Applications}
It is common for learning methods that the results are best when the test data distribution is similar to the training data distribution. For the proposed setup of SSL, this means that the results are best when the test environment is similar in appearance to the training environment (see Section~\ref{sec:Generalization}). This implies that the approach will be most successful on MAVs that operate in a limited variation of environments, such as indoor MAVs flying in a warehouse for keeping track of stock or flying surveillance rounds in an industrial plant. Also, the approach will work well for MAVs that are always flying over forest areas for spotting live stock or flying over the same fields in an agricultural application. This being said, the results in Section~\ref{sec:Generalization} show that if the test environment is dissimilar from the training environment, this can be successfully detected by the MAV with uncertainty measures that can be determined by machine learning methods. In such a case, the MAV can decide to rely again on optical flow in order to adapt its mapping from appearance to obstacle detection.

\section{Conclusion}
\label{sec:Conclusion}
We have introduced a novel setup for SSL, in which optical flow provides the supervised outputs. The surface roughness $\epsilon^*$ from the optical flow algorithm allows obstacle detection and safe landing spot selection with a straightforward landing strategy when the MAV has lateral movement. A regression function is learned that maps texton distributions to the roughness estimate $\epsilon^*$ from optical flow algorithm. We have shown that $\hat{\epsilon}$ from SSL does not only manage to detect, but even segment obstacles in the image, without having to move. A landing strategy analyzing $\hat{\epsilon}$ in nine regions in an image is able to guide the MAV to land on an area without obstacle. Both methods using roughness estimates led to successful landings in indoor experiments with a Parrot AR drone and outdoor experiments with a Parrot Bebop, both running all vision and learning algorithms on-board. In addition, we have investigated the generalization of the SSL method to different environments. Although the results show that the generalization is achieved to some extent, we recommend the use of uncertainty measures inherent to some machine learning methods to detect when the appearance of the environment is significantly different from that of the training environment. We have shown that the Shannon entropy of a Naive Bayes classifier can allow a robot to detect this and fall back on optical flow again in order to re-adapt to the new environment. We conclude that the proposed approach to SSL has the potential to significantly extend the sensory capabilities of MAVs with a single camera.


%

%
%
%
%
%

\ifCLASSOPTIONcaptionsoff
\newpage
\fi



\bibliographystyle{myIEEEtran}
\bibliography{IEEEabrv,Arxiv_RAS_learning}
%
%
%

\end{document}